\documentclass[journal]{IEEEtran}
\ifCLASSINFOpdf \else \fi

\usepackage{graphicx}
\usepackage{subfigure}
\usepackage[cmex10]{amsmath}
\usepackage{amssymb}

\usepackage{algorithmic}
\usepackage{array}
\usepackage{amsthm,amsmath,amssymb,amsfonts}
\ifCLASSOPTIONcompsoc
 \usepackage[caption=false,font=normalsize,labelfont=sf,textfont=sf]{subfig}
\else
 \usepackage[caption=false,font=footnotesize]{subfig}
\fi
\usepackage{url}
\usepackage[noadjust]{cite}
\usepackage{multicol}
\usepackage{multirow}
\usepackage{booktabs}
\usepackage{amsmath}
\usepackage{amssymb}
\usepackage{graphicx}
\usepackage{subfigure}
\usepackage{tabularx}
\usepackage[dvipsnames]{xcolor}
\usepackage{xspace}
\makeatletter
\DeclareRobustCommand\onedot{\futurelet\@let@token\@onedot}
\def\@onedot{\ifx\@let@token.\else.\null\fi\xspace}

\def\eg{\emph{e.g}\onedot} 
\def\ie{\emph{i.e}\onedot}

\def\etal{\emph{et al}\onedot}

\newcommand{\red}[1] {\textcolor[rgb]{1.0,0.0,0.0}{{#1}}}

\makeatother

\begin{document}



\title{High-resolution Depth Maps Imaging via Attention-based Hierarchical Multi-modal Fusion}

\author{Zhiwei Zhong,
Xianming~Liu,~\IEEEmembership{Member,~IEEE,}
Junjun~Jiang,~\IEEEmembership{Member,~IEEE,}
Debin Zhao,~\IEEEmembership{Member,~IEEE,}
Zhiwen Chen,~\IEEEmembership{Member,~IEEE,}
Xiangyang Ji,~\IEEEmembership{Member,~IEEE}
\thanks{This work was supported by National Natural Science Foundation of China under Grants 61922027 and 61932022.}

\IEEEcompsocitemizethanks{
\IEEEcompsocthanksitem Z. Zhong, X. Liu, J. Jiang and D. Zhao are with the School of Computer Science and Technology, Harbin Institute of Technology, Harbin 150001, China, and also with Peng Cheng Laboratory, Shenzhen 518052, China  E-mail: \{zhwzhong, csxm, jiangjunjun\}@hit.edu.cn.
\IEEEcompsocthanksitem Z. Chen is with Taobao (China) Software Co.,Ltd., Beijing, China. E-mail: zhiwen.czw@alibaba-inc.com
\IEEEcompsocthanksitem X. Ji is with the Department of Automation, Tsinghua University, Beijing 100084, China.  E-mail: xyji@tsinghua.edu.cn.
\IEEEcompsocthanksitem Correspondence to: Xianming Liu (csxm@hit.edu.cn)}

\thanks{}}

\maketitle

\begin{abstract}
Depth map records distance between the viewpoint and objects in the scene, which plays a critical role in many real-world applications. However, depth  map captured by consumer-grade RGB-D cameras suffers from low spatial resolution.
Guided depth map super-resolution (DSR) is a popular approach to address this problem, which attempts to restore a high-resolution (HR) depth map from the input low-resolution (LR) depth and its coupled HR RGB image that serves as the guidance.
 The most challenging issue for guided DSR is how to correctly select consistent structures and propagate them, and properly handle inconsistent ones.  In this paper, we propose a novel attention-based hierarchical multi-modal fusion (AHMF) network for guided DSR. Specifically, to effectively extract and combine relevant information from LR depth and HR guidance, we propose a multi-modal attention based fusion (MMAF) strategy for hierarchical convolutional layers, including a feature enhancement block to select valuable features and a feature recalibration block to unify the similarity metrics of modalities with different appearance characteristics.  Furthermore, we propose a bi-directional hierarchical feature collaboration (BHFC)
module to fully leverage low-level spatial information and high-level structure information among multi-scale features.  Experimental results show that our approach outperforms state-of-the-art methods in terms of reconstruction accuracy, running speed and memory efficiency.
\end{abstract}


\IEEEpeerreviewmaketitle

\section{Introduction}

\IEEEPARstart{D}epth information plays a critical role in a myriad of applications such as autonomous driving~\cite{caesar2020nuscenes},  virtual reality~\cite{meuleman2020single}, 3D reconstruction~\cite{Hou_2019_CVPR} and scene understanding~\cite{8738849}. In recent years, with the progress of sensing technology, depth maps can be readily captured by consumer-grade 
depth cameras such as Time-of-Flight (ToF) and Microsoft Kinect. However, the depth map taken from these commercialized cameras usually suffers from low-resolution, which hinders the subsequent depth based applications. Therefore, depth map super-resolution (DSR) has raised a lively interest in the communities of academia and industry.

DSR is inherently an ill-posed problem, as there exist multiple high-resolution (HR) depth maps corresponding to the same low-resolution (LR) degradation. To solve this inverse problem, one popular approach is guided DSR, considering that in practice many smartphones and robots are equipped with a conventional RGB camera as well as a depth camera. The former acquires an intensity image with higher spatial resolution than the depth map. Since the targets they shoot are the same scene, it is thus natural to enhance the resolution of depth by transferring structure from the HR guidance image.  Specifically, guided DSR aims to apply HR guidance image as a prior for reconstructing regions in depth where there is semantically-related and structure-consistent content, and fall back to a plausible reconstruction for regions in depth with inconsistent content of the guidance.

To achieve this goal, two major problems should be carefully addressed. Firstly, it is challenging to select reference structures and propagate them properly by defining hand-crafted rules. Secondly, the guidance based approach makes a basic assumption that the guidance image should contain correct mutual structural information. However, the guidance could be insufficient, or even wrong locally. It is challenging to handle the structure inconsistency problem. For regions with inconsistent structures, it is expected that guidance based approach could reduce the wrong influence of the guidance and predict HR reconstruction properly.

For the first issue, data-driven based strategies have been proposed to remedy the difficulty of hand-crafted design. For instance, Hui \textit{et al.}~\cite{DMSG} proposed a multi-scale guided convolutional network (DMSG) for DSR, which fuses rich hierarchical features at different levels to generate accurate HR depth map. Similarly, in \cite{DepthSR}, Guo \textit{et al.} proposed a DSRNet to infer a HR depth map from its LR version by hierarchical features driven residual learning. Su \textit{et al.}~\cite{PanNet} proposed a pixel-adaptive convolution based network (PacNet), which is actually a fine-grained filtering operation that can effectively learn to leverage guidance information. Similar strategy that uses pixel-wise transformation also appears in \cite{lutio2019guided}. The work \cite{PMBANet} presented a progressive multi-branch aggregation network (PMBAN) for depth SR, which consisted of stacked multi-branch aggregation blocks to progressively recover the degraded depth map. 

For the second issue, recent efforts focus on learning-based selection strategies for the common structures existing in both the target and guidance images. For instance, Li \textit{et al.}~\cite{DJFR} proposed a joint image filtering with deep convolutional networks, which can selectively transfer salient structures that are consistent with multi-modal inputs. The network architecture of DKN \cite{DKN} is similar to DJFR~\cite{DJFR}, but contains a weight and offset learning module to explicitly learn the sparse and spatially-variant kernels. Recently, Deng \textit{et al.} \cite{CUNet} proposed a common and unique information splitting network (CUNet) to automatically determine the common information among different modalities, according to which an adaptive fusion operation was performed.

Although significant progress has been achieved, learning-based DSR is still an open problem. The key challenge is how to achieve a good balance among performance, running time and network complexity, so as to promote its usage in practical scenarios. In this paper, we embrace this challenge and propose a novel attention-based hierarchical multi-modal fusion (AHMF) network to perform structure selection, propagation and prediction simultaneously for guided DSR. Specifically, to effectively explore and combine relevant information from LR depth and HR guidance, we propose a multi-modal attention based fusion (MMAF) for hierarchical convolutional layers. It consists of a feature enhancement block that is tailored to adaptively select useful information and filter out unwanted ones, such as texture information in guidance and noise in depth that would disturb the depth reconstruction; and a feature recalibration block that is designed to adaptively rescale enhanced features to unify the similarity metrics of modalities with different appearance characteristics. Furthermore, considering that in CNN shallower layers encode rich spatial details but lack semantic knowledge while deeper layers are more effective to capture high-level context and structure information but lose spatial information, we propose a bi-directional hierarchical feature collaboration (BHFC) module to fully leverage the complementarity of the hierarchical fused features. To verify the effectiveness of the proposed method, we conduct experiments on widely used benchmark datasets, and the experimental results demonstrate that our method achieves superior performance than the state-of-the-arts. 

\begin{figure}[!tb]
\begin{center}
   \includegraphics[width=0.9\linewidth]{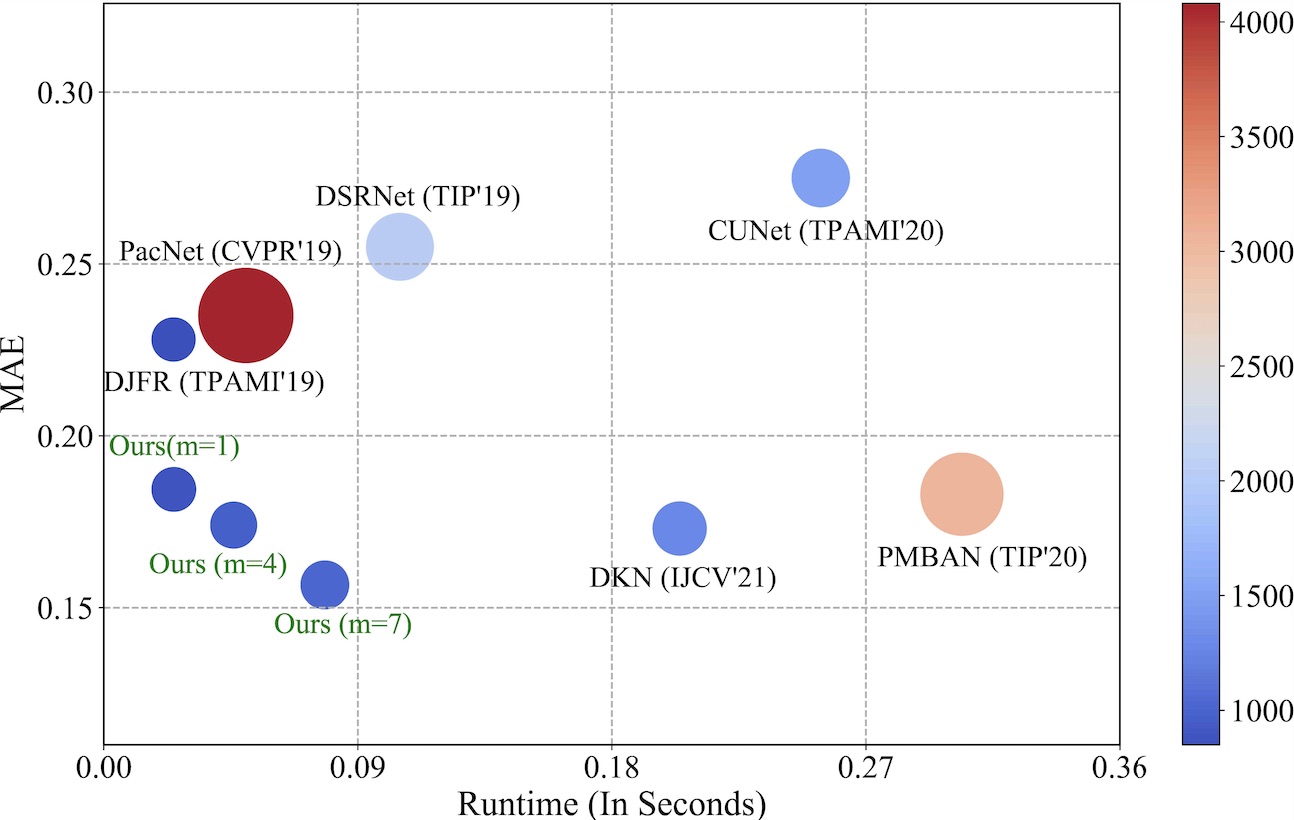}
   \vspace{-0.1in}
   \caption{Comparison of state-of-the-art methods for 4$\times$ DSR on Middlebury 2005~\cite{middleblur_data_1} dataset in terms of MAE (the lower the better), running time and peak GPU memory consumption that is indicted by radius of circles. Our method includes three cases with different $m$, which refers to the number of layers for multi-modal feature fusion. The experiments are evaluated on NVIDIA 1080ti GPU with depth map size 480$\times$640.}
   \label{fig:time_mae}
   \vspace{-0.3in}
\end{center}
\end{figure}

{The main contributions of our method can be summarized as follows: 
\begin{itemize}
    \item We propose a multi-modal attention based fusion module, which can adaptively select and effectively fuse features extracted from the depth and guidance images. It contains a feature enhancement block to select valuable information and a feature recalibration block to rescale multi-modal features. Contrary to existing methods that fuse multi-modal features by simple concatenation or summation, the proposed method can avoid transferring erroneous structures that are not existed in the depth image, \ie, the texture-copying artifacts.
    \item We propose a bi-directional hierarchical feature collaboration module, which can facilitate the hierarchical fused features propagation and collaboration with each other.
    \item We propose an \underline{a}ttention-based \underline{h}ierarchical \underline{m}ulti-modal \underline{f}usion framework (AHMF) for guided depth map super-resolution. With the proposed MMAF and BHFC, the proposed method can effectively explore the complementarity of multi-level and multi-modal features. As shown in Fig.~\ref{fig:time_mae}, our method achieves better DSR performance, faster running speed and moderate memory consumption over state-of-the-art methods.
\end{itemize}}

The remainder of this paper is organized as follows. Section~\ref{rw} reviews related work in the literature.  Section~\ref{ow} introduces the overview of the proposed method. Then, we elaborate on the proposed multi-modal attention based fusion in Section~\ref{mmaf} and hierarchical feature collaboration strategy in Section~V. Extensive experimental results are presented in Section~\ref{hfc}. Finally, Section~\ref{con} concludes this paper. 

\begin{figure*}[!tb]
    \begin{center}
        \includegraphics[width=0.8\linewidth]{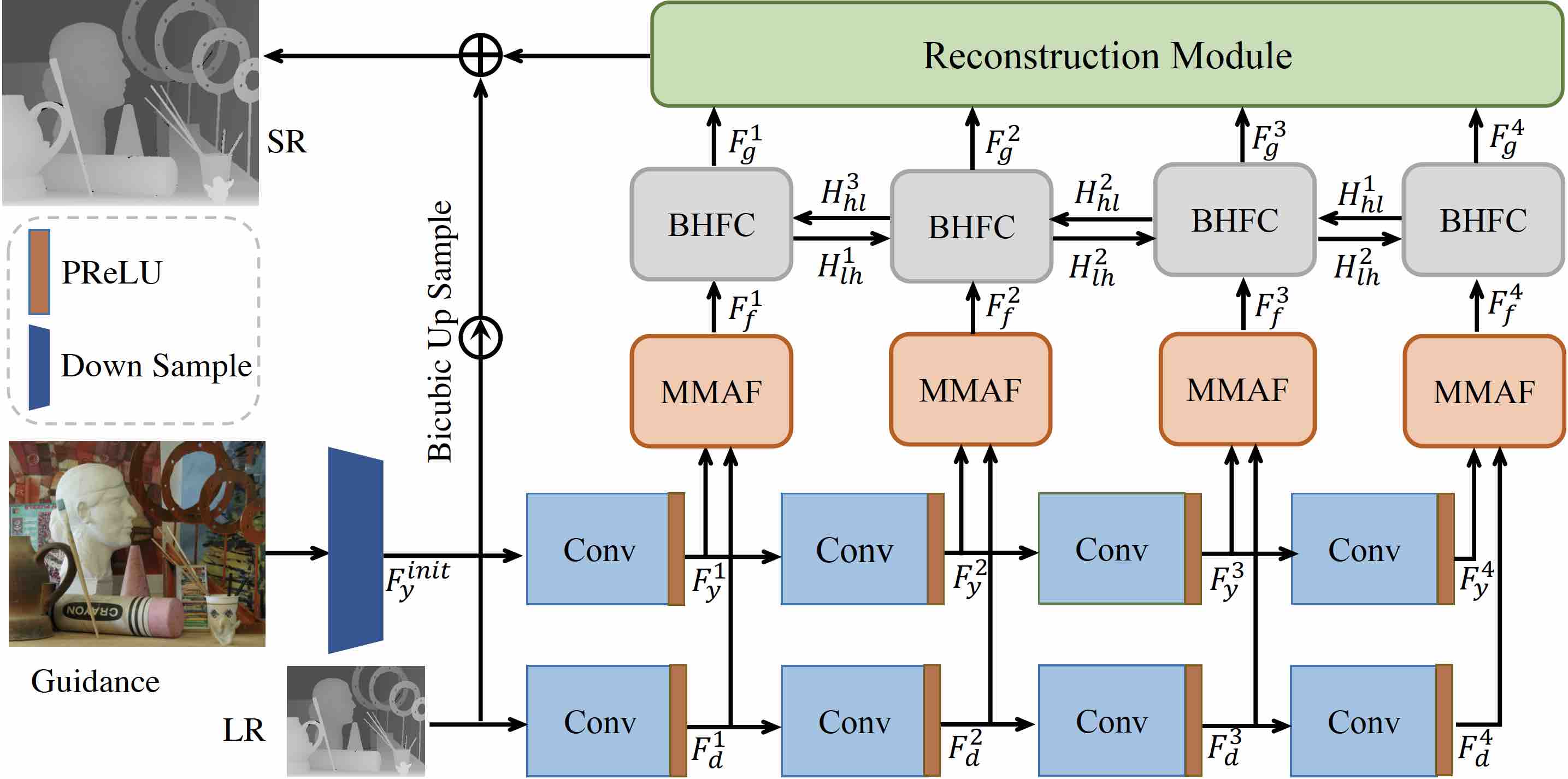}
    \end{center}
    \vspace{-0.13in}
	\caption{The network architecture of our proposed attention-based hierarchical multi-modal fusion (AHMF) network, where MMAF represents the proposed multi-modal attention based fusion module and BHFC represents the bi-directional hierarchical feature collaboration module.}
	\label{fig:net}
	\vspace{-0.25in}
\end{figure*}

\section{Related Work}
\label{rw}
\subsection{Guided Depth Super-resolution}
In the literature, many works have been developed for guided DSR, which can be roughly divided into three categories: filtering based, optimization based and deep learning based. The filtering based approaches, such as~\cite{JBU, 6618873}, reconstruct the depth map by means of the weighted average of target image values with a local filter. Although these filtering based methods are at low computational cost, they cannot maintain the global information as the weights of the filter are calculated by the local content in the guidance image. This would inevitably transfer inaccurate structures to the depth image when the assumption of structure similarity is invalid. Different from filtering based methods, optimization based methods formulate depth image super-resolution as a global optimization problem. They adopt various priors to constrain the high-dimensional solution space for this ill-posed problem. For example, Ferstl~\etal ~\cite{Ferstl2013Image} formulated the DSR problem as a convex optimization problem and employed a high-order Total Generalized Variation (TGV) regularization to get a piece-wise smooth solution. Liu~\etal~\cite{8474366} proposed a DSR method by combining both internal and external priors formulated in the graph domain. However, these hand-crafted priors suffer from limitations in modeling the real world image degradation process. Moreover, solving the optimization problem is usually time-consuming due to the iterative computation process.

Based on U-Net architecture, Guo~\etal~\cite{DepthSR} proposed a hierarchical feature driven network and they claimed that their network could make full use of features extracted from depth and guided image compared to the existing methods. To learn the potential relationship between the guidance and the depth images, Zuo~\etal~\cite{mfr-sr} proposed a coarse-to-fine network by combining both global and local residual learning strategies. Li~\etal~\cite{CMSR} introduced a multi-scale symmetric network, which includes a symmetric unit to restore edge details and a correlation-controlled color guidance block to investigate the inner-channel correlation between the depth and guidance sub-network. Kim~\etal~\cite{DKN} proposed a deformable kernel network for depth SR, they employed the kernel based method but the weights and offsets of the kernel were learned by the network automaticly. Su~\etal~\cite{PanNet} argued that the convolutional operation was content-agnostic and proposed a pixel-adaptive convolution to address this problem, and they conducted plenty of experiments to show that their method could easily adapt to a lot of applications such as joint upsampling, semantic segmentation and CRF inference.

Besides the color-guided depth super-resolution considered in this work, in the field of 3D imaging, there is another approach to achieve high-quality depth imaging by fusing high-quality RGB image and noisy 3D single-photo avalanche detector (SPAD) arrays~\cite{lindell2018single,sun2020spadnet,chan2019long,ruget2021robust}. Our method can also be applied to treat this problem by modifying accordingly to meet the requirements of RGB-SPAD fusion, including:
\begin{enumerate}
    \item Replace the 2D convolution layers for depth image features extraction with 3D convolution layers.
    \item Redesign the multi-modal feature fusion module (MMAF in our model) to make it capable of fusing 2D and 3D data, i.e., add a 2D-3D up-projection module as mentioned in \cite{sun2020spadnet} at the head of MMAF to expand the temporal dimension for 2D features.
    \item  Decrease the temporal dimensions to one before the final reconstruction module for the purpose of generating 2D depth image.
\end{enumerate}
Since this is out of the scope of this paper, we leave the investigation of the effectiveness of our method for RGB-SPAD fusion in the future work.


\subsection{Attention Model}
\par Attention mechanism has shown remarkable performance in deep convolutional neural networks and has been introduced in a large number of computer computer vision tasks. Zhang~\etal~\cite{SAGAN} proposed a self-attention based GAN architecture for image generation problem, the experimental results showed that the self-attention module could effectively capture long-range dependencies. Inspired by the non-local mean method, Wang~\etal~\cite{8578911} proposed a non-local neural network for video classification and the self-attention could be viewed as a special case of it. Hu~\etal~\cite{SE} proposed a squeeze-and-excitation module to let the network pay more attention to important feature maps by explicitly modelling channel-wise interrelations. Li~\etal~\cite{Li_2019_CVPR} proposed a selective kernel network (SKNet) to let the neural adjust its receptive field adaptively. Mei~\etal~\cite{mei2020pyramid} proposed a pyramid attention module for image restoration, their method could capture long range correspondences  in a multi-level fashion. To accelerate the channel attention and decrease the model complexity, Wang~\etal~\cite{ECA} replaced the fully connected layer used in the channel attention module with a 1D convolution. Yu~\etal~\cite{9010689} proposed a gated convolution to solve the problem of the traditional convolutional layer treating all input pixels equally for free-from image inpainting tasks, then Chang~\etal~\cite{9009508} extended the 2D gated convolution to 3D part for free-from video inpainting. Among these works, the most related to our method is \cite{Li_2019_CVPR}, however, there are still some differences between them, firstly the feature recalibration block in our method is to rescale the different modal data while \cite{Li_2019_CVPR} aims to dynamically select the receptive field for each neuron. Secondly, we use mean and standard deviation pooling instead of max pooling used in \cite{Li_2019_CVPR} to get global embedding, which is more suitable for low-level vision. 

\subsection{Multi-level Feature Fusion}
\par In deep convolutional networks, the features of shallower layers usually contain low-level details while the ones of deeper layers are composed of high-level semantic information. In order to make the best use of both high and low level features, enormous methods are proposed. He~\etal~\cite{ResNet} introduced a skip-connection which was also called residual learning operator for training very deep convolutional networks, with the help of this operation, the deeper layer can directly access the low-level features. Instead of pixel-wise addition, Huang~\etal~\cite{8099726} concatenated the low-level and high-level features to fuse the multi-level features. Gu~\etal~\cite{Gu_2019_ICCV} proposed a self-guided network for image denoising, they used high-level features to progressively guide the low-level feature, which could enlarge the receptive field of the shallower layer and enhance the network representing capability. Lin~\etal~\cite{lin2019refinenet} proposed a refinement network in which the low-level features were refined by the high-level semantic levels. Although the performance of the network is greatly improved, there are still some problems, for example, in \cite{ResNet} and \cite{8099726} the low-level features cannot directly contact with high-level information due to the feed-forward nature of convolutional neural network, on the contrary, in the methods of \cite{lin2019refinenet} and \cite{Gu_2019_ICCV} only the low-level features can be refined by the high-level features. To solve these problems, in this paper, we propose a bi-directional hierarchical feature collaboration module, in which the low-level feature and high-level can propagate to each other effectively. 

\section{Overview of the Proposed Method}
\label{ow}
In this section, we provide an overview about the proposed attention-based hierarchical multi-modal fusion network.

Our model takes a LR depth map $\boldsymbol{D}_{LR} \in \mathbb{R}^{H \times W \times 1} $ and a HR guidance image $\boldsymbol{Y} \in \mathbb{R}^{\alpha H \times \alpha W \times C} $ as inputs, where $\alpha$ is the upscale factor, $H,W$ and $C$ represent the height, width and the number of channels respectively. The pipeline of our network is illustrated in Fig.~\ref{fig:net}, consisting of five major modules that are marked by different colors, which are tailored to address the corresponding issues in color guided depth SR:
\begin{itemize}
    \item \textbf{Module of guidance image downsampling.} Considering that the resolution of guidance $\boldsymbol{Y}$ is higher  ($\alpha  \times$) than the corresponding depth $\boldsymbol{D}_{LR}$, to facilitate the subsequent processing, this module is tailored to downsample $\boldsymbol{Y}$ to achieve resolution consistency. Instead of using traditional downsampling strategy such as bicubic, which would result in information loss, we propose to leverage inverse pixel-shuffle \cite{invp} to progressively downsample $\boldsymbol{Y}$ with the upscale factor $\alpha$. In this way, it can not only preserve all the original information of $\boldsymbol{Y}$ but also achieve resolution balance of two inputs. More specifically, taking $\alpha=4$ as an example, $\boldsymbol{Y}$ needs to be downsampled by $2~(\log_2\alpha) $ times. We first employ a $3 \times 3$ convolution layer to expand the feature channels of the guidance image to obtain $\boldsymbol{\hat{Y}}$, then the downsampling process can be described as follows:
\begin{equation}
    \boldsymbol{F}_{y}^{init} = \mathrm{Down} (\sigma (\boldsymbol{W}_{y}^{0} * \mathrm{Down}(\boldsymbol{\hat{Y}} )+ b_{y}^{0})),
\end{equation}
where $\mathrm{Down}(\cdot)$ represents the inverse pixel-shuffle operator with downscale factor 2; $\boldsymbol{W}_y^0$ is a 1$\times$1 convolutional kernel; $b_y^0$ is the bias term; $*$ means convolutional operation; 
$\sigma$ represents a Parametric Rectified Linear Unit (PReLU)~\cite{PReLU} activation function. %
\item \textbf{Module of feature extraction.} This module is employed to fully extract meaningful features from both depth and guidance in a multi-level fashion:
\begin{align}
    \boldsymbol{F}_d^1 &= \sigma \left( \boldsymbol{W}_d^1 * \boldsymbol{D}_{LR} + b_d^1 \right), \\
    \boldsymbol{F}_y^1 &= \sigma \left( \boldsymbol{W}_y^1 * \boldsymbol{F}_{y}^{init}  + b_y^1 \right), \\
    \boldsymbol{F}_d^i &= \sigma \left( \boldsymbol{W}_d^{i} * \boldsymbol{F}_{d}^{i-1}  + b_d^{i} \right), 1< i \leq m, \\
    \boldsymbol{F}_y^i &= \sigma \left( \boldsymbol{W}_{y}^{i}  * \boldsymbol{F}_{y}^{i-1} + b_y^{i} \right), 1< i \leq m,
\end{align}
where $\boldsymbol{W}_d^i$ and $\boldsymbol{W}_y^i$ are convolutional kernels of $i$-th layer that are used for depth and guidance feature extraction respectively; $b_d^i$ and $b_y^i$ are the bias terms; $m$ refers to the number of layers for feature extraction.
\item \textbf{Module of multi-modal attention based fusion.} After extracting multi-modal features, the following question is how to fuse them properly. We do this by the proposed multi-modal attention based fusion (MMAF):
\begin{align}
    \boldsymbol{F}_{f}^i = \mathrm{MMAF}_{i} (\boldsymbol{F}_y^i, \boldsymbol{F}_d^i), 1\leq i \leq m,
    \label{eq6}
\end{align}
which considers the difference between guidance and depth and adaptively combines multi-modal features by the attention mechanism. In  this  way,  it optimally  preserves  consistent  structures  and suppresses inconsistent components in a learning manner.
\item \textbf{Module of bi-directional hierarchical feature collaboration.} MMAF is followed by the proposed bi-directional hierarchical feature collaboration (BHFC), which is tailored to further jointly leverage low-level spatial and high-level semantic information:
\begin{align}
    \boldsymbol{F}_g^i = \mathrm{BHFC}_{i}([\boldsymbol{F}_{f}^i, \boldsymbol{H}_{lh}^{i-1}, \boldsymbol{H}_{hl}^{m-i}]), 1\leq i \leq m,
\end{align}
where $\boldsymbol{H}$ is hidden state and the subscripts $lh$ and $hl$ denote the information propagation direction. $\boldsymbol{H}^0$ is the initial state and is set as zero. 
\item \textbf{Module of final HR depth reconstruction.} Finally, with the refined features $\{\boldsymbol{F}_g^i \}_{i=1}^m$, we arrive at the HR depth reconstruction module, which produces the upsampled depth image by pixel-shuffle \cite{PixelShuffle}:
\begin{align}
	\boldsymbol{F}_{up} &= \mathrm{Up}\left(\sigma(\boldsymbol{W}_{cr} * [\boldsymbol{F}_g^1, \boldsymbol{F}_g^2, \cdots, \boldsymbol{F}_g^m] + b_{cr})\right) \\
	\boldsymbol{D}_{SR} &= \sigma(\boldsymbol{W}_{out} *\boldsymbol{F}_{up} + b_{out}) + \boldsymbol{D}_{LR}^{\uparrow},
\end{align}
where $\boldsymbol{W}_{cr}$ and $b_{cr}$ are the weight and bias of a 1$\times$1 convolutional layer for channel reduction; $\mathrm{Up}(\cdot)$ is denoted as the up-sampling module which employ the pixel-shuffle operation to progressive up-sample the input features; $[\cdot]$ is used to concatenate the refined features; $\boldsymbol{W}_{out}$ and $b_{out}$ are convolution kernels and bias, respectively.  $\boldsymbol{D}_{LR}^{\uparrow}$ is the bicubic upsampled version of the input LR depth image; $\boldsymbol{D}_{SR} \in \mathbb{R}^{\alpha H \times \alpha W \times 1}$ is the final reconstructed HR depth image. 
\end{itemize}

The network is trained using the following loss function:
\begin{equation}
    \boldsymbol{L}(\boldsymbol{\theta}) = \frac{1}{N} \sum_{i=1}^N ||\mathcal{F}(\boldsymbol{D}_{LR}, \boldsymbol{Y}; \boldsymbol{\theta}) - \boldsymbol{D}_{GT} ||_1,
\end{equation}
where $\mathcal{F}(\cdot)$ represents the overall network architecture and $\boldsymbol{\theta}$ are network parameters. $\boldsymbol{D}_{GT}$ is the ground-truth HR depth map, $N$ is the number of training samples.

\begin{figure}[!tb]
    \centering
	\includegraphics[width=0.9\linewidth]{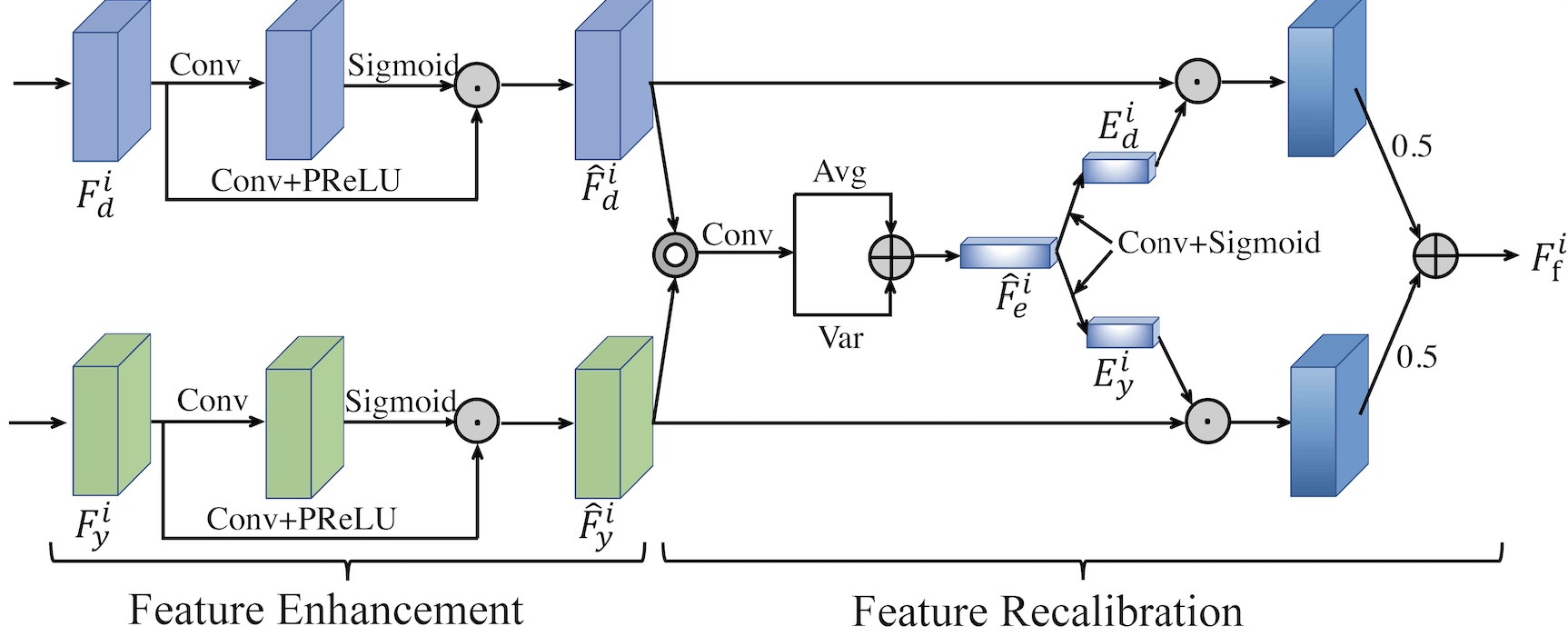}
	\vspace{-0.17in}
	\caption{Multi-modal attention based fusion (MMAF), where $\odot$ is pixel-wise multiplication, $\oplus$ is pixel-wise summation and $\circledcirc$ denotes concatenation. Avg and Var mean the average and variance pooling. The MMAF takes the depth and guidance features as inputs and outputs the fused features.}
	\label{fig:mmam}
	\vspace{-0.25in}
\end{figure}

\section{Multi-modal Attention based  Fusion}
\label{mmaf}
In this section, we introduce the proposed multi-modal attention based fusion strategy. In guided DSR, the core is the act of extracting and combining relevant information from LR depth and HR guidance so as to derive superior performance over using only depth.
One can perform concatenation or summation of the outputs of two separate branches for depth and guidance, respectively, and then use uni-modal CNNs to get the final reconstruction, as done in \cite{DJFR}. Some works suggest mid-level feature fusion could benefit reconstruction. For instance, \cite{DMSG,DepthSR,PMBANet} propose to fuse the hierarchical features of convolutional layers of two branches, which is still conducted by concatenation. However, the intermediate level features of depth and guidance have different semantic meanings,  making the intermediate fusion more challenging. The simple concatenation is not effective for this purpose.

Different from existing guided DSR methods, we propose to hierarchically combine multi-modal features by attention mechanism. We attempt to preserve consistent structures and suppress inconsistent components in a learning manner. As shown in Fig.~\ref{fig:mmam}, our proposed fusion scheme
consists of a feature enhancement block and a feature recalibration block, which are elaborated in the following.  

\textbf{Feature Enhancement Block}: The guidance image contains rich texture information, which would disturb the depth reconstruction, while depth itself contains noise due to the limitation of sensors. In view of these, we propose to leverage feature enhancement block (FEB) to adaptively select useful information and filter out unwanted one. Specifically, inspired by \cite{9010689}, we leverage gated units as the FEB. Given the extracted multi-modal features $\boldsymbol{F}_d^i$ and $\boldsymbol{F}_y^i (1 \leq i \leq m)$ as inputs, the process of FEB can be formulated as follows:
\begin{align}
    \small
    \label{FEB}
    \hat{\boldsymbol{F}}_d^i & = \sigma ( \boldsymbol{W}_{d,1}^i * \boldsymbol{F}_d^i + b_{d,1}^i ) \odot \phi ( \boldsymbol{W}_{d,2}^i * \boldsymbol{F}_d^i + b_{d,2}^i ), \\
    \hat{\boldsymbol{F}}_y^i & = \sigma ( \boldsymbol{W}_{y,1}^i * \boldsymbol{F}_y^i + b_{y,1}^i ) \odot \phi ( \boldsymbol{W}_{y,2}^i * \boldsymbol{F}_y^i + b_{y,2}^i ),
\end{align}
where $\boldsymbol{W}_{d,1}^i$, $\boldsymbol{W}_{d,2}^i$ and $\boldsymbol{W}_{y,1}^i$, $\boldsymbol{W}_{y,2}^i$ are convolutional kernels for depth and guidance respectively, the subscripts $1$ and $2$ represent two different convolutional operations; $b_{d,1}^i$, $b_{d,2}^i$ and $b_{y,1}^i$, $b_{y,2}^i$ are the corresponding bias terms;  $\odot$ denotes element-wise multiplication; $\phi$ is sigmoid function to limit the output of the gating operation within the range of 0 and 1;  $\sigma$ is PReLU  activation function; $\hat{\boldsymbol{F}}_d^i$ and $\hat{\boldsymbol{F}}_y^i$ are the enhanced features of depth and guidance respectively. 

\textbf{Feature Recalibration Block}: Considering depth and guidance images exhibit notably different appearance characteristics due to the difference in imaging principle, similar pixels in guidance image may have quite different values in depth and vice versa. To unify the similarity metrics, we employ multi-modal feature recalibration block (FRB) to recalibrate the enhanced multi-modal features by FEB. The FRB consists of two units: \textit{multi-modal feature squeeze unit}, which is tailored to learn global joint knowledge from all modalities, and \textit{feature excitation unit}, which uses the learned joint knowledge to adaptively emphasize useful features. Specifically, the multi-modal squeeze unit can be formulated as follows:
\begin{align}
    \hat{\boldsymbol{F}}_c^i &= \sigma (\boldsymbol{W}_c^i * ([\hat{\boldsymbol{F}}_d^i, \hat{\boldsymbol{F}}_y^i]) + b_c^i), \\
    \hat{\boldsymbol{F}}_e^i &= \mathrm{AvgPool}~(\hat{\boldsymbol{F}}_c^i ) + \mathrm{VarPool}~(\hat{\boldsymbol{F}}_c^i),
\end{align}
where $[\cdot,\cdot]$ denotes concatenation operation;  $\boldsymbol{W}_c^i$ is a 3$\times$3 convolutional kernel;  $\text{AvgPool}(\cdot)$ and $\text{VarPool}(\cdot)$ mean average pooling and variance pooling respectively.  $\hat{\boldsymbol{F}}_e^i$ is the global joint knowledge learned from all modalities, which is further fed into the feature excitation unit:
\begin{align}
    \boldsymbol{E}_d^i &=  \phi (\boldsymbol{W}_{d}^i * \hat{\boldsymbol{F}}_e^i + b_{d}^i), \\
    \boldsymbol{E}_y^i &=  \phi (\boldsymbol{W}_{y}^i * \hat{\boldsymbol{F}}_e^i + b_{y}^i), \\
    \boldsymbol{F}_{f}^i &= 0.5 \times \boldsymbol{E}_d^i \odot \hat{\boldsymbol{F}}_d^i  + 0.5 \times \boldsymbol{E}_y^i \odot \hat{\boldsymbol{F}}_y^i,
\end{align}
where $\boldsymbol{W}_{d}^i$, $\boldsymbol{W}_{y}^i$ and $b_{d}^i$, $b_{y}^i$ are 1$\times$1 convolutional kernels and bias terms for depth and guidance respectively; $\phi$ is the sigmoid function; $\boldsymbol{E}_d^i$ and $\boldsymbol{E}_y^i$ are the excitation signals for depth feature and guidance feature respectively; $\boldsymbol{F}_{f}^i$ denotes the fused multi-modal feature.

\begin{figure}[!tb]
    \centering
	\includegraphics[width=0.9\linewidth]{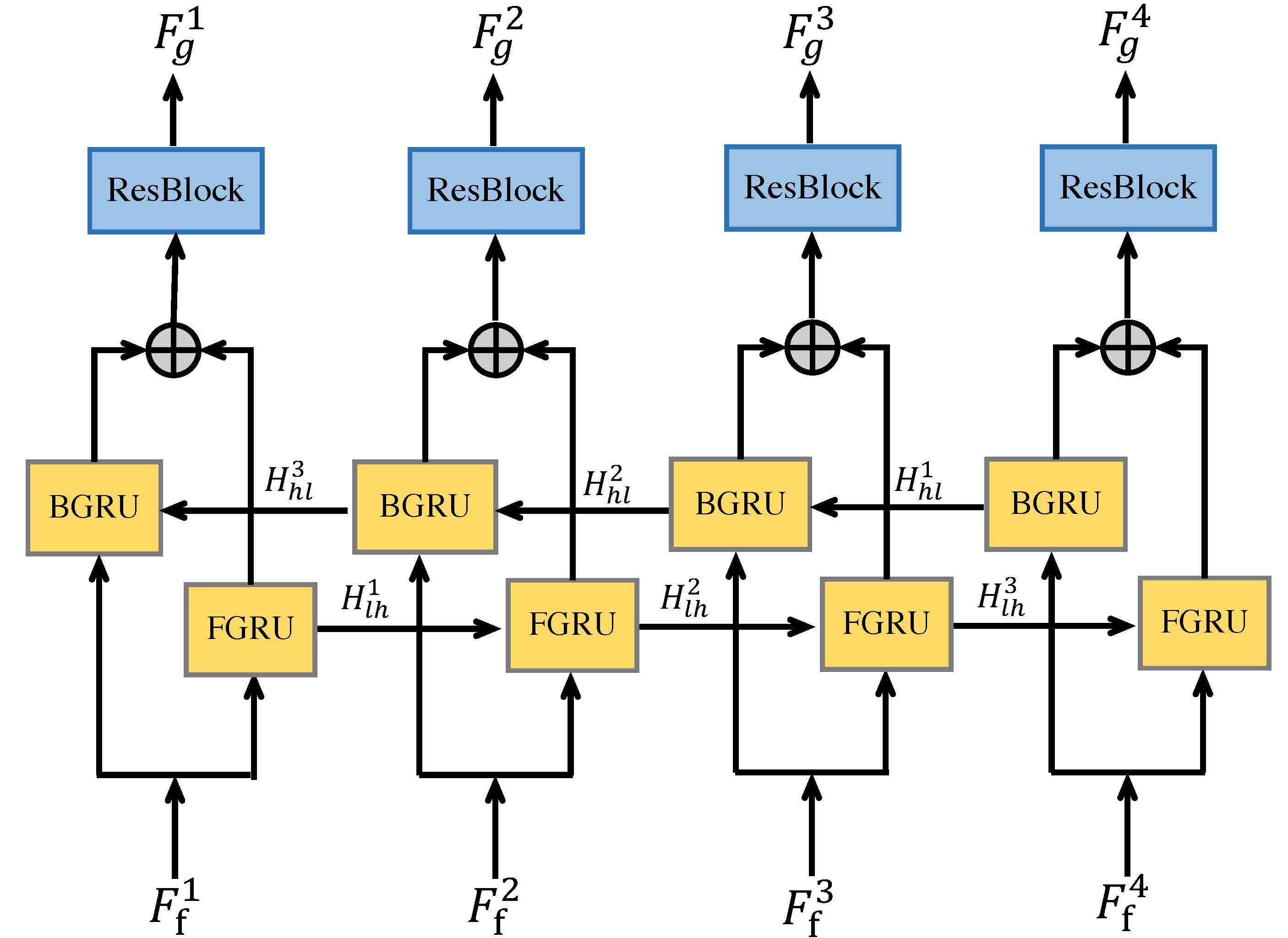}
	\vspace{-0.17in}
	\caption{Hierarchical feature collaboration with four BHFCs, where $\oplus$ means pixel-wise summation. It takes the fused multi-modal features as inputs and sends them to a bi-directional GRU for multi-level features collaboration.}
	\label{fig:bf}
	\vspace{-0.24in}
\end{figure}
\begin{table*}[!tb]\setlength{\tabcolsep}{3.6pt}
    \centering
    \caption{\label{tab:mpi_result}MAE performance comparison for scale factors $4 \times, 8 \times$ and $16 \times$ with bicubic degradation on Middlebury dataset. The best performance is shown in \textbf{bold} and second best performance is the \underline{underscored} ones (Lower MAE values, better performance).}
    \vspace{-0.1in}
    \renewcommand{\arraystretch}{1.2}
    \begin{tabular}{l|ccc|ccc|ccc|ccc|ccc|ccc|ccc}
    \toprule
    \multirow{2}{*}{Method} & \multicolumn{3}{c|}{Art} & \multicolumn{3}{c|}{Books} & \multicolumn{3}{c|}{Dools} & \multicolumn{3}{c|}{Laundry} & \multicolumn{3}{c|}{Mobeius} & \multicolumn{3}{c|}{Reindeer} & \multicolumn{3}{c}{Average} \\
    \cline{2-22}
    ~ &$4\times$ & $8 \times$  & $16\times$  & $4\times$ & $8 \times$  & $16\times$  &$4\times$ & $8 \times$  & $16\times$ &$4 \times$ & $8 \times$  & $16\times$ &$4\times$ & $8 \times$  & $16\times$ &$4\times$ & $8 \times$  & $16\times$ & $4\times$ & $8 \times$ & $16\times$ \\
    \hline
    Bicubic & 1.10 & 2.10 & 4.02 & 0.35 & 0.67 & 1.29 & 0.36 & 0.68 & 1.25 & 0.60 & 1.12 & 2.13 & 0.36 & 0.70 & 1.32 & 0.58 & 1.09 & 2.18 & 0.558 & 1.060 & 2.032 \\   
    CLMF1~\cite{clmf} & 0.74 & 1.44 & 2.87 & 0.28 & 0.51 & 1.02 & 0.34 & 0.66 & 1.01 & 0.50 & 0.80 & 1.67 & 0.29 & 0.51 & 0.97 & 0.51 & 0.84 & 1.55 & 0.443 & 0.793 & 1.515  \\
    ATGV~\cite{AGTV} & 0.65 & 0.81 & 1.42 & 0.43 & 0.51 & 0.79 & 0.41 & 0.52 & 0.56 & 0.37 & 0.89 & 0.94 & 0.38& 0.45 & 0.80 & 0.41& 0.58 & 1.01 & 0.442 & 0.627 & 0.920 \\
    DMSG~\cite{DMSG} & 0.46 & 0.76 & 1.53 & 0.15 & 0.41 & 0.76 & 0.25 & 0.51 & 0.87 & 0.30 & 0.46 & 1.12 & 0.21 & 0.43 & 0.76 & 0.31 & 0.52 & 0.99 & 0.280 & 0.515 & 1.005 \\
    DGDIE~\cite{gu2017learning}&0.48&1.20&2.44&0.30&0.58&1.02&0.34&0.63&0.93&0.35&0.86&1.56&0.28&0.58&0.98&0.35&0.73&1.29 & 0.350 & 0.763 & 1.370 \\
    DEIN~\cite{DEIN} & 0.40 & 0.64 & 1.34 & 0.22 & 0.37 & 0.78 & 0.22 & 0.38 & 0.73 & 0.23 & 0.36 & 0.81 & 0.20 & 0.35 & 0.73 & 0.26 & 0.40 & 0.80 & 0.255 & 0.417 & 0.865 \\
    DSRNet~\cite{DepthSR} & 0.63 & 1.24 & 2.44 & 0.23 & 0.45 & 0.84 & 0.24 & 0.47 & 0.85 & 0.36 & 0.70 & 1.35 & 0.24 & 0.45 & 0.87 & 0.38 & 0.69 & 1.19 & 0.347 & 0.667 & 1.257 \\
    CCFN~\cite{wen2018deep} &0.43&0.72&1.50&0.17&0.36&0.69&0.25&0.46&0.75&0.24&0.41&0.71&0.23&0.39&0.73&0.29&0.46&0.95 & 0.268 & 0.467 & 0.888\\
    GSPRT~\cite{lutio2019guided} &0.48&0.74&1.48&0.21&0.38&0.76&0.28&0.48&0.79&0.33&0.56&1.24&0.24&0.49&0.80&0.31&0.61&1.07 &  0.308 & 0.543 & 1.023\\
    DJFR~\cite{DJFR}  & 0.35 & 0.76 & 1.88 & 0.17 & 0.34 & 0.74 & 0.22 & 0.41 & 0.79 & 0.21 & 0.48 & 1.10 & 0.19 & 0.37 & 0.75 & 0.23 & 0.44 & 0.99 & 0.228 & 0.467 & 1.042 \\ 
    PacNet~\cite{PanNet} &  0.34 & 1.12 & 2.13 & 0.19 & 0.62 & 1.13 & 0.23 & 0.70 & 1.18 & 0.22 & 0.74 & 1.25 & 0.19 & 0.51 & 0.92 & 0.24 & 0.74 & 1.20  & 0.235 & 0.738 & 1.302\\
    CUNet~\cite{CUNet}  & 0.38 & 0.99 & 2.34 & 0.23 & 0.54 & 1.41 & 0.28 & 0.60 & 1.24 & 0.30 & 0.72 & 1.85 & 0.21 & 0.47 & 1.08 & 0.25 & 0.59 & 1.37 & 0.275 & 0.652 & 1.548  \\
    MDSR~\cite{MDDR} & 0.46&0.62& 1.87& 0.24& 0.37&0.73& 0.29& 0.51 &0.79 & 0.32 & 0.53 & 1.11& 0.19& 0.37& 0.74&0.41 & 0.55 & 0.95 & 0.318 & 0.492 & 1.032 \\
    PMBAN~\cite{PMBANet} & 0.26 & \textbf{0.51} & \textbf{1.22} & 0.15 & \underline{0.26} & 0.59 & 0.19 & \textbf{0.32} & \textbf{0.59} & 0.17 & 0.34 & \textbf{0.71} & 0.16 & \textbf{0.26} & 0.67 & \underline{0.17} & 0.34 & 0.74 & 0.183 & \underline{0.338} & \underline{0.753} \\
    DKN~\cite{DKN} & \underline{0.25} & \underline{0.52} & 1.34 & 0.14 & 0.27 & \underline{0.58} &  \underline{0.17} & \textbf{0.32} & 0.61 & 0.17 &  \underline{0.33} & 0.85 & \underline{0.14} & \underline{0.27} & \underline{0.53}& \underline{0.17} & \underline{0.33} & \underline{0.73} & 0.173 & 0.340 & 0.773  \\
    CTKT~\cite{CTKT}  & \underline{0.25} & 0.53 & 1.44 & \textbf{0.11} & \underline{0.26} & 0.67 & \textbf{0.16} & \underline{0.36} & 0.65 & \underline{0.16} & 0.36 & 0.76 & \textbf{0.13} & \underline{0.27} & 0.69 & \underline{0.17} & 0.35 & 0.77 & \underline{0.163} & 0.355 & 0.830\\
    AHMF & \textbf{0.22} & \textbf{0.51} & \underline{1.26} & \underline{0.13} & \textbf{0.25} &\textbf{0.48} & \underline{0.17}& \textbf{0.32} & \underline{0.62} & \textbf{0.15} & \textbf{0.32} & \underline{0.74} & \underline{0.14} & \textbf{0.26} & \textbf{0.52} & \textbf{0.15} & \textbf{0.31} & \textbf{0.62} & \textbf{0.157} & \textbf{0.327} & \textbf{0.706} \\
    \bottomrule
    \end{tabular}
    \vspace{-0.25in}
\end{table*}

\section{Hierarchical Feature Collaboration}
\label{hfc}
It is known to all that in CNN shallower layers encode rich spatial details but lack semantic knowledge, while deeper layers are more effective to capture high-level context and structure information but lose spatial information. Motivated by this observation, we argue that low-level spatial details and high-level structure information should collaborate to boost by each other. Accordingly, we propose a bi-directional hierarchical feature collaboration (BHFC) module to fully leverage the hierarchical fused features. 

Specifically, as shown in Fig.~\ref{fig:bf}, we propose to use bi-directional convolutional gated recurrent units (GRU) for hierarchical feature collaboration by propagating information from one layer to all other layers. BHFC takes the fused multi-modal features $\{\boldsymbol{ F}_f^1, \cdots, \boldsymbol{ F}_f^m \}$ as inputs. Without loss of generality, we take the $i$-th input $\boldsymbol{ F}_f^i$ as an example. It is processed by forward GRU (FGRU) and backward GRU (BGRU), whose results are added together and passed through a residual block \cite{EDSR} for further boosting the hierarchical features. This procedure is formulated as follows:
\begin{equation}
    \boldsymbol{F}_g^i = \mathrm{Res}( \mathrm{FGRU} (\boldsymbol{F}_f^i, \boldsymbol{H}_{lh}^{i-1}) + \mathrm{BGRU} (\boldsymbol{F}_f^{i}, \boldsymbol{H}_{hl}^{m - i}) ),
\end{equation}
where $\text{Res}(\cdot)$ represents a residual block; 
$\boldsymbol{H}_{lh}$ and $\boldsymbol{H}_{hl}$ are the hidden states of FGRU and BGRU respectively. 
The mathematical model of FGRU can be formulated as follows:
\begin{align}
    \boldsymbol{Z}^i_{lh} & = \sigma (\boldsymbol{W}_{z}^i *[\boldsymbol{H}^{i-1}_{lh}, \boldsymbol{F}_f^i] + b_{z}^i),\\
    \boldsymbol{R}^i_{lh} & = \sigma (\boldsymbol{W}_{r}^i * [\boldsymbol{H}^{i-1}_{lh}, \boldsymbol{F}_f^i] + b_{r}^i), \\
    \boldsymbol{\hat{H}}^i_{lh} & = tanh (\boldsymbol{W}_{h}^i * [\boldsymbol{R}^i_{lh} \odot \boldsymbol{H}^{i-1}_{lh}, \boldsymbol{F}_f^i] + b_{h}^i ),  \\
    \boldsymbol{H}^i_{lh} &= \boldsymbol{Z}^i_{lh} \odot \boldsymbol{\hat{H}}^i_{lh}  + (1 - \boldsymbol{Z}^i_{lh}) \odot \boldsymbol{H}^{i-1}_{lh},
\end{align}
where $\boldsymbol{W}$ denotes 3$\times$3 convolutional kernel, the subscripts of which represent different convolutional layers; $b$ is the bias term; $\boldsymbol{Z}^i_{lh} $, $\boldsymbol{R}^i_{lh} $ and $ \boldsymbol{H}^i_{lh} $ represent update gate, reset gate and hidden states, respectively. We set the initial states to zero. $\text{BGRU}(\cdot)$ is with the same formulation as $\text{FGRU}(\cdot)$. $\text{FGRU}(\cdot)$ controls the information flow from low-to-high layers, while $\text{BGRU}(\cdot)$ does it from high-to-low layers.

\section{Experiments}
\label{exp}
\par In this section, we conduct several experiments to evaluate the performance of the proposed method. Firstly, we introduce the datasets and evaluation metrics used in our experiments in subsection~\ref{dataset}. Besides, the experimental settings are presented in subsection~\ref{id}. Moreover, we compare the proposed method with other state-of-the-art depth image SR approaches in subsection~\ref{cs}. Then, ablation studies are presented in subsection~\ref{abs} to analyze the design choices of our method. At last, the network complexity comparisons and discussion on limitations of our method are presented in subsection~\ref{cc} and subsection~\ref{Limitations}, respectively.
\begin{table*}[!tb]\setlength{\tabcolsep}{5.2pt}
    \centering
    \caption{\label{tab:mpi_rmse}RMSE performance comparison for scale factors $4 \times, 8 \times$ and $16 \times$ with bicubic degradation on Middlebury dataset. The best performance is shown in \textbf{bold} and second best performance is the \underline{underscored} ones (Lower MAE values, better performance).}
    \vspace{-0.1in}
    \renewcommand{\arraystretch}{1.2}
    \begin{tabular}{l|ccc|ccc|ccc|ccc|ccc|ccc}
    \toprule
    \multirow{2}{*}{Method}  & \multicolumn{3}{c|}{Art} & \multicolumn{3}{c|}{Books} & \multicolumn{3}{c|}{Dools} & \multicolumn{3}{c|}{Laundry} & \multicolumn{3}{c|}{Mobeius} & \multicolumn{3}{c}{Reindeer} \\
    \cline{2-19}
    ~  &$4\times$ & $8 \times$  & $16\times$  & $4\times$ & $8 \times$  & $16\times$  &$4\times$ & $8 \times$  & $16\times$ &$4 \times$ & $8 \times$  & $16\times$ &$4\times$ & $8 \times$  & $16\times$ &$4\times$ & $8 \times$  & $16\times$   \\
    \hline
    Bicubic & 3.87 & 5.46 & 8.17 & 1.27 & 2.34 & 3.34 & 1.31 & 1.86 & 2.62 & 2.06 & 3.45 & 5.07 & 1.33 & 1.97 & 2.85 & 2.42 & 3.99 & 5.86 \\ 
    DMSG~\cite{DMSG} & 1.47 & 2.46 & 4.57 & 0.67 & 1.03 & 1.60 & \underline{0.69} & \underline{1.05} & \underline{1.60} & \underline{0.79} & 1.51 & 2.63 & 0.66 & 1.02 & 1.63 & 0.98 & 1.76 & 2.92 \\
    DGDIE~\cite{gu2017learning} & 2.00 & 3.84 & 6.16 & 0.91 & 1.68 & 2.67 & 0.84 & 1.54 & 2.34 & 1.30 &3.37 &4.14 & 0.85 &1.86 &2.31 & 1.52 &3.88 &4.30  \\
    DEIN~\cite{DEIN} & 3.26 & 4.20 & 6.40 & 1.38 & 2.12 & 3.38 & 1.20 & 1.64 & 2.27 & 2.00 & 2.59 & 4.07 & 1.20 & 1.75 & 2.97 & 2.27 & 2.95 & 4.21 \\
    DSRNet~\cite{DepthSR} & 1.20 &2.22 &  \textbf{3.90} & 0.60 & 0.89 & \underline{1.51} & 0.84 & 1.14 & \textbf{1.52} & 0.78 & \underline{1.31} & 2.26 & 0.96 & 1.19 & \underline{1.58} & 0.96 & \underline{1.57} & \textbf{2.47}   \\
    DJFR~\cite{DJFR} &1.62 & 3.08 & 5.81 & 0.54 & 1.11 & 2.24 & 0.78 & 1.27 & 2.02 & 0.90 & 1.83 & 3.65 & 0.68 & 1.22 & 2.21 & 1.25 & 2.38 & 4.22   \\ 
    PacNet~\cite{PanNet}& 1.66 & 2.92 & 6.02 & 0.58 & 1.03 & 2.43 & 0.81 & 1.25 & 2.38 & 0.92 & 1.79 & 3.61 & 0.70 & 1.18 & 2.25 & 1.16 & 2.12 & 4.11   \\
    CUNet~\cite{CUNet} & 1.57 & 2.60 & 4.74 & 0.56 & 0.97 & 1.96 & 0.76 & 1.14 & 1.97 & 0.88 & 1.50 & 3.00 & 0.67 & 1.04 & 2.00 & 1.10 & 1.97 & 3.25 \\
    CGN~\cite{CGN} & 1.50 & 2.69 & \underline{4.14} & 0.60 & 0.97 & 1.73 & 0.88 & 1.20 & 1.80 & 0.98 & 1.57 & 2.57 & 0.69 & 1.06 & 1.69 & 1.22 & 2.02 & 3.60   \\
    DSR\_N~\cite{DSR-N} & 1.77 & 3.32 & 6.08 & 0.73 & 1.39 & 2.38 & 0.81 & 1.35 & 2.05 & 0.98 & 2.06 & 4.09 & 0.74 & 1.37 & 2.21 & 1.24 & 2.39 & 4.08 \\
    MDAR~\cite{MDDR} & 2.57 & 3.20 & 4.87 & 1.33 & 1.46 & 2.51 & 1.07  & 1.19 & 1.90 & 2.00 & 2.11 & 4.07 & 0.85 & 1.10 & 1.98 & 1.07 & 1.19 & 3.44  \\ 
    MFR-SR~\cite{mfr-sr} & 1.54 & 2.71 & 4.35 & 0.63 & 1.05 & 1.78 & 0.89 & 1.22 & 1.74 & 1.23 & 2.06 & 3.74 & 0.72 & 1.10 & 1.73 & 1.23 & 2.06 & 3.74 \\
    PMBAN~\cite{PMBANet} & 1.19 & 2.47 & 4.37 & \underline{0.43} & 1.10 & \underline{1.51} & 0.66 & \underline{1.08} & 1.75 & 0.80 & 1.54 & 2.72 & \underline{0.55} & 1.13 & \underline{1.62} & \underline{0.92} & \underline{1.76} &  2.86 \\
    DKN ~\cite{DKN} & \underline{1.12} & \underline{2.46} & 4.61 & 0.44 & \underline{0.82} & 1.71 & 0.71 & 1.14 & 1.75 & \underline{0.79} & 1.46 & \underline{2.23} & 0.59 & \underline{0.97} & 1.68 & \underline{0.92} & 1.83 & 3.30 \\
    AHMF & \textbf{1.09} & \textbf{2.14} & 4.20 & \textbf{0.38} & \textbf{0.72} & \textbf{1.49} & \textbf{0.62} & \textbf{1.03} & \underline{1.66} & \textbf{0.64} & \textbf{1.22} & \textbf{2.14} & \textbf{0.54} & \textbf{0.88} & \textbf{1.53} & \textbf{0.85} & \textbf{1.56} & \underline{2.84} \\
    \bottomrule
    \end{tabular}
    \vspace{-0.25in}
\end{table*}
\subsection{Datasets and Metrics}
\label{dataset}
\textbf{Middlebury Dataset:} Middlebury dataset is a widely used dataset to evaluate the performance of depth super-resolution algorithms. Following~\cite{PMBANet}, we use 36 RGB-D image pairs (6, 21, and 9 pairs from 2001~\cite{2001}, 2006~\cite{middleblur_data_2} and 2014~\cite{2014} datasets, respectively) from Middlebury dataset
to train our model. 6 RGB-D image pairs (\textit{Art, Books, Dools, Laundry, Mobeius, Reindeer}) from Middlebury 2005~\cite{middleblur_data_1} are utilized as testing dataset to evaluate the performance of the proposed method. To evaluate the generalization ability of the proposed method, we select 6 RGB-D image pairs from Lu~\cite{Lu} dataset as another test dataset for our method. For middlebury dataset, we utilize the hole-filled depth maps collected by~\cite{DMSG, 2001}. As suggested by~\cite{2001}, the whole process to fill the depth holes can be divided into three steps: 1) detect the occluded regions by using cross-checking~\cite{fill1, fill2}, 2) apply a median filter to remove spurious mismatches, 3) fill the holes by surface fitting or by distributing neighboring disparity estimates~\cite{fill3, fill4}. We train two kinds of models for two different tasks: (1) depth map super-resolution and (2) joint depth map super-resolution and denoising. Similar to other works~\cite{DepthSR, PMBANet, DSR-N, CTKT}, we generate the low-resolution depth maps by using Bicubic interpolation and employ Mean Absolute Error (MAE) and root mean squared error (RMSE) to evaluate the objective performance of the proposed method.
Following~\cite{DepthSR, PMBANet, DSR-N}, we quantize all recovered depth maps to 8-bits before calculating the MAE or RMSE values for fair evaluation. For both metrics, lower values indicate better performance.

 \textbf{NYU v2 Dataset:} NYU v2 dataset~\cite{NYU} consists of 1449 RGB-D image pairs captured by Microsoft Kinect sensors. Following the similar settings of previous depth map super-resolution methods~\cite{DJFR, DKN}, our method is trained on the first 1000 RGB-D image pairs, and tested on the remained 449 RGB-D images pairs. Following the experimental protocol of Kim \etal~\cite{DKN}, we use the Bicubic and direct down-sampling to generate the low-resolution depth maps, and utilize RMSE as the default metric to evaluate the performance of the proposed method. To show the robustness of the proposed method, we also conduct experiments on depth maps which are captured by different sensors, such as Lu dataset~\cite{Lu} and Middlebury dataset~\cite{middleblur_data_1}. Since the acquired depth maps usually have missing values, the depth maps are in-painted by the official toolbox\footnote{http://cs.nyu.edu/~silberman/code/toolbox\_nyu\_depth\_v2.zip} which employs the colorization framework~\cite{levin2004colorization} to fill the missing values. 

\subsection{Implementation Details}
\label{id}
Our model has only one specific hyper-parameter $m$, which is used to control the number of convolutional layers for multi-modal feature extraction. To balance the efficiency and network performance, we set $m=4$ as default. We set the channel number of all intermediate layers as 64. The ablation experiments presented below will verify the effectiveness of our configurations. The kernel size of a convolutional layer is set as 3$\times$3, except for those in the upsampling and downsampling module that are set as 1$\times$1, 3$\times$3 and 5$\times$5 with a stride size of 0, 1, 2 in 4$\times$, 8$\times$, 16$\times$ super-resolution, respectively.  We use PReLU \cite{PReLU} as the default activation function. During training, we randomly select 32 HR depth maps with the size of 256$\times$256 as the ground truth, and the LR depth maps are generated by using the down-sampling operator. Our model is optimized by Adam~\cite{adam:} with $\beta_{1}=0.9, \beta_{2}=0.999,$ and $\epsilon=1 \mathrm{e}-8$; the initial learning rate is $2 \times 10^{-4}$ and decreased by multiplying 0.5 for every 100 epochs. It takes about 25 hours to train our model. The proposed model is implemented by PyTorch \cite{PyTorch:} and trained on a NVIDIA V100 GPU. 
\begin{figure*}[!tb]
    \centering
    \includegraphics[width=\linewidth]{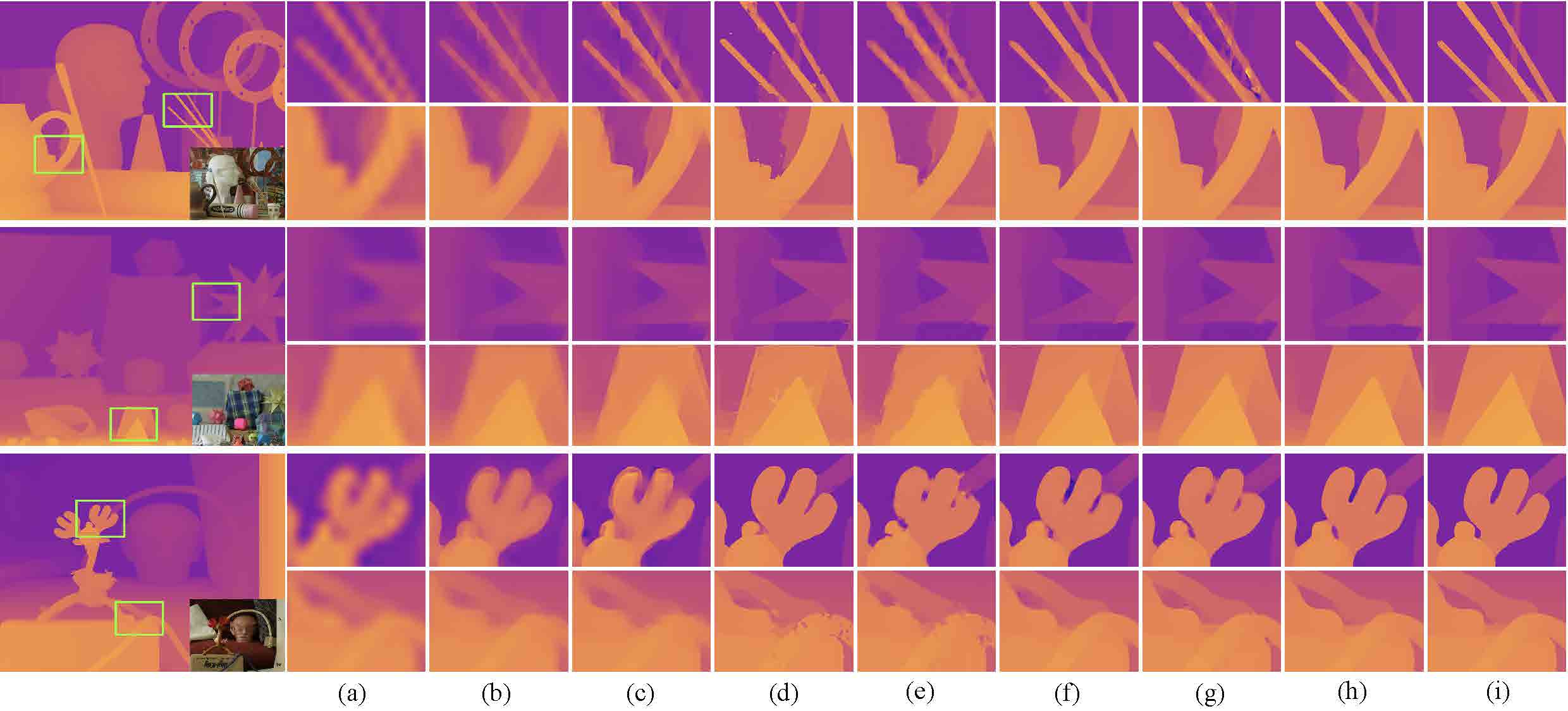}
    \vspace{-0.3in}
    \caption{Visual comparison of $16 \times$ upsampling results on \textit{Art}, \textit{Moebius} and \textit{Reindeer} from Middlebury dataset~\cite{middleblur_data_1}: (a): Bicubic, (b): CLMF1~\cite{clmf}, (c): DGDIE~\cite{gu2017learning}, (d): DEIN~\cite{DEIN}, (e): DJFR~\cite{DJFR}, (f): PMBAN~\cite{PMBANet}, (g): DKN~\cite{DKN}, (h): Ours, (i): GT. Please enlarge the PDF for more details.}
    \label{fig:mpi_vis}
    \vspace{-0.15in}
\end{figure*}
\begin{table*}[!tb]\setlength{\tabcolsep}{5.2pt}
    \centering
    \caption{\label{tab:mpi_noisy}MAE comparison for scale factors $4 \times, 8 \times$ and $16 \times$ with ToF-like degradation on Middlebury dataset. The best performance is shown in \textbf{bold} and the second best performance is the \underline{underscored} ones (Lower MAE values, better performance).}
    \vspace{-0.1in}
    \renewcommand{\arraystretch}{1.2}
    \begin{tabular}{l|ccc|ccc|ccc|ccc|ccc|ccc}
    \toprule
    \multirow{2}{*}{Method}  & \multicolumn{3}{c|}{Art} & \multicolumn{3}{c|}{Books} & \multicolumn{3}{c|}{Dools} & \multicolumn{3}{c|}{Laundry} & \multicolumn{3}{c|}{Mobeius} & \multicolumn{3}{c}{Reindeer}\\
    \cline{2-19}
    ~ & $4\times$ & $8 \times$  & $16\times$  & $4\times$ & $8 \times$  & $16\times$  &$4\times$ & $8 \times$  & $16\times$ &$4 \times$ & $8 \times$  & $16\times$ &$4\times$ & $8 \times$  & $16\times$ &$4\times$ & $8 \times$  & $16\times$ \\
    \hline
    Bicubic & 3.23 & 4.00 & 5.58 &2.62 & 2.83 & 3.27 &2.61&2.75&3.10&2.79&3.96&4.14&2.63&2.84&3.21&2.84&3.18&4.05 \\   
    GF \cite{GF} &1.91&2.87&4.90&1.13&1.84&2.80&1.14&1.86&2.69&1.31&2.12&3.39&1.17&1.87&2.77&1.33&2.10&3.56 \\
    DMSG \cite{DMSG} & 0.84 & 1.57 & 2.98 & 0.62 & 1.18 & 1.48 & 0.84 & 1.12 & 1.78 & 0.78 & 1.03 & 1.89 & 0.66 & 1.13 & 1.76 & 0.57 & 1.12 & 1.87 \\
    DGDIE~\cite{gu2017learning} & 0.99 & 1.84 & 3.34 & 0.81 & 1.29 & 2.04 & 0.95 & 1.39 & 2.05 & 1.10 & 1.73 & 2.67 & 0.84 & 1.37 & 2.16 & 0.79 & 1.33 & 2.19 \\
    DSRNet~\cite{DepthSR}&2.19&2.57&3.69&1.81&1.82&2.16&1.84&1.91&2.21&1.93&2.07&2.66&1.81&1.84&2.26&1.95&2.09&2.59 \\
    GSPRT ~\cite{lutio2019guided} & 0.68 & 1.33 & 2.47 & 0.52 & 0.87 & 1.37 & 0.78 & 1.26 & 2.03 & 0.76 & 1.24 & 1.86 & 0.65 & 1.03 & 1.68 & 0.55 & 1.04 & 1.70 \\
    DJFR~\cite{DJFR}&0.82 & 1.44 & 2.77 & 0.62 & 1.03 & 1.67 & 0.73 & 1.12 & 1.66 & 0.72 & 1.23 & 2.02 & 0.66 & 1.05 & 1.71 & 0.65 & 1.11 & 1.88  \\ 
    PacNet~\cite{PanNet}& 0.72 & 1.40 & 2.45 & 0.49 & 0.83 & 1.47 & 0.71 & 1.15 & 1.64 & 0.64 & 1.22 & 1.92 & 0.56 & 0.93 & 1.86 & 0.57 & 1.03 & 1.38\\
    CUNet~\cite{CUNet} & 0.81 & 1.25 & 2.35 & 0.54 & 0.85 & 1.47 & 0.74 & 1.07 & 1.78 & 0.66 & 1.05 & 2.00 & 0.60 & 0.93 & 1.64 & 0.61 & 0.96 & 1.73\\
    PMBAN~\cite{PMBANet}& \textbf{0.59} & 0.98 & 1.89 & \underline{0.44} & \textbf{0.71} & 1.23 & \textbf{0.64} & 1.01 & 1.56 & \underline{0.54} & \underline{0.89} & \textbf{1.62} & \underline{0.48} & 0.81 & \underline{1.30} & \textbf{0.47} & \underline{0.78} & 1.52 \\
    DKN ~\cite{DKN} & 0.68 & \underline{0.95} & \underline{1.86} & 0.51 & \underline{0.80} & \underline{1.21} & \underline{0.67} & \underline{0.99} & \textbf{1.46} & 0.62 & 1.00 & 1.72 & 0.55 & \underline{0.80} & 1.37 & \underline{0.54} & \underline{0.78} & \underline{1.50} \\
    AHMF (Ours)  & \underline{0.62} & \textbf{0.93} & \textbf{1.85} & \textbf{0.41} & \textbf{0.71} & \textbf{1.19} & 0.68 & \textbf{0.94} & \underline{1.47} & \textbf{0.52} & \textbf{0.88} & \underline{1.63} & \textbf{0.46} & \textbf{0.79} & \textbf{1.28} & \textbf{0.47} & \textbf{0.76} & \textbf{1.48}  \\
    \bottomrule
    \end{tabular}
    \vspace{-0.16in}
\end{table*}

\subsection{Comparison with the State-of-the-arts}
\label{cs}
\subsubsection{Experimental Results on Middlebury Dataset (Noiseless Case)}
\begin{table*}[!tb]\setlength{\tabcolsep}{5.2pt}
    \centering
    \caption{\label{tab:mpi_noisy_rmse} RMSE performance comparison for scale factors $4 \times, 8 \times$ and $16 \times$ with ToF-like degradation on Middlebury dataset. The best performance is shown in \textbf{bold} and second best performance is the \underline{underscored} ones (Lower MAE values, better performance).}
    \vspace{-0.1in}
    \renewcommand{\arraystretch}{1.2}
    \begin{tabular}{l|ccc|ccc|ccc|ccc|ccc|ccc}
    \toprule
    \multirow{2}{*}{Method}  & \multicolumn{3}{c|}{Art} & \multicolumn{3}{c|}{Books} & \multicolumn{3}{c|}{Dools} & \multicolumn{3}{c|}{Laundry} & \multicolumn{3}{c|}{Mobeius}& \multicolumn{3}{c}{Reindeer} \\
    \cline{2-19}
    ~ & $4\times$ & $8 \times$  & $16\times$  & $4\times$ & $8 \times$  & $16\times$  &$4\times$ & $8 \times$  & $16\times$ &$4 \times$ & $8 \times$  & $16\times$ &$4\times$ & $8 \times$  & $16\times$ &$4\times$ & $8 \times$  & $16\times$ \\
    \hline
    Bicubic & 5.74 & 6.89 &  9.12 & 4.57 & 4.86 & 5.28 & 4.50 & 4.69 & 5.02 & 4.91 & 5.47 & 6.48 & 4.48 & 4.72 & 5.23 & 5.11 & 5.80 & 7.12 \\
    DMSG \cite{DMSG} & 3.06 & 4.48 & 6.44 & 1.90 & 3.13 & 4.63 & 1.99 & 3.17 & 4.51 & 2.25 & 3.60 & 5.39 & 1.97 & 3.22 & 4.50 & 2.32 & 3.75 & 5.35  \\
    DGDIE~\cite{gu2017learning} & 4.29 &5.94 &9.78 &1.99 &2.65 &4.16 &1.74 &2.28 &3.18 &3.00 &4.21 &6.80 &1.69 &2.37 &3.54 &3.40 &4.47 &7.86 \\
    DSRNet~\cite{DepthSR} & -- & -- & 6.24 & -- & -- & 3.36 & -- & -- & -- & -- & -- & 4.95 & -- & -- & -- & -- & -- & 4.65  \\
    DJFR~\cite{DJFR}& 2.96 & 4.81 & 7.30 & 1.41 & 2.58 & 3.47 & 1.91 & 2.57 & 4.33 & 2.55 & 3.07 & 5.43 & 2.43 & 2.62 & 4.35 & 2.60 & 3.21 & 5.66  \\ 
    PacNet~\cite{PanNet}& 2.95 & 4.41 & 7.80 & 1.37 & 2.05 & 3.80 & 1.76 & 2.45 & 3.87 & 2.01 & 3.09 & 4.99 & 1.61 & 2.26 & 3.84 & 2.07 & 3.24 & 5.79  \\
    CUNet~\cite{CUNet} & 2.85 & \underline{3.84} & 6.12 & 1.42 & 1.95 & 2.95 & 1.71 & 2.24 & \underline{2.93} & 1.93 & 2.82 & 4.23 & 1.55 & 2.08 & \textbf{2.98} & 2.01 & 3.01 & \underline{4.20} \\
    PMBAN~\cite{PMBANet} & \textbf{2.58} & 3.92 & 6.67 & \underline{1.25} & \underline{1.82} & 2.94 & \underline{1.52} & 2.18 & 3.10 & \underline{1.69} & 2.71 & \textbf{3.67} & \underline{1.38} & \textbf{1.93} & 3.08 & \underline{1.81} & \textbf{2.72} & 4.58\\
    DKN~\cite{DKN} & \underline{2.77} & \textbf{3.79} & \underline{6.01} & 1.36 & 1.91 & \underline{2.93} & 1.68 & \underline{2.16} & \textbf{2.85} & 1.89 & \underline{2.69} & 4.15 & 1.49 & 2.10 & \underline{2.99} & 1.93 & 2.99 & 4.24 \\
    AHMF (Ours)  & \textbf{2.58} & 3.85 & \textbf{6.00} & \textbf{1.21} & \textbf{1.80} & \textbf{2.75} & \textbf{1.47} & \textbf{2.04} & \textbf{2.85} & \textbf{1.63} & \textbf{2.45} & \underline{3.79} & \textbf{1.33} & \underline{2.03} & \textbf{2.98} & \textbf{1.78} & \underline{2.80} &\textbf{4.18} \\
    \bottomrule
    \end{tabular}
    \vspace{-0.25in}
\end{table*}
\begin{figure}[!tb]
	\begin{center}
		\subfigure[Guidance]{
		\hspace{-0.1in}
		\begin{minipage}[b]{0.24\linewidth}
			\includegraphics[width=1\linewidth]{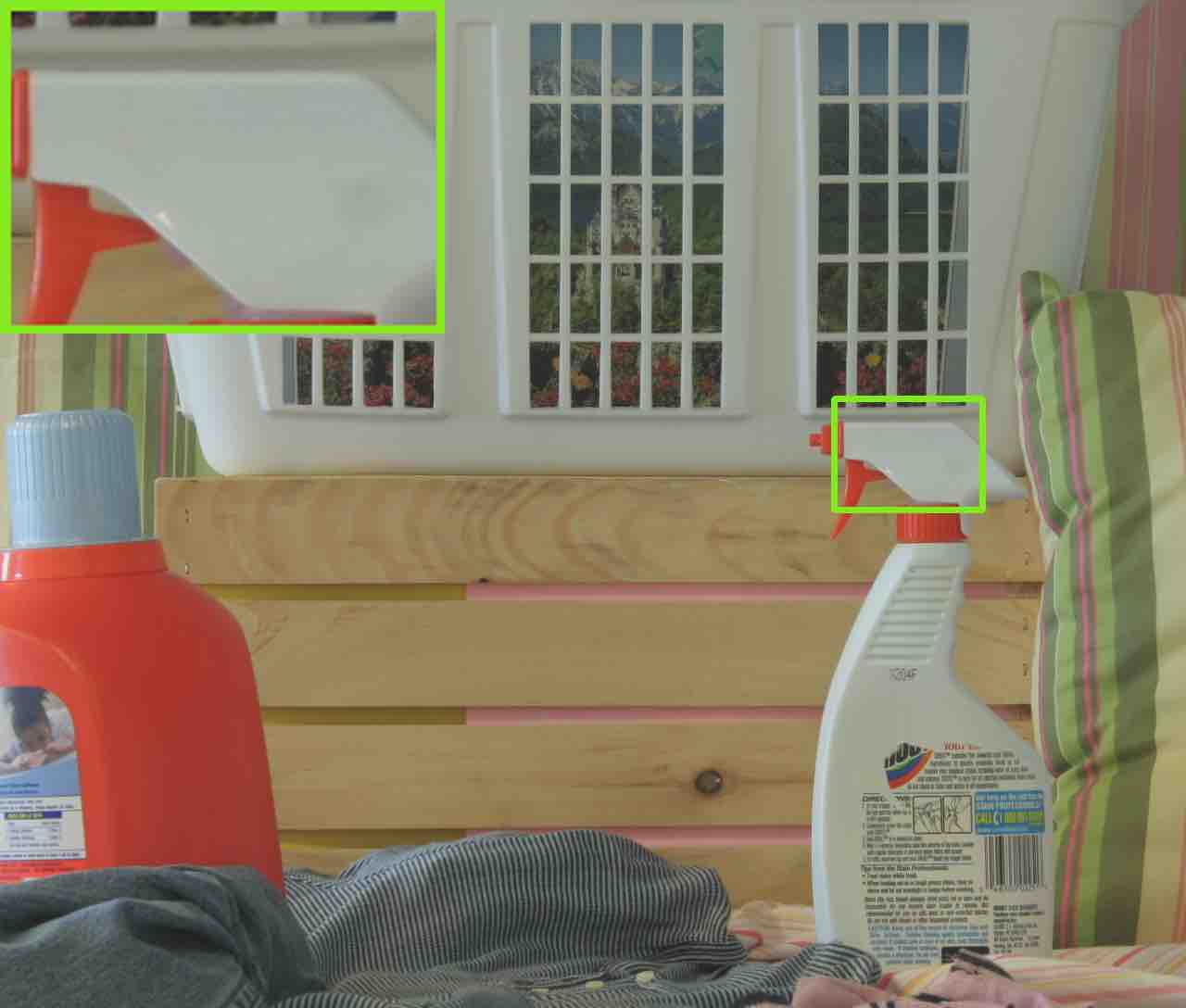} 
		\end{minipage}
		\hspace{-0.1in}
	}\subfigure[Bicubic]{
    		\begin{minipage}[b]{0.24\linewidth}
   		 	\includegraphics[width=1\linewidth]{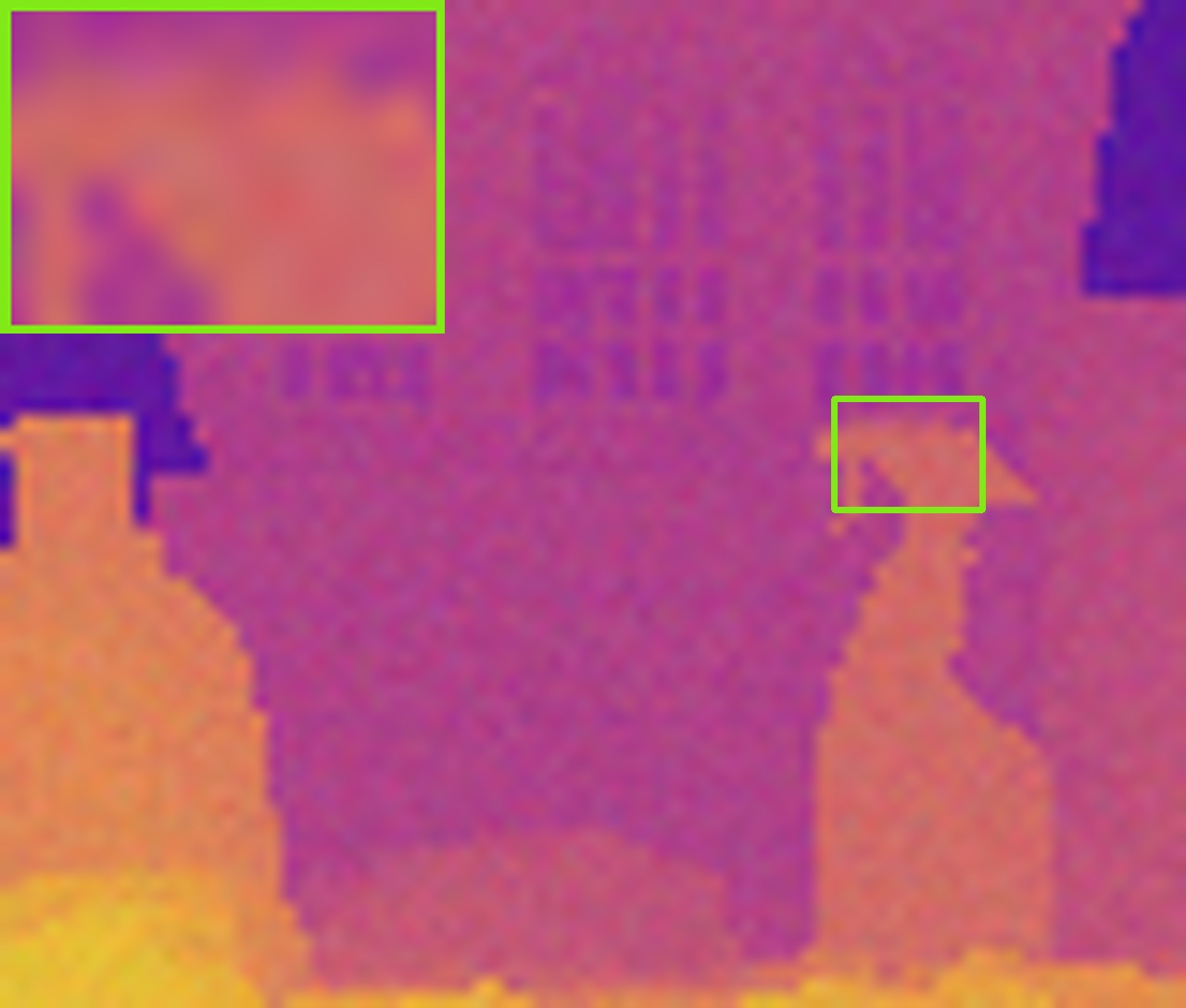}
    		\end{minipage}
    		\hspace{-0.1in}
    	}\subfigure[DJFR~\cite{DJFR}]{
    		\begin{minipage}[b]{0.24\linewidth}
   		 	\includegraphics[width=1\linewidth]{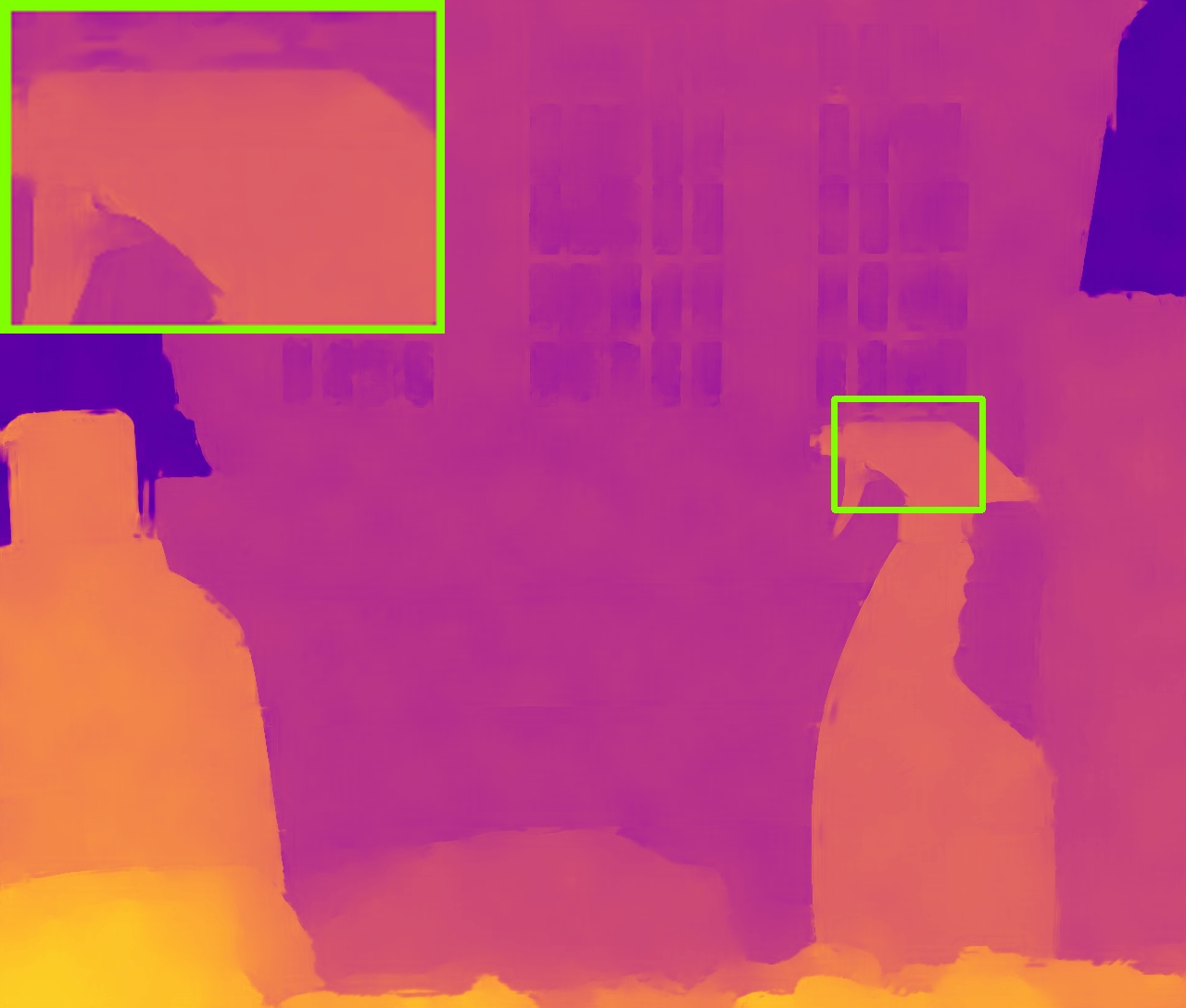}
    		\end{minipage}
    		\hspace{-0.1in}
    	}\subfigure[PacNet~\cite{PanNet}]{
		\begin{minipage}[b]{0.24\linewidth}
			\includegraphics[width=1\linewidth]{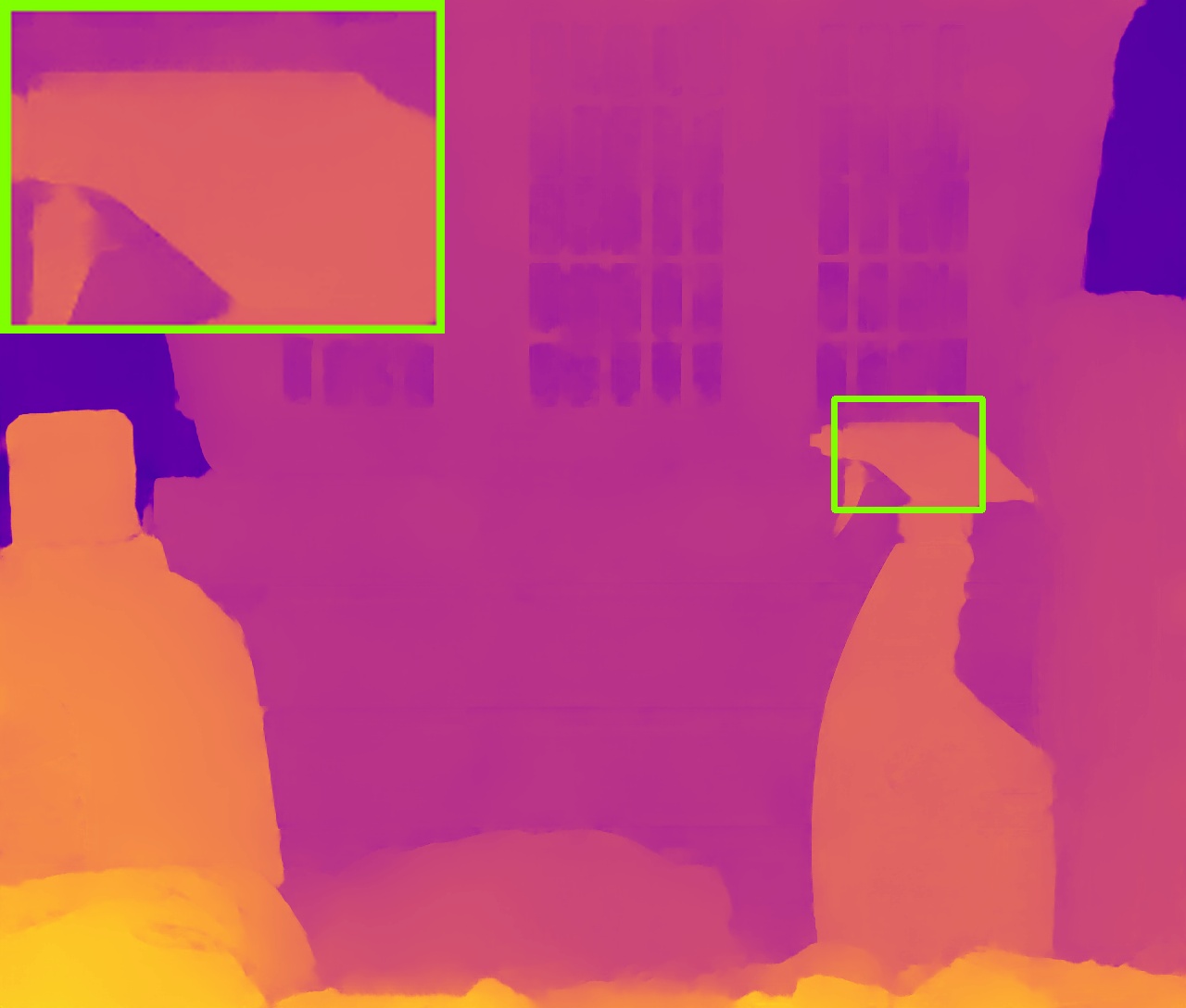} 
		\end{minipage}
	}
	\subfigure[CUNet~\cite{CUNet}]{
	        \hspace{-0.1in}
    		\begin{minipage}[b]{0.24\linewidth}
		 	\includegraphics[width=1\linewidth]{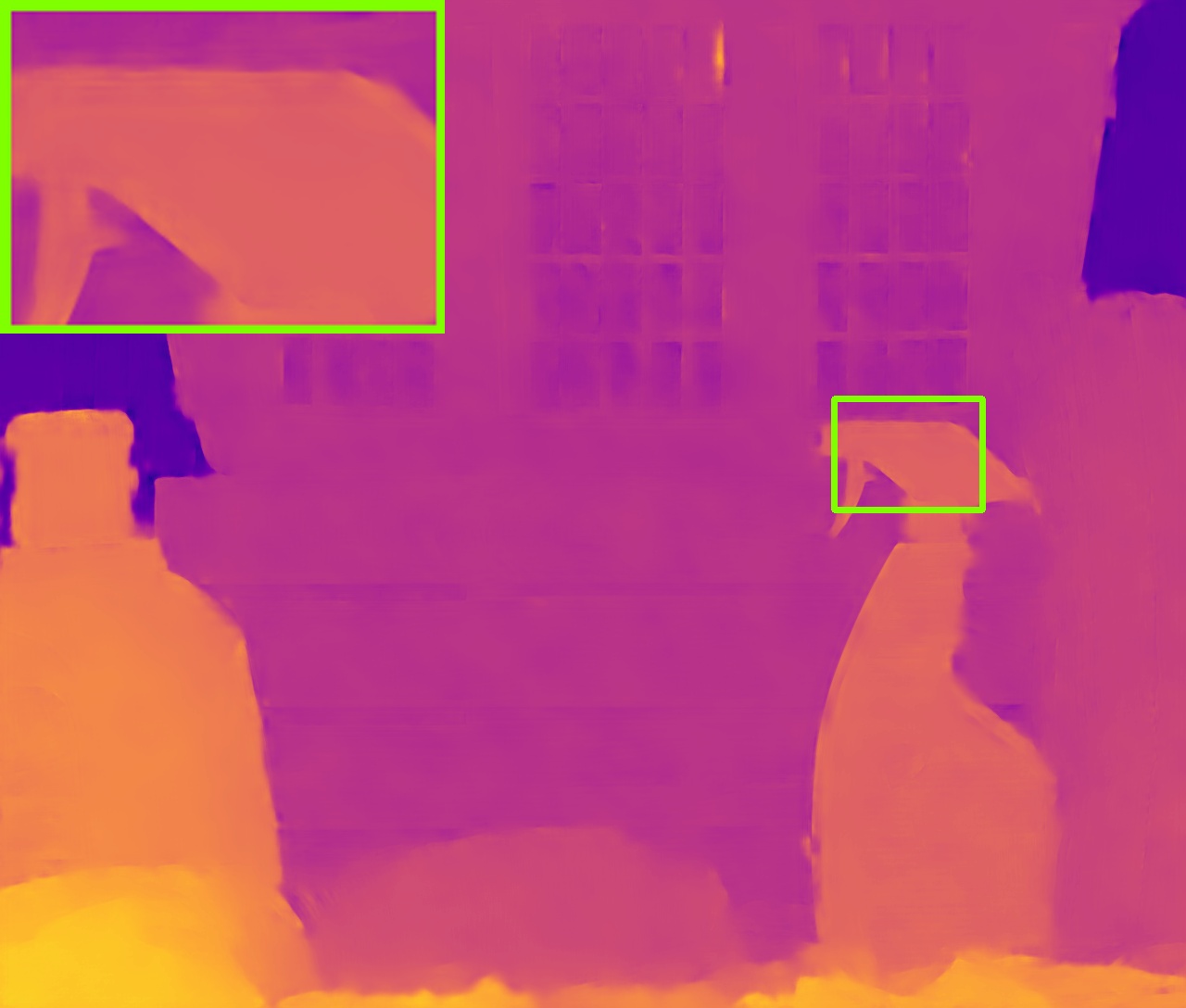}
    		\end{minipage}
    		\hspace{-0.1in}
    	}\subfigure[DKN~\cite{DKN}]{
    		\begin{minipage}[b]{0.24\linewidth}
   		 	\includegraphics[width=1\linewidth]{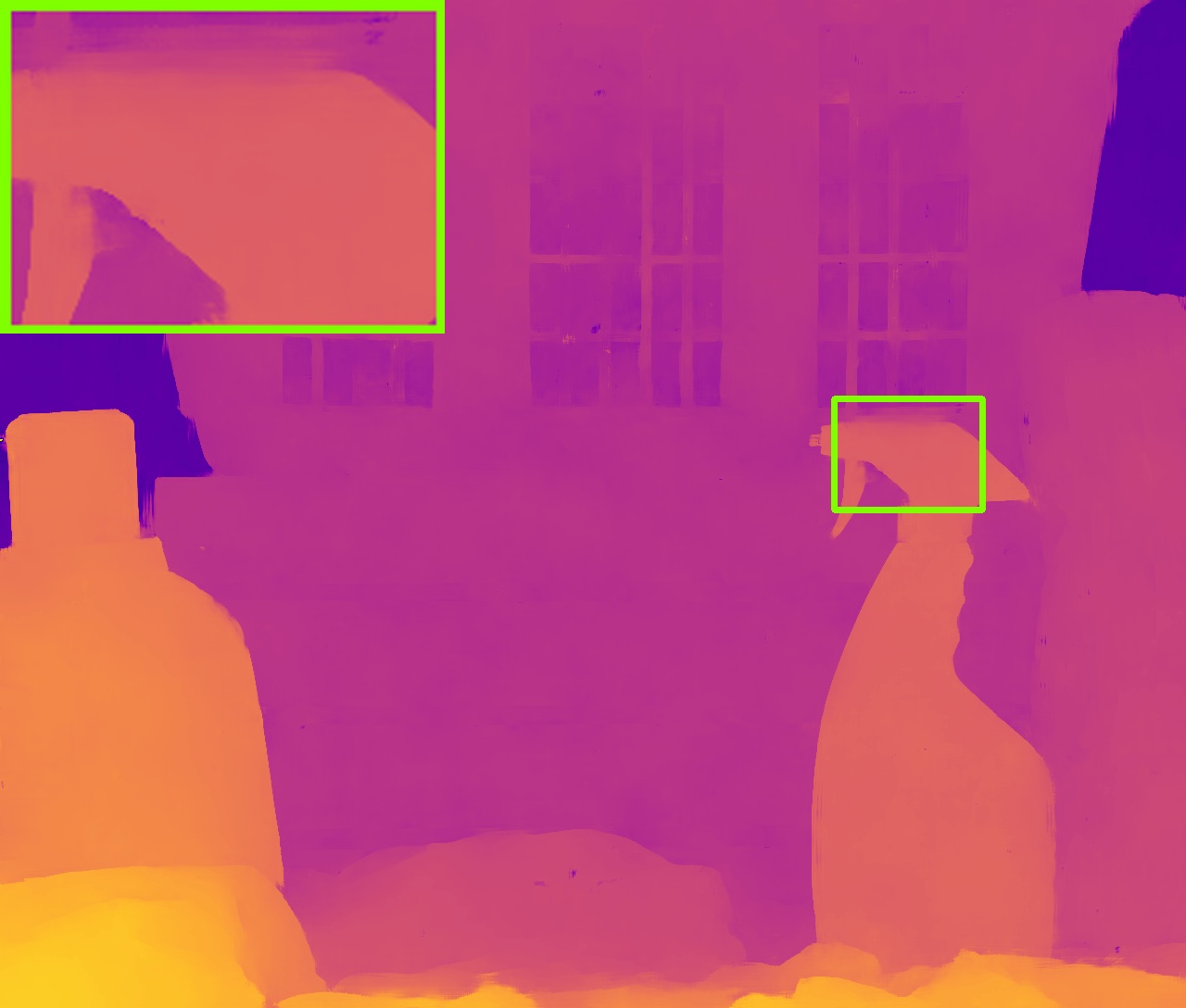}
    		\end{minipage}
    		\hspace{-0.1in}
    	}\subfigure[Ours]{
    		\begin{minipage}[b]{0.24\linewidth}
   		 	\includegraphics[width=1\linewidth]{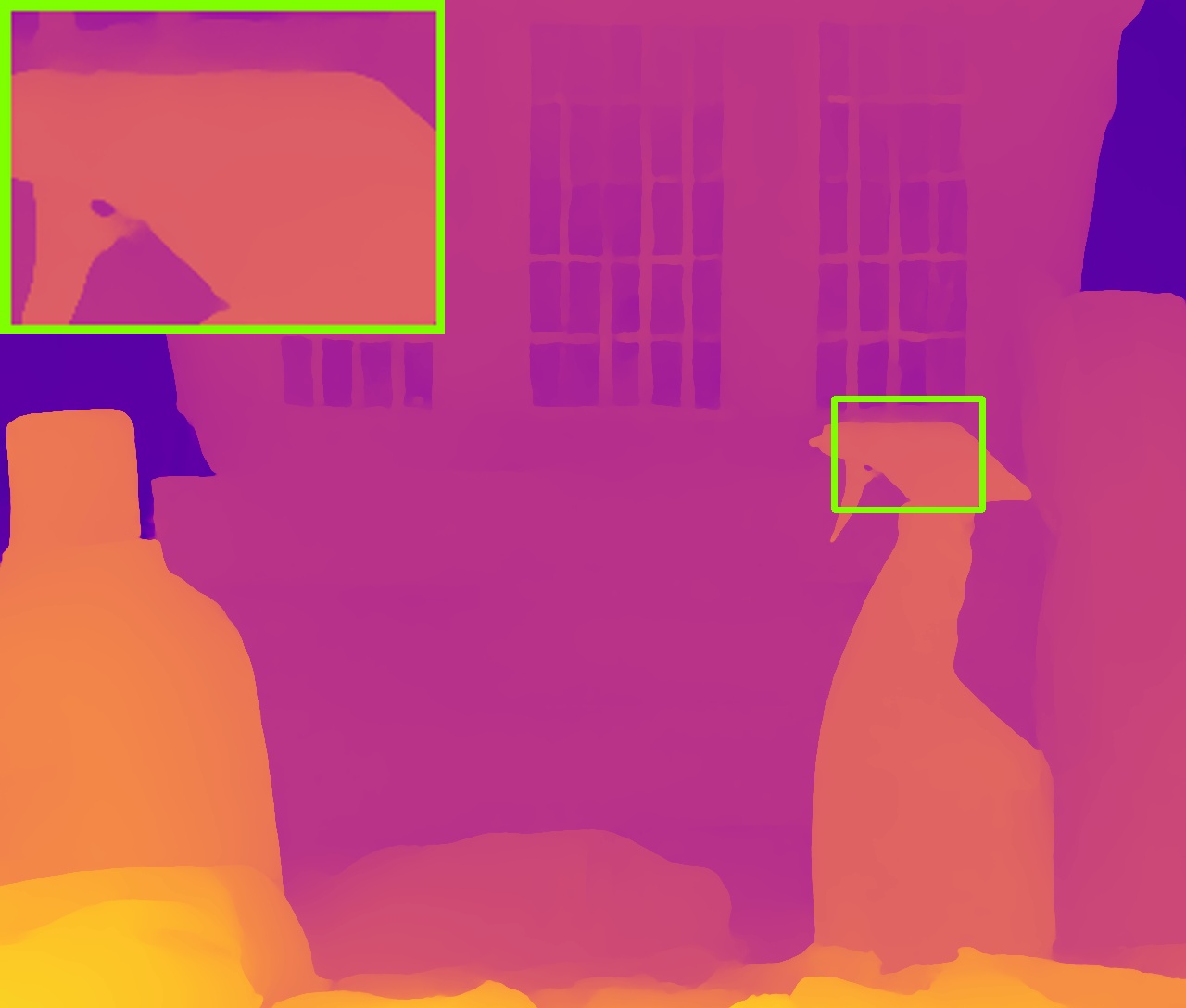}
    		\end{minipage}
    		\hspace{-0.1in}
    	}\subfigure[Ground Truth]{
    		\begin{minipage}[b]{0.24\linewidth}
   		 	\includegraphics[width=1\linewidth]{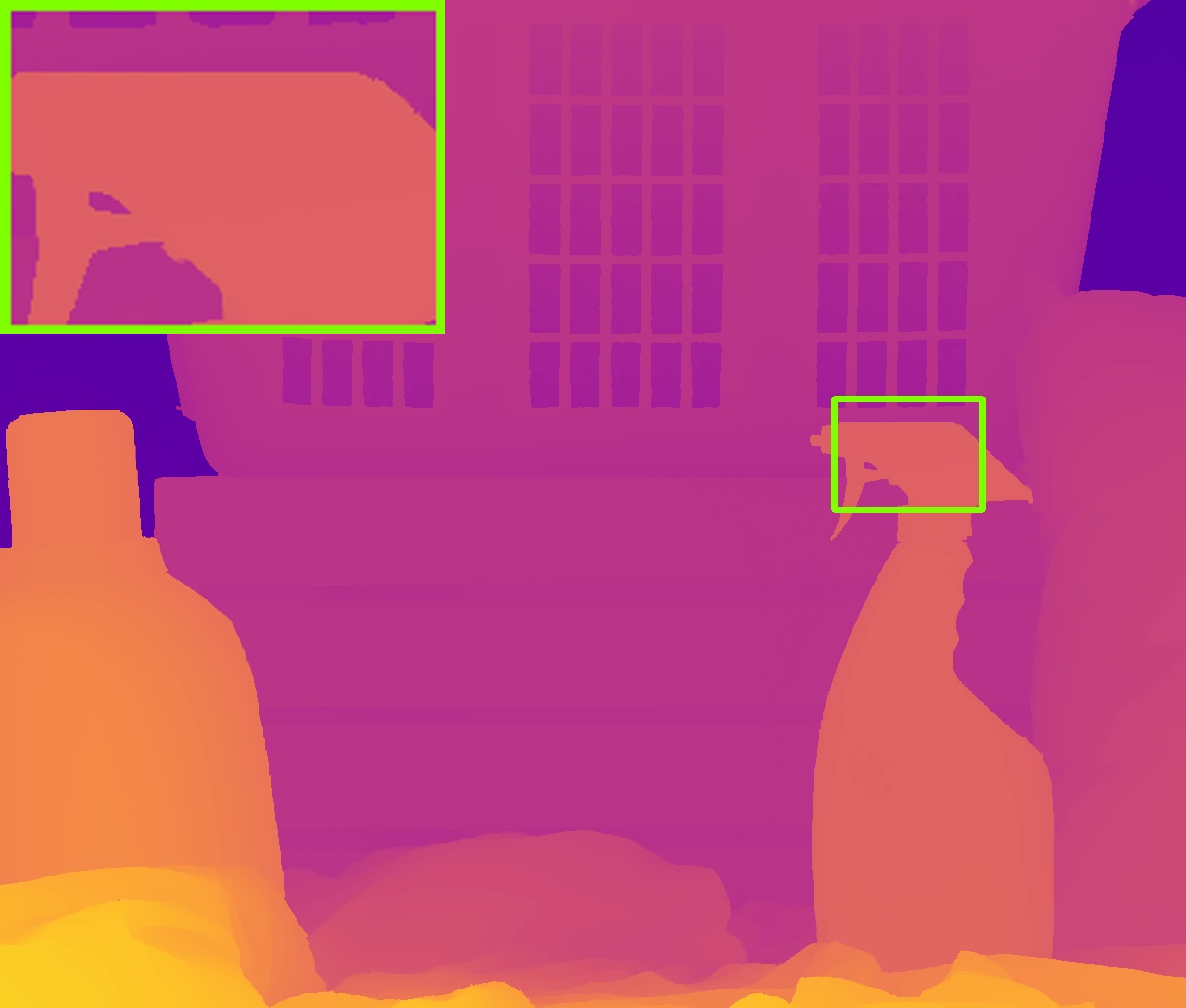}
    		\end{minipage}
    	}
	\end{center}
	\vspace{-0.15 in}
	\caption{Visual comparison of $16\times$ results on \textit{Laundry} from Middlebury~\cite{middleblur_data_1} dataset. Please enlarge the PDF for more details.}
	\label{fig:mpi_noisy_vis}
	\vspace{-0.25in}
\end{figure}
\begin{table*}[!tb]\setlength{\tabcolsep}{5.2pt}
	\centering
	\renewcommand{\arraystretch}{1.2}
    \caption{\label{tab:Lu_result}MAE values for scale factors $4 \times, 8 \times$ and $16 \times$ with bicubic degradation on Lu dataset. The best performance is shown in \textbf{bold} and second best performance is the \underline{underscored} ones (Lower MAE values, better performance).}
    \vspace{-0.1in}
		\begin{tabular}{l|ccc|ccc|ccc|ccc|ccc|ccc}
		\toprule
		\multirow{2}{*}{Method}  & \multicolumn{3}{c|}{Image\_01} & \multicolumn{3}{c|}{Image\_02} & \multicolumn{3}{c|}{Image\_03} & \multicolumn{3}{c|}{Image\_04} & \multicolumn{3}{c|}{Image\_05} & \multicolumn{3}{c}{Image\_06} \\
    \cline{2-19}
     ~& $4\times$ & $8 \times$  & $16\times$  & $4\times$ & $8 \times$  & $16\times$  &$4\times$ & $8 \times$  & $16\times$ &$4 \times$ & $8 \times$  & $16\times$ &$4\times$ & $8 \times$  & $16\times$ &$4\times$ & $8 \times$  & $16\times$ \\
    \hline
    Bicubic & 0.57 & 1.32 & 2.46 & 0.82  & 2.06 & 4.34 & 0.40 & 0.93 & 1.85 & 0.73  & 1.73 & 3.44 & 0.56 & 1.47 & 3.06 & 0.52 & 1.24 & 2.58  \\
    DSRNet~\cite{DepthSR}& 0.57& 1.01& 1.89& 0.82& 1.48& 3.08& 0.40 & 0.83& 1.56& 0.73& 1.21& 2.52& 0.56& 1.10 & 2.12& 0.52& 0.85& 1.85 \\
    DEDIE~\cite{gu2017learning} & 0.53 &1.19 &2.11 &0.83 &1.81 &3.57 &0.45 &1.02 &1.73 &0.56 &1.61 &2.81 &0.55 &1.30 &2.39 &0.48 &1.19 &2.19 \\
    CUNet~\cite{CUNet}& 0.57& 1.01& 1.89& 0.82& 1.48& 3.08& 0.4& 0.83& 1.56& 0.73& 1.21& 2.52& 0.56& 1.10 & 2.12& 0.52& 0.85& 1.85 \\
    PacNet~\cite{PanNet}& 0.61 & 1.34 & 2.77 & 0.84 & 1.65 & 3.52 & 0.52 & 1.14 & 2.22 & 0.61 & 1.32 & 2.55 & 0.57 & 1.12 & 2.38 & 0.52 & 1.03 & 2.15 \\
    DJFR~\cite{DJFR}& 0.53 & 1.11 & 2.35 & 0.74 & 1.47 & 3.08 & 0.41 & 0.86 & 1.80 & 0.55 & 1.09 & 2.36 & 0.53 & 0.93 & 1.94 & 0.43 & 0.75 & 1.53\\
    PMBAN~\cite{PMBANet}&  0.54 & \textbf{0.79} & 1.83 & 0.74 &  \underline{1.17} & 2.83 & 0.42 & \underline{0.72} & \underline{1.36} & 0.57 & \underline{0.84} & \textbf{1.86} & 0.55 & \textbf{0.74} & \underline{1.61} & 0.49 & \underline{0.64} & \underline{1.13} \\
    DKN~\cite{DKN} & \underline{0.48} & \underline{0.80} &  \underline{1.80} &  \underline{0.69} & \textbf{1.16} &  \underline{2.86} &  \underline{0.38} &  0.73 & 1.75 &  \underline{0.50} &  0.92 &  2.34 &  \underline{0.50} &  0.78 &  1.84 &  \underline{0.40} &  \underline{0.64} &  1.26 \\
    AHMF (Ours) & \textbf{0.42} & 0.82 & \textbf{1.76} & \textbf{0.64} & 1.19 & \textbf{2.78} &\textbf{0.36} & \textbf{0.67} & \textbf{1.32}& \textbf{0.45} &\textbf{0.79} & \underline{1.90} & \textbf{0.46} &\underline{0.76} & \textbf{1.53} & \textbf{0.39} & \textbf{0.61} & \textbf{1.07} \\
	\bottomrule
	\end{tabular}
	\vspace{-0.2in}
\end{table*}
\begin{table*}[!tb]\setlength{\tabcolsep}{5.2pt}
	\centering
	\renewcommand{\arraystretch}{1.2}
    \caption{\label{tab:Lu_result_rmse}RMSE values for scale factors $4 \times, 8 \times$ and $16 \times$ with bicubic degradation on Lu dataset. The best performance is shown in \textbf{bold} and second best performance is the \underline{underscored} ones (Lower RMSE values, better performance).}
        \vspace{-0.1in}
		\begin{tabular}{l|ccc|ccc|ccc|ccc|ccc|ccc}
		\toprule
		\multirow{2}{*}{Method}  & \multicolumn{3}{c|}{Image\_01} & \multicolumn{3}{c|}{Image\_02} & \multicolumn{3}{c|}{Image\_03} & \multicolumn{3}{c|}{Image\_04} & \multicolumn{3}{c|}{Image\_05} & \multicolumn{3}{c}{Image\_06} \\
    \cline{2-19}
     ~& $4\times$ & $8 \times$  & $16\times$  & $4\times$ & $8 \times$  & $16\times$  &$4\times$ & $8 \times$  & $16\times$ &$4 \times$ & $8 \times$  & $16\times$ &$4\times$ & $8 \times$  & $16\times$ &$4\times$ & $8 \times$  & $16\times$  \\
    \hline
    Bicubic& 2.18 & 3.93 & 6.10 & 3.55 & 6.72 & 11.19 & 1.36 & 2.42 & 4.18 & 2.71 & 4.76 & 7.55 & 2.39 & 4.92 & 7.88 & 2.32 & 4.53 & 7.37 \\
    DGDIE~\cite{gu2017learning} & 1.90 &3.33 &5.50 &3.33 &6.09 &10.41 &1.21 &2.46 &4.20 &1.81 &4.69 &7.23 &2.17 &4.20 &6.57 &2.03 &4.34 &7.01 \\
    DSRNet~\cite{DepthSR} & 1.47 & \underline{2.34} & 5.25 &  \textbf{2.01} & \textbf{3.45}  & 8.18 & 0.92 & \textbf{1.60}  & 3.36 &  1.53 & 2.28  & 4.92 &  1.65 & \textbf{2.29}  & 3.65 &  \underline{1.39}  & \textbf{1.83} & \underline{2.97}\\
    CUNet~\cite{CUNet}& 1.55 & 2.47 & \underline{5.23} & 3.08 & 4.04 & 7.98 & 1.15 & 2.11 & 3.95 & 1.73 & 2.17 & 4.93 & 2.31 & 2.80 & 3.98 & 2.12 & 2.43 & 3.35 \\
    PacNet~\cite{PanNet}& 1.81 & 2.83 & 5.66 & 3.54 & 4.29 & 8.69 & 1.38 & 2.08 & 4.42 & 1.98 & 2.35 & 5.27 & 2.63 & 2.79 & 4.53 & 2.46 & 2.37 & 3.79\\
    DJFR~\cite{DJFR}& 2.10 & 2.98 & 7.46 & 3.53 & 4.69 & 10.60 & 1.92 & 2.14 & 5.81 & 3.46 & 2.97 & 9.21 & 2.96 & 2.99 & 7.20 & 2.48 & 2.65 & 5.34\\
    PMBAN~\cite{PMBANet} & \underline{1.35} & 2.51 & 5.35 & {2.37} & 3.98 & \underline{7.60} & \underline{0.85} & 1.87 & \underline{3.04} & 1.37 & \textbf{2.12} & 5.07 & 1.66 & 2.74 & 3.58 & 1.58 & 2.39 & 3.03\\
    DKN~\cite{DKN} & 1.37 & 2.67 & 5.46 & 2.65 & 4.38 & 7.80 & 0.87 & 1.82 & 4.24 & \underline{1.36} & 2.46 & \underline{4.75} & \underline{1.44} & 2.94 & \underline{3.56} & 1.57 & 2.57 & 3.70 \\
    AHMF (Ours) & \textbf{1.23} & \textbf{2.22} & \textbf{4.92} & \underline{2.21} & \underline{3.97} & \textbf{7.47} & \textbf{0.72} & \underline{1.65} & \textbf{3.03} & \textbf{1.11} & \underline{2.14} & \textbf{4.63} & \textbf{1.36} & \underline{2.69} & \textbf{3.51} & \textbf{1.34} & \underline{2.36} & \textbf{2.87} \\
	\bottomrule
	\end{tabular}
	\vspace{-0.2in}
\end{table*}
\begin{figure}[!tb]
	\begin{center}
		\subfigure[Bicubic]{
		\hspace{-0.1in}
		\begin{minipage}[b]{0.24\linewidth}
			\includegraphics[width=1\linewidth]{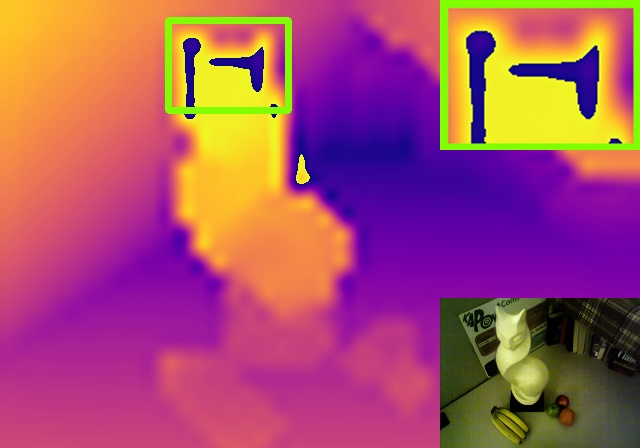} 
		\end{minipage}
		\hspace{-0.11in}
	}\subfigure[DGDIE~\cite{GF}]{
    		\begin{minipage}[b]{0.24\linewidth}
   		 	\includegraphics[width=1\linewidth]{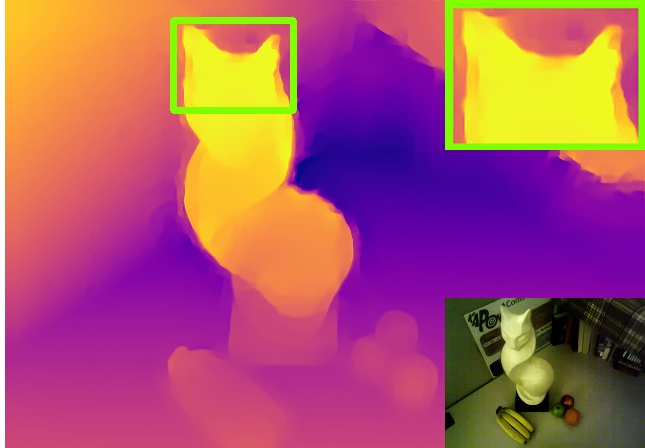}
    		\end{minipage}
    		\hspace{-0.1in}
    	}\subfigure[CUNet~\cite{CUNet}]{
		\begin{minipage}[b]{0.24\linewidth}
			\includegraphics[width=1\linewidth]{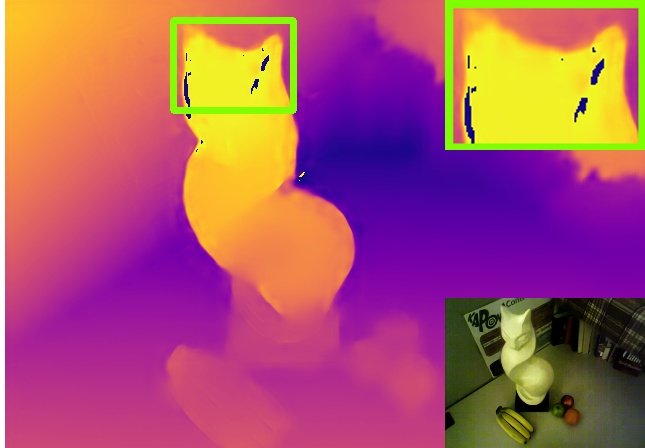} 
		\end{minipage}
		\hspace{-0.1in}
	}\subfigure[PacNet~\cite{DJFR}]{
    		\begin{minipage}[b]{0.24\linewidth}
		 	\includegraphics[width=1\linewidth]{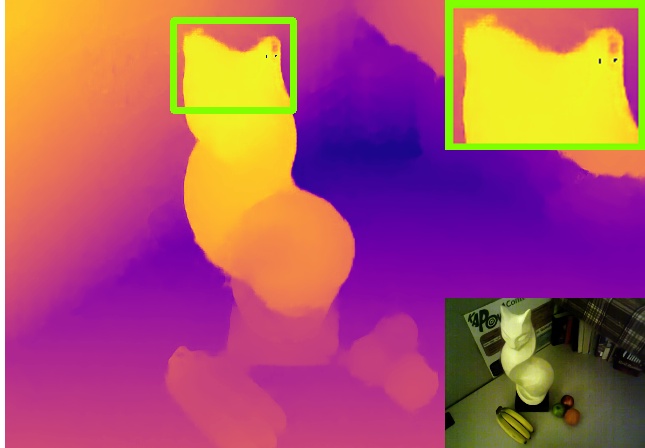}
    		\end{minipage}
    	}
    \subfigure[DJFR~\cite{PMBANet}]{
    \hspace{-0.1in}
    		\begin{minipage}[b]{0.24\linewidth}
   		 	\includegraphics[width=1\linewidth]{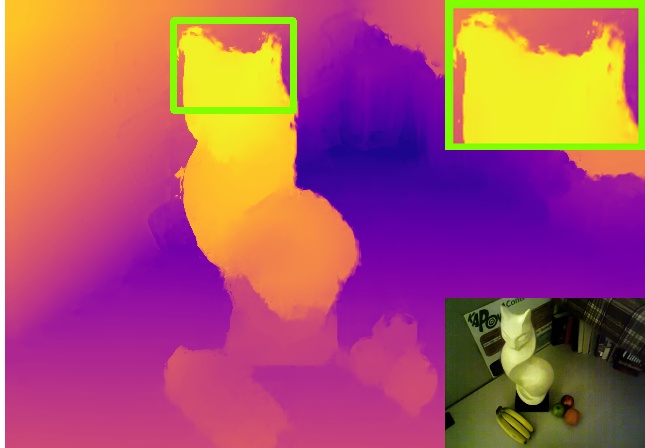}
    		\end{minipage}
    		\hspace{-0.1in}
    	}\subfigure[DKN~\cite{DKN}]{
    		\begin{minipage}[b]{0.24\linewidth}
   		 	\includegraphics[width=1\linewidth]{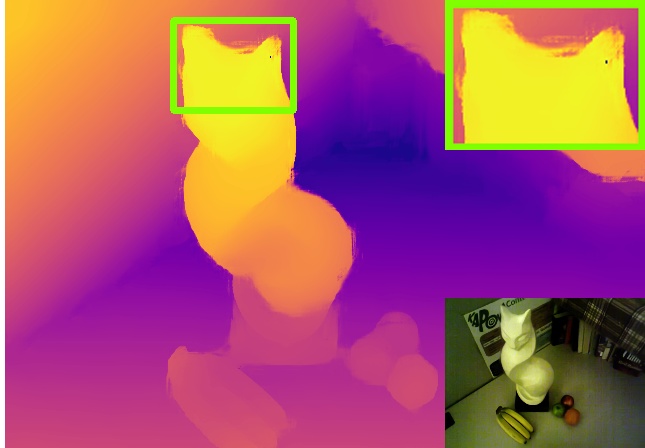}
    		\end{minipage}
    		\hspace{-0.1in}
    	}\subfigure[Ours]{
    		\begin{minipage}[b]{0.24\linewidth}
   		 	\includegraphics[width=1\linewidth]{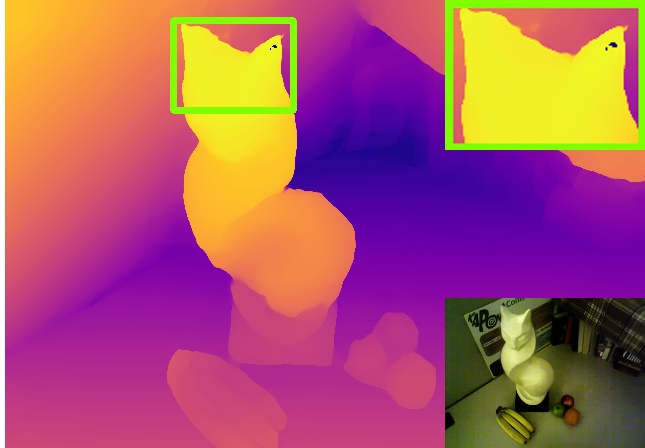}
    		\end{minipage}
    		\hspace{-0.1in}
    	}\subfigure[Ground Truth]{
    		\begin{minipage}[b]{0.24\linewidth}
   		 	\includegraphics[width=1\linewidth]{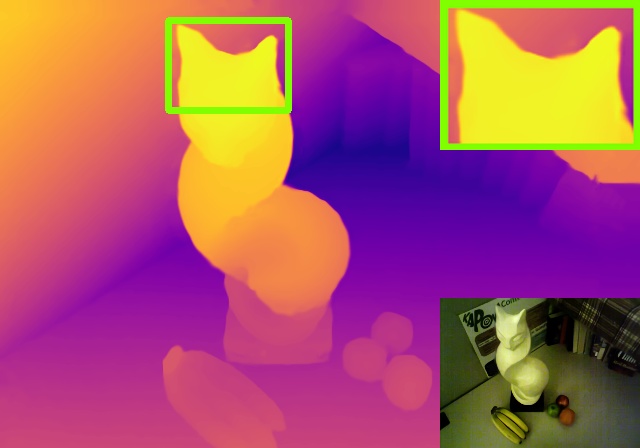}
    		\end{minipage}
    	}
	\end{center}
	\vspace{-0.15in}
	\caption{Visual comparison of $16\times$ results on \textit{Image\_05} from Lu dataset~\cite{Lu}. Please enlarge the PDF for
more details.}
	\label{fig:lu_vis}
	\vspace{-0.25in}
\end{figure}
For Middlebury dataset, we compare the proposed method with 15 state-of-the-art DSR methods: cross-based local multipoint filtering (CLMF1)~\cite{clmf}, anisotropic total generalized variation network (ATGV)~\cite{AGTV}, deep multi-scale guidance network (DMSG)~\cite{DMSG}, learned dynamic guidance network (DEDIE)~\cite{gu2017learning}, deep edge inference network (DEIN)~\cite{DEIN}, depth super-resolution network (DSRNet)~\cite{DepthSR}, color-guided coarse-to-fine network (CCFN)~\cite{wen2018deep}, guided super-resolution via pixel-to-pixel transformation (GSPRT)~\cite{lutio2019guided}, deep joint image filter (DJFR)~\cite{DJFR}, pixel-adaptive convolution neural network (PacNet)~\cite{PanNet}, common and unique network (CUNet)~\cite{CUNet}, multi-direction dictionary learning (MDSR)~\cite{MDDR}, progressive multi-branch aggregation network (PMBAN)~\cite{PMBANet}, deformable kernel network (DKN)~\cite{DKN} and cross-task knowledge network (CTKT)~\cite{CTKT}. For fair comparison, we re-train DJFR~\cite{DJFR}, Pacnet~\cite{PanNet}, CUNet~\cite{CUNet} DKN~\cite{DKN} and PMBAN~\cite{PMBANet} with the same datasets as ours. The results of other compared methods are obtained from the authors. We report MAE and RMSE values for $4\times$, $8\times$ and $16\times$ depth map super-resolution in Table~\ref{tab:mpi_result} and Table~\ref{tab:mpi_rmse}, respectively. As can be seen from Table~\ref{tab:mpi_result}, with respect to the average MAE, the proposed method achieves the best performance for all scaling factors, especially for $16\times$ case that is most difficult to recover. The superior performance benefits from the proposed attention-based  hierarchical  multi-modal fusion strategy, which can recover structure-consistent content and make a plausible prediction for regions  with inconsistent contents of the guidance.

We also show the zoomed results of various methods in Fig.~\ref{fig:mpi_vis}, from which we can see that most of the existing approaches cannot generate clear boundaries and suffer from various artifacts. Take \textit{Art} as an example, the results of CLMF1~\cite{clmf} and DEDIE~\cite{gu2017learning} are blur. The results of DEIN~\cite{DEIN} suffer from broken edge artifacts. The results of DJFR~\cite{DJFR} and DKN~\cite{DKN} cannot generate continuous boundaries of the highlighted pencil region. In contrast, the proposed method can produce sharp boundaries with less artifacts. We attribute this to the proposed feature fusion and collaboration modules, which fully exploit the relevant information from LR depth and HR guidance.

\subsubsection{Experimental Results on Middlebury Dataset (Noisy Case)}
Following~\cite{DepthSR}, we simulate the ToF-like degradation by adding Gaussian noise with standard deviation of 5 to the LR depth maps. We list the MAE and RMSE values for $4\times$, $8\times$ and $16\times$ upsampling on Middlebury dataset in Table~\ref{tab:mpi_noisy} and Table~\ref{tab:mpi_noisy_rmse}, respectively. From these tables, it can be found that the proposed method can better deal with the effect of noise when upsampling the depth maps, even compared to the recently proposed deep learning methods, \textit{i.e.}, CUNet~\cite{CUNet}, PMBAN~\cite{PMBANet} and DKN~\cite{DKN}. We further present the qualitative results of $16\times$ downsampling and noisy degradation in Fig.~\ref{fig:mpi_noisy_vis}. We can see that our proposed method achieves the clearest and sharpest results. Both CUNet~\cite{CUNet} and DKN~\cite{DKN} generate competitive MAE scores, but suffer from edge diffusion artifacts.

\subsubsection{Evaluation on Generalization Ability across Datasets}

To verify the generalization ability of the proposed method, we test our method on Lu dataset, which is from different resources to the training dataset. The comparison study is conducted with DepthSR~\cite{DepthSR}, CUNet~\cite{CUNet}, PacNet~\cite{PanNet}, DJFR~\cite{DJFR}, PMBAN~\cite{PMBANet} and DKN~\cite{DKN}, which are all trained on the same datasets for fair comparison. The quantitative results are illustrated in Table~\ref{tab:Lu_result} (MAE) and Table~\ref{tab:Lu_result_rmse} (RMSE), from which we can see that the proposed method obtains the best performance among compared methods for most upsampling factors. We show the visual comparison of the compared methods in Fig~\ref{fig:lu_vis}. From Fig.~\ref{fig:lu_vis}(a), it can be found that, in the initial bicubic upsampling, there are holes in a large region. The method DGDIE~\cite{gu2017learning} works well in handling the holes due to that it performs guided filtering iteratively. The deep neural networks based methods, such as CUNet~\cite{CUNet}, PacNet~\cite{PanNet}, DJFR~\cite{DJFR}, DKN~\cite{DKN} and ours, work in an end-to-end learning manner, which all suffer from the hole artifact more or less. Compared with other deep neural networks based methods, our scheme produces much sharper edges, as shown in the highlighted region.


\begin{table*}[!tb]\setlength{\tabcolsep}{10.57pt}
	\begin{center}
	\caption{\label{tab:nyu_result}Average RMSE performance comparison for scale factors $4 \times, 8 \times$ and $16 \times$ with direct downsampling degradation. The best performance is shown in \textbf{bold} and second best performance is the \underline{underscored} ones. For NYU v2~\cite{NYU} we calculate in centimeter, for other datasets we calculate RMSE with depth value scaled to [0, 255] (Lower RMSE values, better performance).}
	    \vspace{-0.1in}
		\begin{tabular}{l|rrr|rrr|rrr|rrr}
		\toprule
		\multirow{2}{*}{Method} & \multicolumn{3}{c|}{Middlebury \cite{middleblur_data_1}} & \multicolumn{3}{c|}{Lu \cite{Lu}} & \multicolumn{3}{c|}{NYU v2 \cite{NYU}}  & \multicolumn{3}{c}{Average} \\
		\cline{2-13}
		~ &$4\times$ & $8 \times$  & $16\times$  & $4\times$ & $8 \times$  & $16\times$  &$4\times$ & $8 \times$  & $16\times$ &$4 \times$ & $8 \times$  & $16\times$ \\
		\hline
		Bicubic & 4.44 & 7.58 & 11.87 & 5.07 & 9.22 & 14.27 & 8.16 & 14.22 & 22.32 & 5.89 & 10.34 & 16.15 \\
		MRF \cite{MRF} & 4.26 & 7.43 & 11.80 & 4.90 & 9.03 & 14.19 & 7.84 & 13.98 & 22.20 & 5.67 & 10.15 & 16.06 \\
		GF \cite{GF} & 4.01 & 7.22 & 11.70 & 4.87 & 8.85 & 14.09 & 7.32 & 13.62 & 22.03 & 5.40 & 9.90 & 15.94\\
		TGV \cite{TGV} & 3.39 & 5.41 & 12.03 & 4.48 & 7.58 & 17.46 & 6.98 & 11.23 & 28.13 & 4.95 & 8.07 & 19.21 \\
		Park \cite{Park} & 2.82 & 4.08 & 7.26 & 4.09 & 6.19 & 10.14 & 5.21 & 9.56 & 18.10 & 4.04 & 6.61 & 11.83 \\
		Ham \cite{Ham} & 3.14 & 5.03 & 8.83 & 4.65 & 7.73 & 11.52 & 5.27 & 12.31 & 19.24 & 4.35 & 8.36 &  13.20 \\
		JBU \cite{JBU} & 2.44 & 3.81 & 6.13 & 2.99 & 5.06 & 7.51 & 4.07 & 8.29 & 13.35 & 3.17 & 5.72 & 9.00\\
		DGF~\cite{DGF} & 3.92& 6.04& 10.02& 2.73& 5.98& 11.73& 4.50& 8.98& 16.77 & 3.72 & 7.00 & 12.84 \\
		DJF \cite{DJF} & 2.14 & 3.77 & 6.12 & 2.54 & 4.71 & 7.66 & 3.54 & 6.20 & 10.21 & 2.74 & 4.89 & 8.00 \\
		DMSG \cite{DMSG} & 2.11 & 3.74 & 6.03 & 2.48 & 4.74 & 7.51 & 3.37 & 6.20 & 10.05 & 2.65 & 4.89 & 7.86 \\
		PacNet~\cite{PanNet} & 1.91 & 3.20 & 5.60 & 2.48 & 4.37 & 6.60 & 2.82 & 5.01 & 8.64 & 2.40 & 4.19 & 6.95 \\
        DJFR \cite{DJFR} & 1.98 & 3.61 & 6.07 & \underline{2.21} & \textbf{3.75} & 7.53 & 3.38 & 5.86 & 10.11 & 2.52 & 4.41 & 7.90 \\
        DSRNet~\cite{DepthSR} & 2.08 & 3.26 &  5.78 &  2.57 &  4.46 &  6.45 &  3.49 &  5.70  & 9.76  & 2.71 & 4.47 & 7.30 \\
        FDKN \cite{DKN}& 2.21 & 3.64 & 6.15 & 2.64 & 4.55 & 7.20 & 2.63 & 4.99 & 8.67 & 2.49 & 4.39 & 7.34\\
		DKN \cite{DKN} & \underline{1.93} & \underline{3.17} & \underline{5.49} & 2.35 & 4.16 & \underline{6.33} & \underline{2.46} & \underline{4.76} & \underline{8.50} & \underline{2.25} & \underline{4.03} & \underline{6.77} \\
		AHMF (Ours) & \textbf{1.79} & \textbf{2.81} & \textbf{5.02} & \textbf{2.00} & \underline{3.83} & \textbf{6.21} & \textbf{2.25} & \textbf{4.50} & \textbf{8.10} & \textbf{2.01} & \textbf{3.71} & \textbf{6.41} \\
		\bottomrule
		\end{tabular}
	\end{center}
	\vspace{-0.2in}
\end{table*}
\begin{figure*}[!tb]
    \centering
    \includegraphics[width=\linewidth]{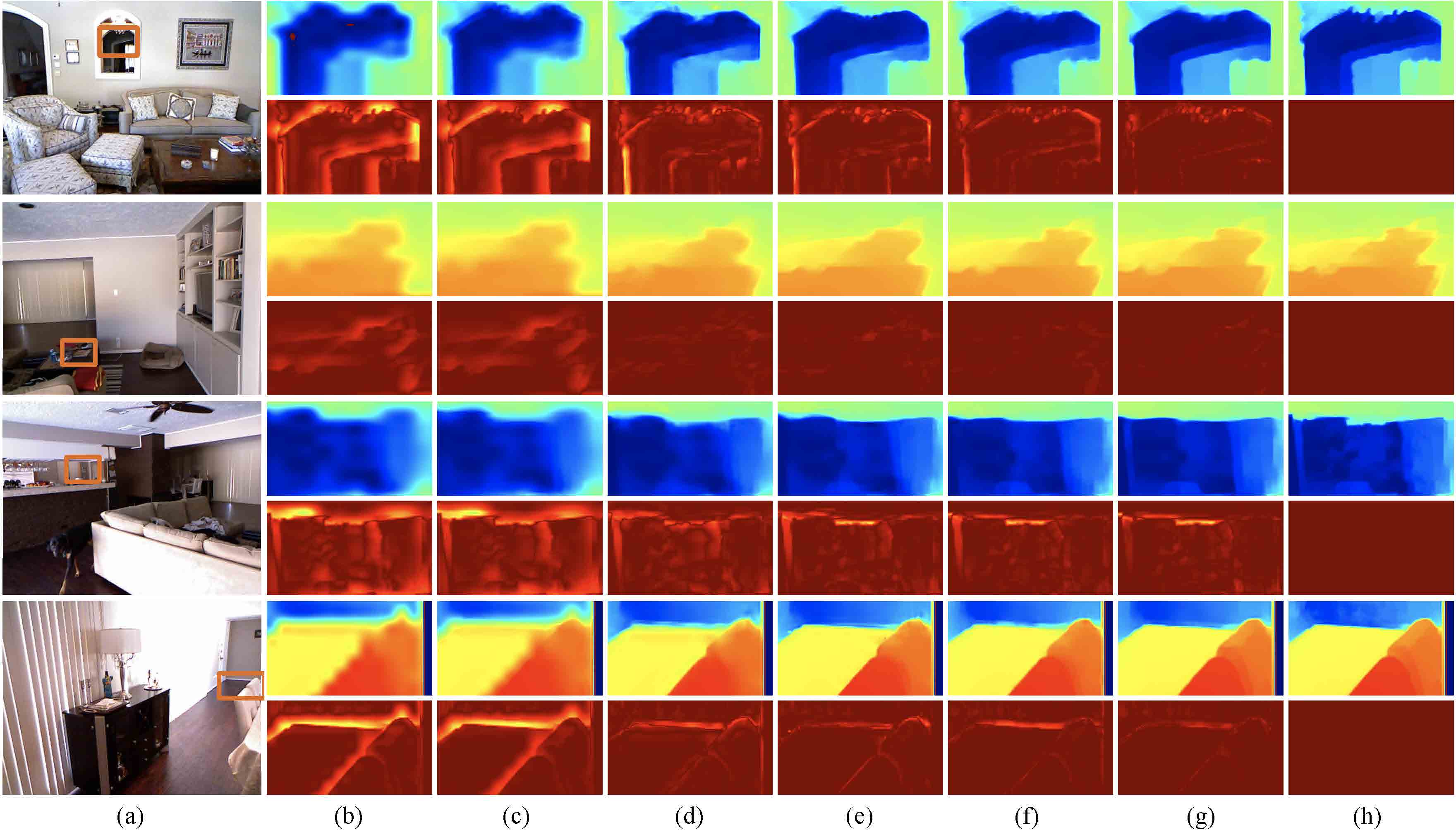}
    \vspace{-0.3in}
    \caption{Visual comparison of $8\times$ results on \textit{Image\_1339}, \textit{Image\_1242}, \textit{Image\_1241} and \textit{Image\_1360} from NYU v2~\cite{NYU} dataset: (a): Guidance Image, (b): Bicubic, (c): GF~\cite{GF} (d): DJFR~\cite{DJFR}, (e): DKN~\cite{DKN}, (f): PacNet~\cite{PanNet}, (g): Ours, (h): GT. For each sample, the second row shows the error map between the results and ground truth. In the error map, brighter area means the larger error. Please enlarge the PDF for more details.}
    \label{fig:nyu_vis}
    \vspace{-0.1in}
\end{figure*}
\begin{table*}[!tb]\setlength{\tabcolsep}{10.7pt}
	\begin{center}
	\caption{\label{tab:nyu_result_bic} Average RMSE performance comparison for scale factors $4 \times, 8 \times$ and $16 \times$ with bicubic degradation. The best performance is shown in \textbf{bold} and second best performance is the \underline{underscored} ones. For NYU v2~\cite{NYU} we calculate in centimeter, for other datasets we calculate RMSE with depth value scaled to [0, 255] (Lower RMSE values, better performance).}
	\vspace{-0.1in}
		\begin{tabular}{l|rrr|rrr|rrr|rrr}
		\toprule
		\multirow{2}{*}{Method} & \multicolumn{3}{c|}{Middlebury \cite{ middleblur_data_1}} & \multicolumn{3}{c|}{Lu \cite{Lu}} & \multicolumn{3}{c|}{NYU v2~\cite{NYU}}  & \multicolumn{3}{c}{Average} \\
		\cline{2-13}
		~ &$4\times$ & $8 \times$  & $16\times$  & $4\times$ & $8 \times$  & $16\times$  &$4\times$ & $8 \times$  & $16\times$ &$4 \times$ & $8 \times$  & $16\times$ \\
		\hline
		Bicubic & 2.47 & 4.65 & 7.49 &  2.63  & 5.23 & 8.77 & 4.71 & 8.29 & 13.17 & 3.27 & 6.06& 9.81   \\
		GF \cite{GF} & 3.24 & 4.36 & 6.79 & 4.18 & 5.34 & 8.02 & 5.84 & 7.86 & 12.41 & 4.42 & 5.85 & 9.07\\		
		TGV \cite{TGV} & 1.87 & 6.23 & 17.01 & 1.98 & 6.71 & 18.31 & 3.64 & 10.97 & 39.74 & 2.50 & 7.97 & 25.02 \\
		DGF~\cite{DGF}  & 1.94 & 3.36 & 5.81 & 2.45 & 4.42 & 7.26 & 3.21 & 5.92 & 10.45 & 2.53 & 4.57 &  7.84 \\
		DJF~\cite{DJF} & 1.68 & 3.24 & 5.62 & 1.65 & 3.96 & 6.75 & 2.80 & 5.33 & 9.46 & 2.04 & 4.18 & 7.28 \\
		DMSG~\cite{DMSG} & 1.88 & 3.45 & 6.28 & 2.30 & 4.17 & 7.22 & 3.02 & 5.38 & 9.17 & 2.40 & 4.33 & 7.17  \\
		DJFR~\cite{DJFR}& 1.32 & 3.19 & 5.57 & 1.15 & 3.57 & 6.77 & 2.38 & 4.94 & 9.18  & 1.62 & 3.90  & 7.17\\
		DSRNet~\cite{DepthSR} & 1.77 &  3.05 &  4.96 & 1.77 &  3.10 &  \underline{5.11} & 3.00 &  5.16 &  8.41 & 2.18  & 3.77 & 6.16 \\
		PacNet~\cite{PanNet} &  1.32 & 2.62 & 4.58 & 1.20 & 2.33 & 5.19 & 1.89 & 3.33 & 6.78& 1.47  & 2.76 & 5.53 \\
		FDKN~\cite{DKN} & \underline{1.08} & 2.17 &  4.50 & \textbf{0.82} & 2.10 & 5.05 & 1.86 & 3.58 & 6.96 & 1.25 & 2.62 & 5.50 \\
		DKN~\cite{DKN} &  1.23 & \underline{2.12} &\underline{4.24} & 0.96 & \underline{2.16} & \underline{5.11} & \underline{1.62} & \underline{3.26} & \underline{6.51} & 1.27 & 2.51 &5.29 \\
		AHMF (Ours) & \textbf{1.07} & \textbf{1.63} & \textbf{3.14} & \underline{0.88}& \textbf{1.66} &\textbf{3.71} & \textbf{1.40} & \textbf{2.89} & \textbf{5.64} & \textbf{1.12} & \textbf{2.06} & \textbf{4.16} \\
		\bottomrule
		\end{tabular}
	\end{center}
	\vspace{-0.25in}
\end{table*}

\begin{figure*}[!tb]
    \centering
    \includegraphics[width=\linewidth]{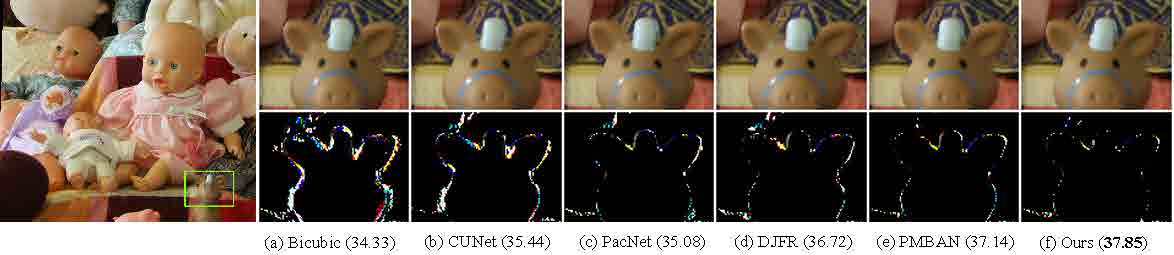}
    \vspace{-0.33in}
    \caption{Visual comparisons of depth image based rendering on \textit{Dolls} from Middlebury~\cite{middleblur_data_1} dataset. The second row shows the error maps between the results and ground truth, in which brighter areas mean larger errors. Please enlarge the PDF for more details.}
    \label{fig:dibr_vis}
    \vspace{-0.25in}
\end{figure*}
\begin{figure}[!tb]
    \centering
    \hspace{-0.1in}
    \includegraphics[width=\linewidth]{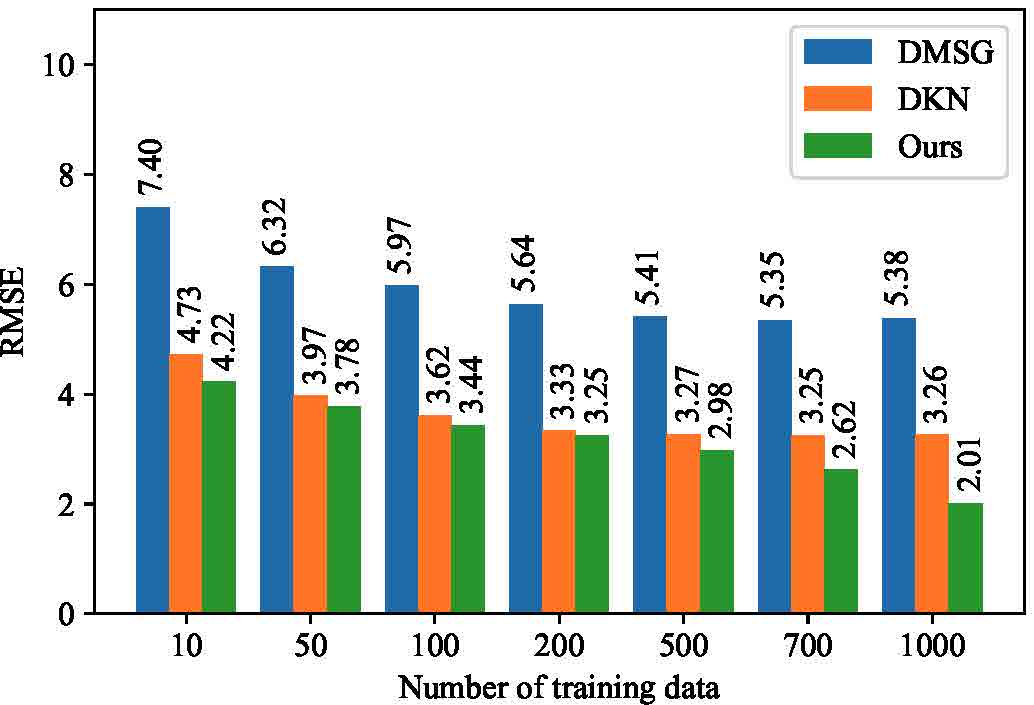}
    \vspace{-0.1in}
    \caption{RMSE comparison by varying the number of training data on NYU v2~\cite{NYU} dataset for $4\times$ depth map super-resolution.}
    \label{fig:diff_nyu}
    \vspace{-0.2in}
\end{figure}

\subsubsection{Experimental Results on NYU v2 Dataset}

To evaluate the effectiveness of the proposed method, we conduct experiments on NYU v2 dataset. We compare our method with DJFR~\cite{DJFR}, DKN~\cite{DKN} and PacNet~\cite{PanNet}. For fairness, we re-train PacNet~\cite{PanNet} and DSRNet~\cite{DepthSR} with the same training dataset as ours. Since these methods are evaluated by RMSE index in their original papers, we report the RMSE performance for direct downsampling and Biucbic downsampling in Table~\ref{tab:nyu_result} and Table~\ref{tab:nyu_result_bic}, respectively. Obviously, our method outperforms all compared methods with a large margin for average RMSE values. To further analyze the performance of our method, we show visual results at 8$\times$ upsampling for \textit{Image\_1241}, \textit{Image\_1242}, \textit{Image\_1339} and \textit{Image\_1360} in Fig.~\ref{fig:nyu_vis}. As can be seen, the results of GF~\cite{GF} are over-smoothed due to that the local filter cannot capture global information. The results of DJFR~\cite{DJFR} and DKN~\cite{DKN} suffer from diffusion artifacts. The result of PacNet~\cite{PanNet} can preserve the local details, but cannot reconstruct the boundary well. On the contrary, our method generates depth maps with smaller reconstruction errors than other compared methods. Besides, to demonstrate the robustness of our method for different size of training data, we train our model by varying the number of training data, and report the RMSE values in Fig.~\ref{fig:diff_nyu}. As the result shows, our method achieves the best performance for all case. In particular, the proposed method trained with only 500 images obtains the similar performance with DKN~\cite{DKN} which is the state-of-the-art approach on NYU v2~\cite{NYU} dataset.

\subsubsection{Experimental Results on Depth Image based Rendering}
\begin{table}[!tb]\setlength{\tabcolsep}{4.1pt}
	\begin{center}
	    \caption{\label{dibr_res}PSNR values for depth image based rendering.}
	    \vspace{-0.1in}
	    \renewcommand{\arraystretch}{1.1}
		\begin{tabular}{l|cccccc}
		\toprule
		Method & Art & Books & Dolls & Laundry & Moebius & Reindeer\\
		\hline
        Bicubic& 25.43& 32.42& 34.33& 26.19& 32.53 & 29.76  \\  
        DSRNet~\cite{DepthSR}& 27.49& 33.71& 34.65& 28.10& 33.60 & 30.00  \\   
        CUNet~\cite{CUNet}& 29.78& 33.77& 35.44& 28.00& 34.09 & 33.58  \\   
        PacNet~\cite{PanNet}& 31.52& 35.08& 35.50& 30.50& 34.85 & 35.45  \\ 
        DJFR~\cite{DJFR}& 32.31& 35.95& 36.72& 30.55& 35.54 &  34.37   \\ 
        PMBAN~\cite{PMBANet}& \underline{33.01}& \underline{37.44}& \underline{37.14}& \underline{31.82}& \underline{35.96} &  \underline{35.98} \\
        Ours& \textbf{34.57}& \textbf{37.87}& \textbf{37.85}& \textbf{32.64}& \textbf{37.22} & \textbf{38.49} \\
		\bottomrule
		\end{tabular}
	\end{center}
	\vspace{-0.25in}
\end{table}
Similar to \cite{zhang2019color}, we take depth image based rendering (DIBR) as a measurement to evaluate the performance of the depth SR methods. The stereo depth map used for DIBR is down-sampled by bicubic with 8$\times$ scaling factor, then the super-resolved depth maps are used to perform DIBR. Table~\ref{dibr_res} tabulates the  PSNR values for the compared methods on Middlebury dataset. It can be observed that the proposed method achieves the best performance for all test images. We present the visual comparison results and error maps for image \textit{Dolls} in Fig.~\ref{fig:dibr_vis}. Clearly, the proposed AHMF method can produce visually appealing results with smaller error maps.

\subsection{Ablation Study}
\label{abs}
\begin{table*}[!tb]\setlength{\tabcolsep}{12.7pt}
	\begin{center}
	    \caption{\label{tab:Ablation_Study}Ablation study of multi-modal attention based fusion (MMAF) and bi-directional hierarchical feature collaboration (BHFC). The experiments are conducted on Middlebury 2005~\cite{middleblur_data_1} dataset. We report the average MAE values for all variants.}
	    \vspace{-0.11in}
		\renewcommand{\arraystretch}{1.2}
		\begin{tabular}{l|cc|ccc} 
		\toprule
		\multirow{2}{*}{Model} & \multirow{2}{*}{Multi-Modal Fusion} & \multirow{2}{*}{Hierarchical Feature Collaboration} & \multicolumn{3}{c}{MAE} \\
		\cline{4-6}
		~ & ~ & ~ &  $4\times$& $8\times$& $16\times$\\
		\hline 
		\texttt{Model1} &  \texttt{Addition} & \texttt{Forward+Backward} &  0.1621 & 0.3391 & 0.7204\\
		 \texttt{Model2} &  \texttt{Concatenation} & \texttt{Forward+Backward} &0.1586 & 0.3307 & 0.7126 \\
		 \texttt{Model3} &  \texttt{MMAF} & \texttt{w/o Feature Collaboration} &0.1640 & 0.3329 & 0.7411 \\
		 \texttt{Model4} & \texttt{MMAF} & \texttt{w/o Backward} & 0.1592 & 0.3315 & 0.7337 \\ 
		 \texttt{Model5} & \texttt{MMAF} &\texttt{w/o Forward} & 0.1584 & 0.3305 & 0.7318  \\
		 \hline
		 \texttt{Model6 (our full model)} & \texttt{MMAF} & \texttt{Forward+Backward (BHFC)} &  \textbf{0.1574 }& \textbf{0.3269} & \textbf{0.7058}  \\
		\bottomrule
		\end{tabular}
	\end{center}
	\vspace{-0.2in}
\end{table*}
\begin{figure*}[!tb]
    \centering
    \includegraphics[width=\linewidth]{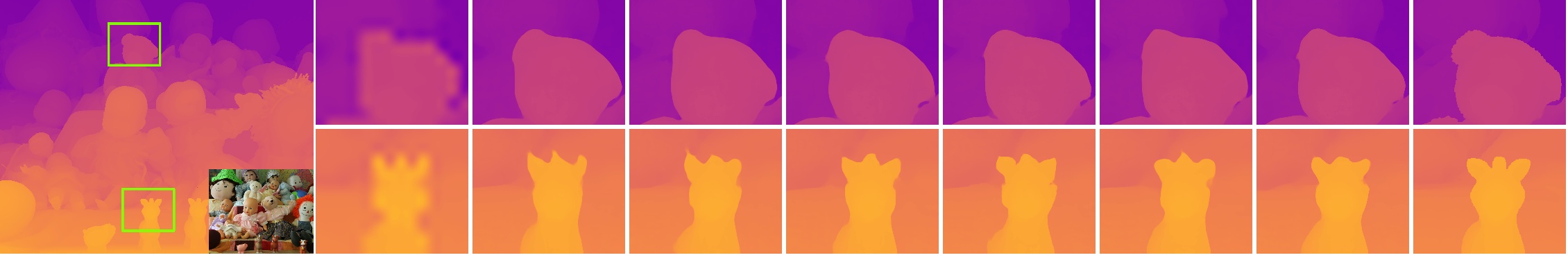}
    \vspace{-0.33in}
    \caption{Visual comparison of $16\times$ depth map super-resolution on \emph{Dools} from Middlebury 2005 dataset~\cite{middleblur_data_1}: (a): Bicbubic, (b): Model1, (c): Model2, (d): Model3, (e): Model4, (f): Model5, (g): Model6 (ours full models, AHMF). Please enlarge the PDF for more details.}
    \label{fig:abl1}
    \vspace{-0.25in}
\end{figure*}
\begin{figure}[!tb]
	\begin{center}
		\subfigure[$4\times$ super-resolution]{
		\begin{minipage}[b]{0.48\linewidth}
			\includegraphics[width=0.9\linewidth]{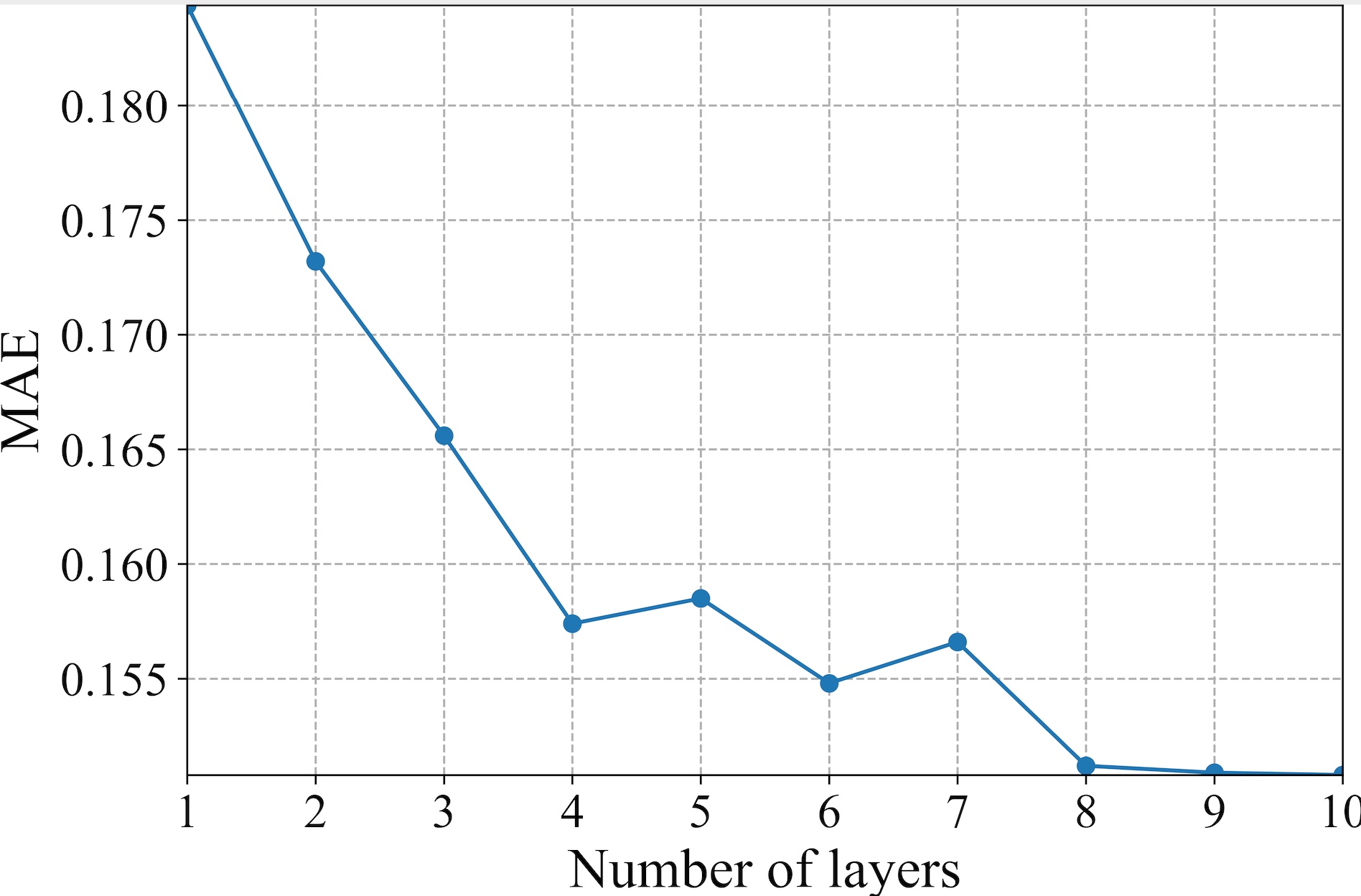} 
		\end{minipage}
		\hspace{-0.1in}
	}\subfigure[$8\times$ super-resolution]{
    		\begin{minipage}[b]{0.48\linewidth}
  		 	\includegraphics[width=0.9\linewidth]{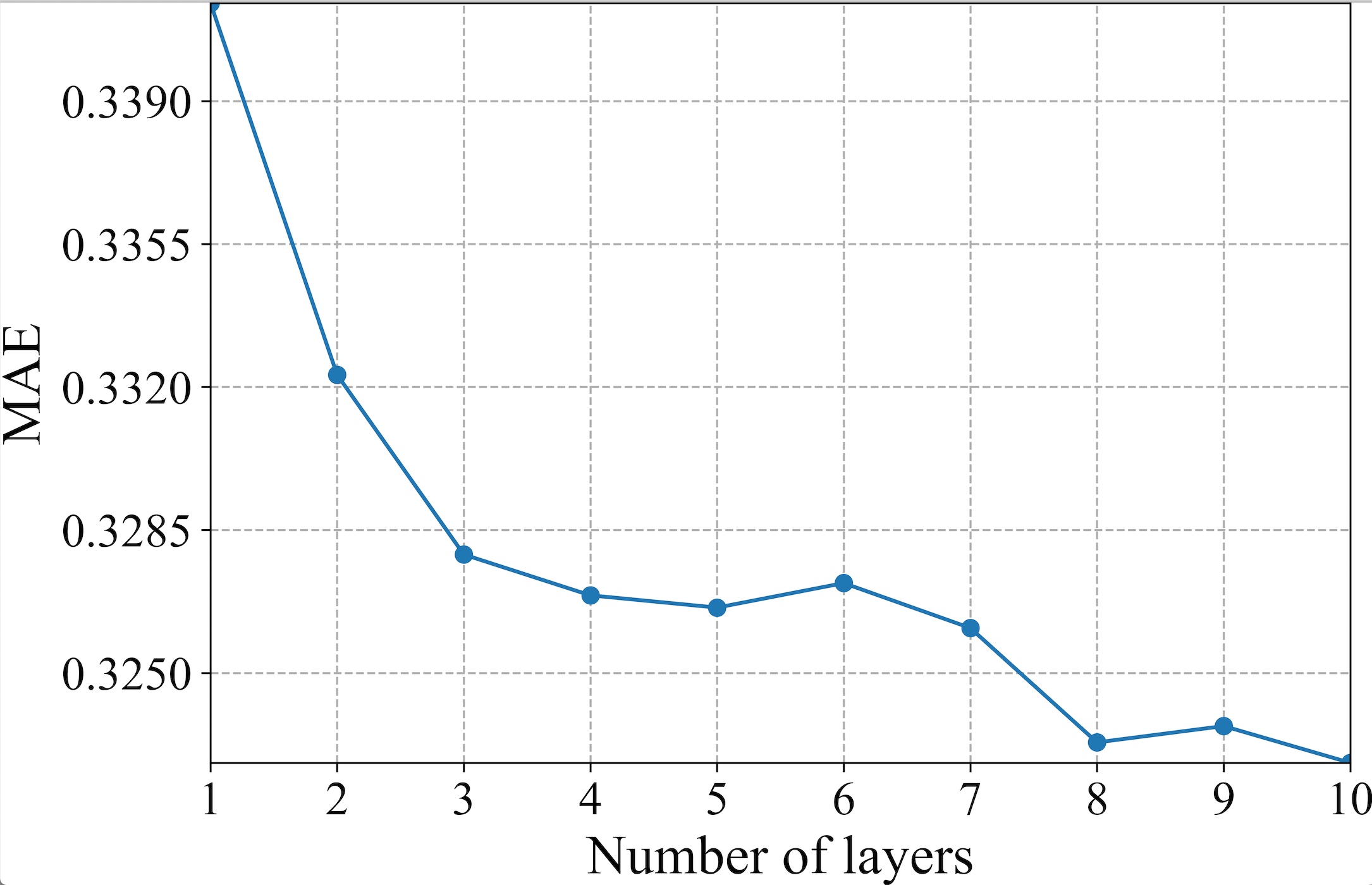}
    		\end{minipage}
    	}
	\end{center}
	\vspace{-.15in}
	\caption{Ablation study of the number of feature extraction layers $m$ on Middlebury dataset~\cite{middleblur_data_1}. We report the average MAE for all variants.}
	\label{fig:sm}
	\vspace{-0.2in}
\end{figure}
\begin{figure}[!tb]
	\begin{center}
		\subfigure[$4\times$ super-resolution]{
		\begin{minipage}[b]{0.48\linewidth}
			\includegraphics[width=0.9\linewidth]{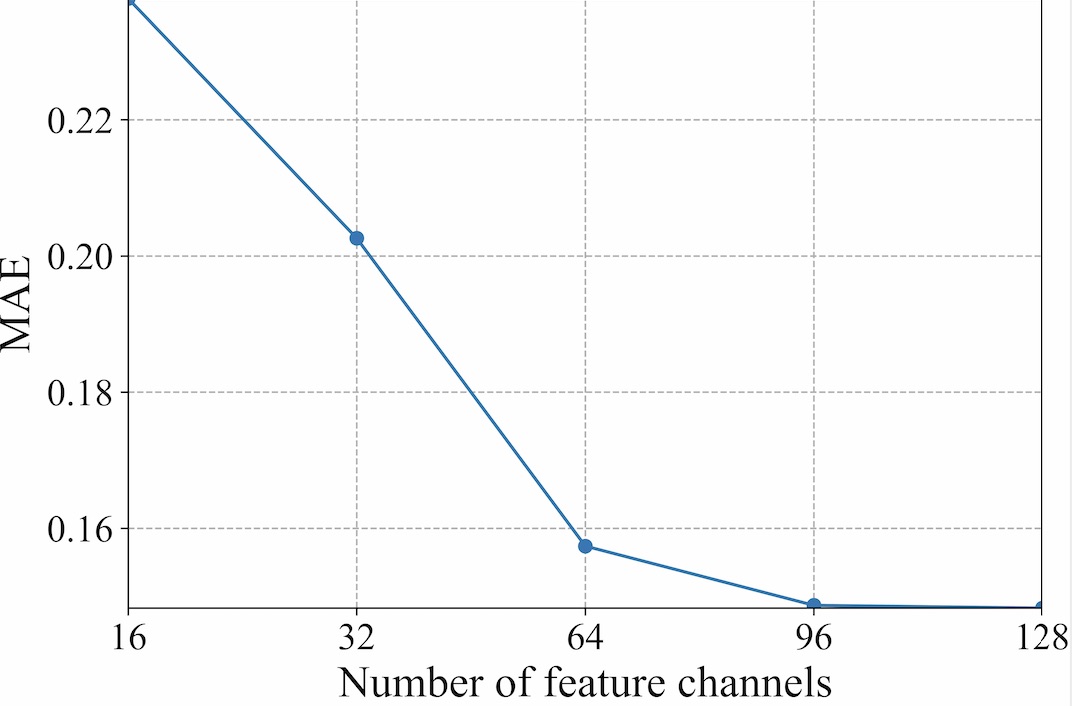} 
		\end{minipage}
		\hspace{-0.1in}
	}\subfigure[$8\times$ super-resolution]{
    		\begin{minipage}[b]{0.48\linewidth}
  		 	\includegraphics[width=0.9\linewidth]{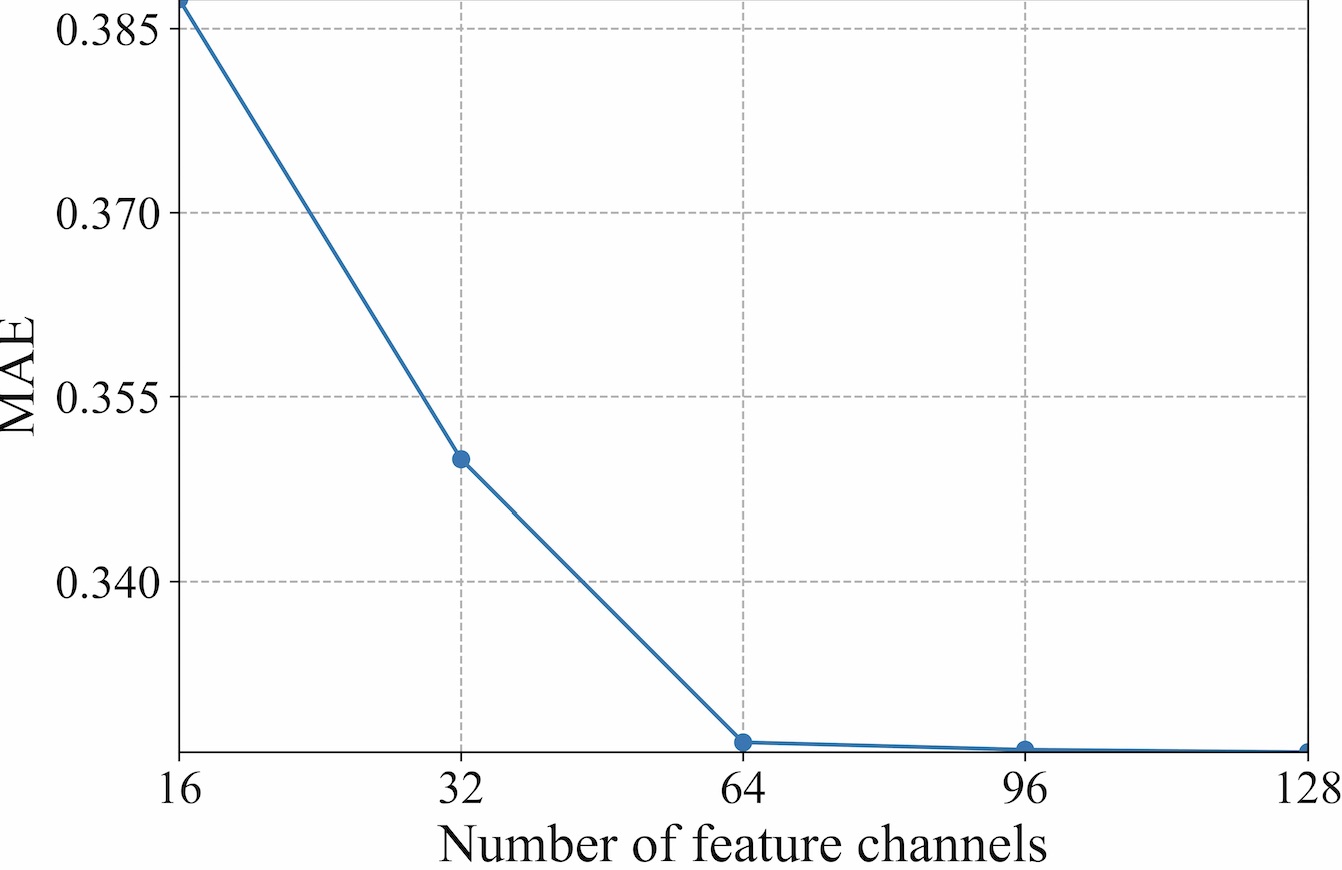}
    		\end{minipage}
    	}
	\end{center}
	\vspace{-.15in}
	\caption{Ablation study of the number of feature channels $n$ on Middlebury dataset~\cite{middleblur_data_1}. We report the average MAE for all variants.}
	\label{fig:sc}
	\vspace{-0.2in}
\end{figure}

In this subsection,  we conduct ablation studies to verify influence of different configurations to the final performance and the effectiveness of the proposed multi-modal attention fusion (MMAF) and bi-directional hierarchical feature collaboration (BHFC) module.

\subsubsection{Influence of Different Configurations to the Final Performance}
We first investigate effect of various parameter settings and effect of different loss functions to the final performance. 

\textbf{Effect of various parameter settings.} To train the proposed deep model, there are some critical parameters needed to carefully set, including the number of feature extraction layers $m$ and the number of feature channels $n$. We provide experimental analysis about the influence of various settings to the final performance. Fig.~\ref{fig:sm} shows the MAE trend with the layer number $m$. It can be found that the MAE values decrease rapidly before $m=4$, and the larger $m$ tends to lead to the better reconstruction performance, which however is at the expense of network complexity. For example, the model's parameters are 4.20M when $m=8$, which are about twice larger than that of $m=4$ (2.54M). In the proposed method, we empirically choose $m=4$ to obtain a good trade-off between the network complexity and reconstruction performance. Moreover, we plot the MAE curve with the feature channel number $n$ in Fig.~\ref{fig:sc}. It can be found that the MAE values decrease rapidly before $n=64$. With similar purpose as $m$, we set $n$ as 64 in our model. 

\begin{table}[!tb]\setlength{\tabcolsep}{3.4pt}
	\begin{center}
	    \caption{\label{tab:loss}Performance comparison (MAE / RMSE) for $4 \times$ depth map super-resolution with different loss functions.}
	    \vspace{-0.1in}
		\renewcommand{\arraystretch}{1.2}
		\begin{tabular}{l|ccc}
		\toprule
		Loss &  Middlebury~\cite{middleblur_data_1} & NYU v2~\cite{NYU} & Lu~\cite{Lu}\\
		\hline 
		 $\mathbf{L}_1$   & \textbf{0.1574} / 0.7021 & \textbf{0.4533} / \textbf{1.4013} &  \textbf{0.4467} / \textbf{1.3283} \\
		 \textbf{L}$_2$  & 0.2160 / \textbf{0.6972} & 0.5514 / 1.4844 & 0.4492 /  1.4843 \\
		 \textbf{L}$_1$+Perceptual Loss & 0.2237 / 0.7594 & 0.5834 / 1.6131 & 0.4837 / 1.5190 \\
		\bottomrule
		\end{tabular}
	\end{center}
	\vspace{-0.3in}
\end{table}
\textbf{Effect of different loss functions.}
We investigate the influence of different loss functions to the final performance. The comparison study group includes $\textbf{L}_1$, $\textbf{L}_2$ and perceptual loss. The experimental analysis is conducted on Middlebury~\cite{middleblur_data_1}, NYU v2~\cite{NYU} dataset and Lu~\cite{Lu} dataset. The comparison results are provided in Table~\ref{tab:loss}, from which it can be found that, the $\textbf{L}_1$ loss achieves the best results with respect to MAE for all test datasets, and achieves the best results with respect to RMSE for two datasets. This is mainly because $\textbf{L}_1$ is more robust to outliers, thus the depth boundaries are preserved better than $\textbf{L}_2$ and perceptual loss. According to this analysis, we choose $\textbf{L}_1$ as our loss function.

\subsubsection{Influence of Different Modules to the Final Performance}

We further investigate the role of multi-modal attention fusion and hierarchical feature collaboration to the final performance. As shown in Table~\ref{tab:Ablation_Study}, the experimental analysis is conducted on Mddilbury 2005 dataset~\cite{middleblur_data_1} with five different variants. For fair comparison, we carefully adjust the feature channels of the networks to guarantee that different variants have roughly the same size of parameters as that of our full model. Specifically, for $4\times$, $8\times$ and $16\times$ super-resolution, the sizes of parameters of all variants are around 2.54M, 3.36M and 5.75M, respectively.


\textbf{Effect of Multi-modal Attention Fusion.} The multi-modal attention fusion model is used to fuse the extracted depth and guidance features. To validate the effectiveness of our proposed strategy, we compare it with two alternatives: 1) \texttt{Model1}, we replace the MMAF by addition operation to fuse the extracted features but keep the BHFC unchanged, 2) \texttt{Module2}, in which we use concatenation operation to fuse the extracted features but keep the BHFC unchanged. We report the results in Table~\ref{tab:Ablation_Study}. Compared with the proposed MMAF, simply fusing the multi-modal features by addition or concatenation significantly worsen the results, especially for the large up-sampling factor. We can observe that the result of \texttt{Module2} is a slightly better than that of \texttt{Module1}. This is easy to understand, since the features of depth and guidance are heterogeneous, and the same pixel values may represent totally different objects. Directly fuse them by addition would destroy the useful information. Fig.~\ref{fig:abl1} further shows the visual comparisons of different variants for $16\times$ depth map super-resolution. As can be seen, \texttt{Model1} and \texttt{Model2} cannot generate accurate depth boundaries for small objects, \eg, head of the toy, since the input low-resolution depth image is severely damaged. In contrast, our method clearly reconstructs the depth boundaries and obtains the best performance.

\begin{figure}[!tb]
	\begin{center}
		\subfigure[Guidance]{
		\begin{minipage}[b]{0.23\linewidth}
			\includegraphics[width=1\linewidth]{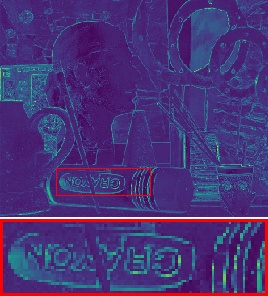} 
		\end{minipage}
		\hspace{-0.12in}
	}\subfigure[Depth]{
    		\begin{minipage}[b]{0.23\linewidth}
  		 	\includegraphics[width=1\linewidth]{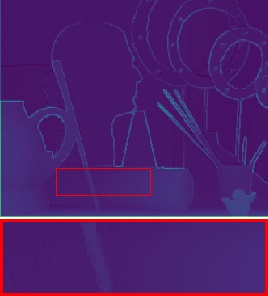}
    		\end{minipage}
    		\hspace{-0.12in}
    	}\subfigure[Concat$_1$]{
    		\begin{minipage}[b]{0.23\linewidth}
  		 	\includegraphics[width=1\linewidth]{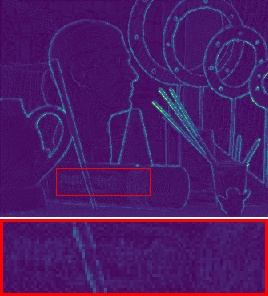}
    		\end{minipage}
    		\hspace{-0.12in}
    }\subfigure[Concat$_2$]{
    		\begin{minipage}[b]{0.23\linewidth}
  		 	\includegraphics[width=1\linewidth]{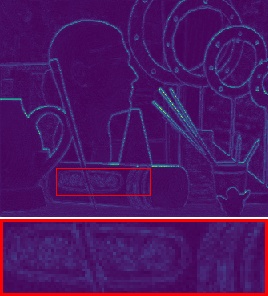}
    		\end{minipage}
    }
    \subfigure[Guidance]{
		\begin{minipage}[b]{0.23\linewidth}
			\includegraphics[width=1\linewidth]{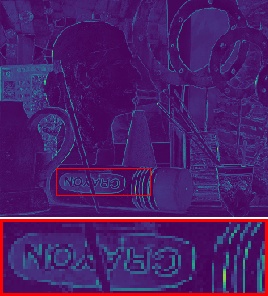} 
		\end{minipage}
		\hspace{-0.12in}
	}\subfigure[Depth]{
    		\begin{minipage}[b]{0.23\linewidth}
  		 	\includegraphics[width=1\linewidth]{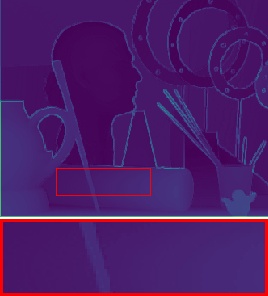}
    		\end{minipage}
    		\hspace{-0.12in}
    	}\subfigure[MMAF$_1$]{
    		\begin{minipage}[b]{0.23\linewidth}
  		 	\includegraphics[width=1\linewidth]{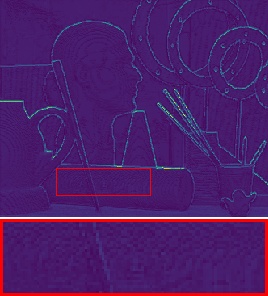}
    		\end{minipage}
    		\hspace{-0.12in}
    }\subfigure[MMAF$_2$]{
    		\begin{minipage}[b]{0.23\linewidth}
  		 	\includegraphics[width=1\linewidth]{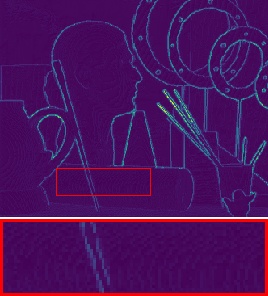}
    		\end{minipage}
    }
	\end{center}
	\vspace{-.15in}
	\caption{Visualization of feature maps (\emph{Art} for $4\times$ depth map super-resolution). (a) and (e) features maps extracted from guidance image; (b) and (f) features maps extracted from depth image; (c) and (g) feature maps of the first layer fused by concatenation (the first row) and the proposed MMAF (the second row), respectively; (d) and (h) feature maps of the second layer fused by concatenation (the first row) and the proposed MMAF (the second row), respectively. Please enlarge the PDF for more details.}
	\label{fig:map}
	\vspace{-0.25in}
\end{figure}

Moreover, we visualize the input and output feature maps of Fusion by Concatenation (the first row) and  MMAF (the second row) in Fig.~\ref{fig:map}. Taking the first row as an example, Fig.~\ref{fig:map} (a) and Fig.~\ref{fig:map} (b) are extracted guidance and depth feature maps, respectively. Fig.~\ref{fig:map} (c) ($\boldsymbol{F}_{f}^1$ in Eq.~\ref{eq6}) and Fig.~\ref{fig:map} (d) ($\boldsymbol{F}_{f}^2$  in Eq.~\ref{eq6}) show the features fused by Concatenation. From Fig.~\ref{fig:map}, we can find that: 1) the extracted guidance feature contains a lot of redundant texture information; 2) both of Concatenation and MMAF can fully fuse the multi-modal features of the consistent regions, \eg, the depth boundaries of the large-scale objects are enhanced; 3) for the inconsistent components, as highlighted in Fig.~\ref{fig:map} (d) and Fig.~\ref{fig:map} (h), the proposed MMAF can sufficiently filter out the redundant texture information while Concatenation suffers from texture-copying artifacts. This is mainly because that the proposed MMAF selects useful information and neglects the erroneous structures by the attention mechanism.

\textbf{Effect of Hierarchical Feature Collaboration.} In deep convolution neural networks, the low-level features typically contain rich spatial details, while the high-level features usually contain sufficient structure information, and the hierarchical features are complementary to each other. To fully leverage these hierarchical features, we propose a bi-directional hierarchical feature fusion module (BHFC) and it contains a forward and a backward process to facilitate the hierarchical features propagation and collaboration with each other. To validate this, we compare it with three different models, 1): \texttt{Model3}, we remove the BHFC in our model; 2): \texttt{Model4}, we remove the backward feature propagation process for this model, 3): \texttt{Model5}, we remove the forward feature propagation process for this model. The quantitative results are illustrated in Table~\ref{tab:Ablation_Study} and the qualitative results are presented in Fig.~\ref{fig:abl1}. The performance drop of the \texttt{Model3} verifies our motivation that fuse hierarchical features is necessary in depth map super-resolution. In \texttt{Model4}, only the low-level features can propagate to the high-level features, suffer from a slightly performance drop. The same conclusion can be drawn for the model \texttt{Model5}. Compare with these three variants, our full model \texttt{Model6} with both forward and backward process achieves the best performance, especially for large scaling factors which is more to recover. This phenomenon further reinforces the hypothesis that the proposed BHFC plays a significant role in our model.

\begin{figure}[!tb]
    \centering
    \includegraphics[width=0.9\linewidth]{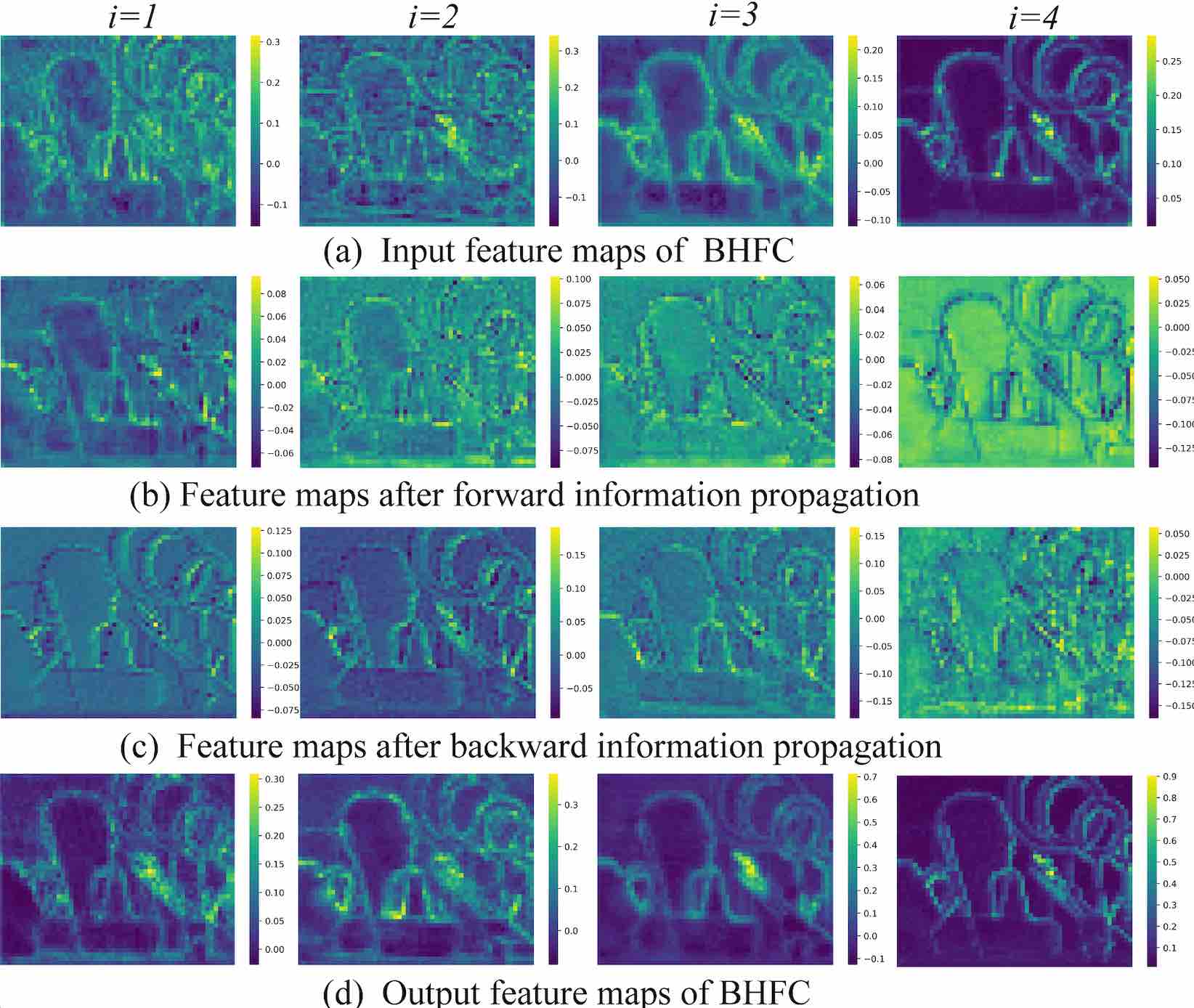}
    \vspace{-0.17in}
    \caption{Average feature maps before and after BHFC, i is the $i$-th layer.}
    \label{fig:bf_vis}
    \vspace{-0.18in}
\end{figure}

To further clarify the mechanism of hierarchical feature collaboration, we visualize the average feature maps of BHFC in Fig.~\ref{fig:bf_vis}. Two observations can be obtained. First, the features encoded by different layers have different attributes: shallower features contain rich details while deeper features contain clear structure information (Fig.~\ref{fig:bf_vis} (a)); Second, our proposed BHFC can propagate information from one layer to all other layers (Fig.\ref{fig:bf_vis} (b)-(c)), thus the output features of BHFC can maintain both of fine-grained details and clear structure information (Fig.\ref{fig:bf_vis} (d)). Thanks to this hierarchical feature collaboration mechanism, our proposed method achieves superior performance than the state-of-the-arts.

\red{\begin{table}[!tb]\setlength{\tabcolsep}{6.9pt}
	\begin{center}
	\caption{\label{tab:ncy}Average running time, GPU memory and network parameters comparison for $4\times$ super-resolution of different methods.}
	\vspace{-0.1in}
	    \renewcommand{\arraystretch}{1.2}
		\begin{tabular}{l|cccc}
		\toprule
		Method & Time (ms) & Memory (MB) & Paras (M) & MAE\\
		\hline
        DMSG~\cite{DMSG} & 32.59 & 1215 & 0.33 & 0.280  \\
        DJFR~\cite{DJFR} & 24.87 & 851 & 0.08 &0.228 \\
        DSRNet~\cite{DepthSR} & 105.38 & 2039  &45.49&0.255  \\
        PacNet~\cite{PanNet} & 50.33 & 4081  &  0.18 &0.235\\
        CUNet~\cite{CUNet}  & 104.91  & 1497    & 0.49& 0.275 \\
        DKN~\cite{DKN} & 204.11  & 1353  & 1.16 &0.173\\
        PMBAN~\cite{PMBANet}& 476.61 & 3145  & 25.06 & 0.183 \\
        AHMF& 78.32 & 1035 & 2.54 & 0.157 \\
		\bottomrule
		\end{tabular}
	\end{center}
	\vspace{-0.25in}
\end{table}}
\subsection{Comparison of Network Complexity}
\label{cc}
In this subsection, we provide experimental analysis about the comparison of our method with other state-of-the-art methods with respect to running time, memory cost and parameter size. In Table~\ref{tab:ncy}, we report the average running time (ms), GPU memory consumption (MB) and network parameter size (M) for $4\times$ depth map super-resolution on the Middlebury~\cite{middleblur_data_1}. The compared methods are run on the same server with an Intel Core i7 3.6GHz CPU and a NVIDIA GTX 1080ti GPU. To obtain the average running time and memory consumption, we crop the test images into the size of $480 \times 640$, and run these methods on these images for 500 times to calculate the average values. From Table~\ref{tab:ncy}, it can be found that the proposed method achieves the best MAE performance, while with less running time and memory cost than PMBAN~\cite{PMBANet}, DKN~\cite{DKN} and CUNet~\cite{CUNet}, which are the most recently published methods. The size of parameters of our method is also much smaller than PMBAN~\cite{PMBANet}. This result shows that our scheme achieves a better trade-off between reconstruction performance and network complexity.

\begin{figure}[!tb]
	\begin{center}
		\subfigure[Guidance]{
		\begin{minipage}[b]{0.194\linewidth}
			\includegraphics[width=1\linewidth]{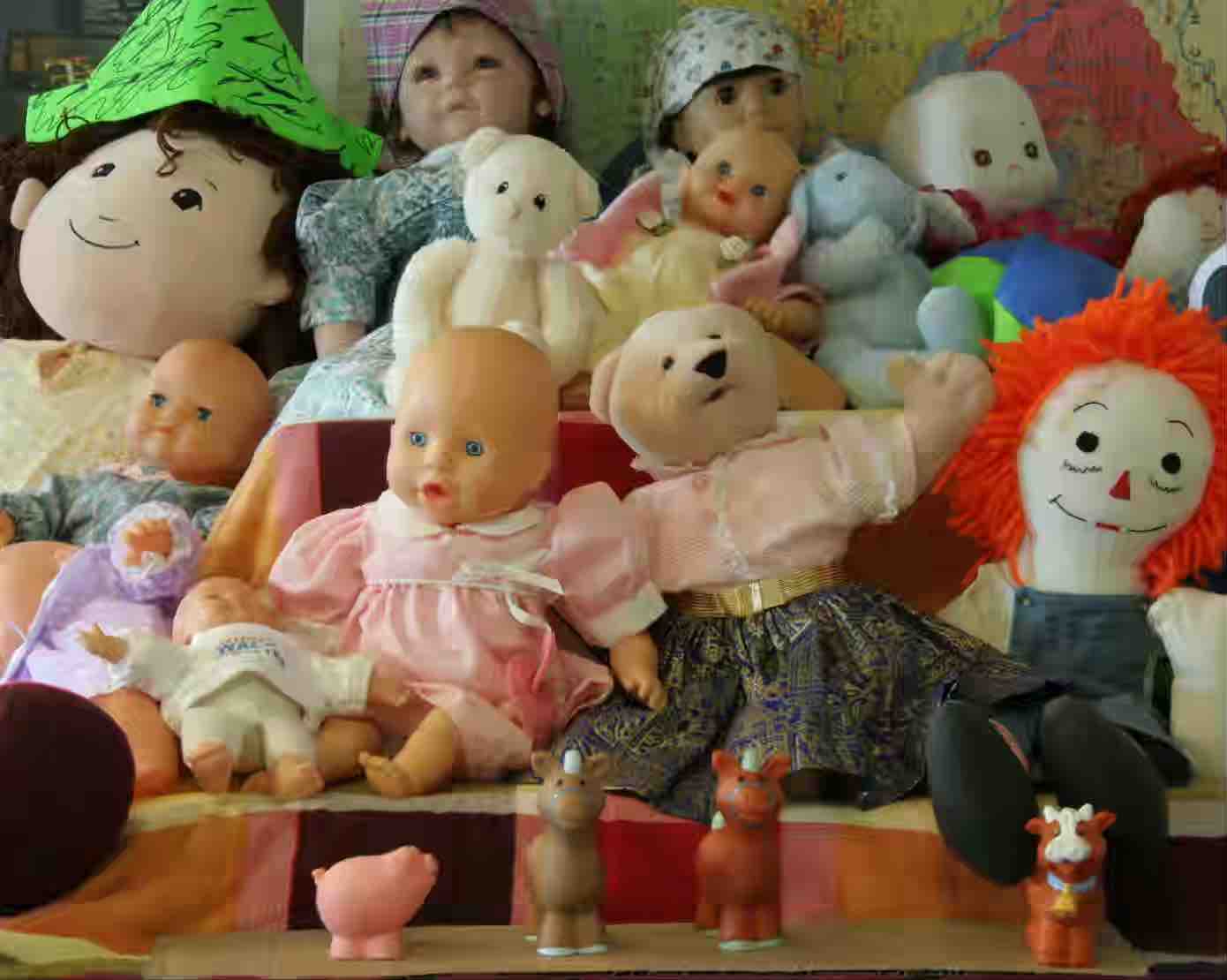} 
		\end{minipage}
		\hspace{-0.13in}
	}\subfigure[Bicubic]{
    		\begin{minipage}[b]{0.194\linewidth}
  		 	\includegraphics[width=1\linewidth]{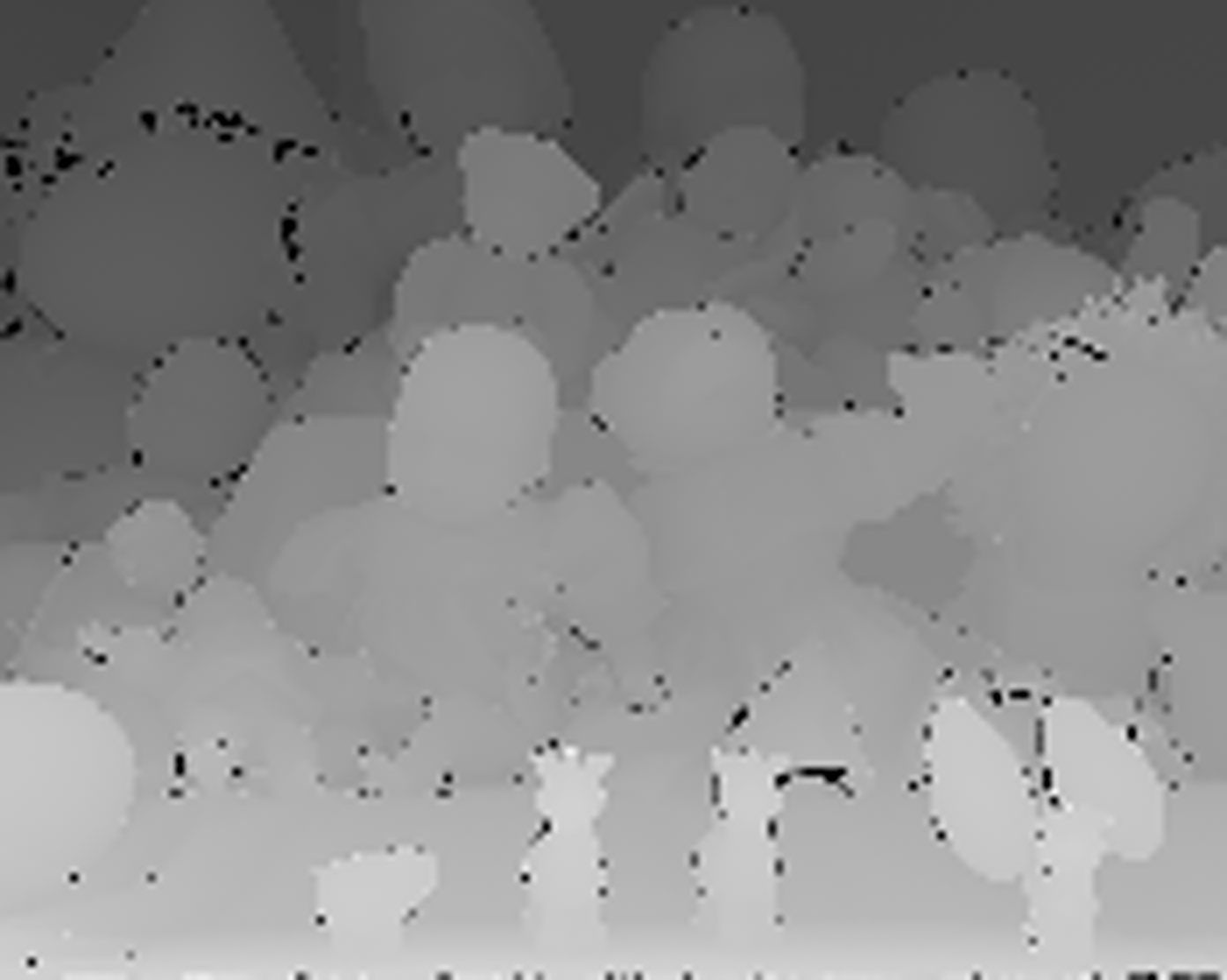}
    		\end{minipage}
    		\hspace{-0.13in}
    	}\subfigure[DJFR]{
    		\begin{minipage}[b]{0.194\linewidth}
\includegraphics[width=1\linewidth]{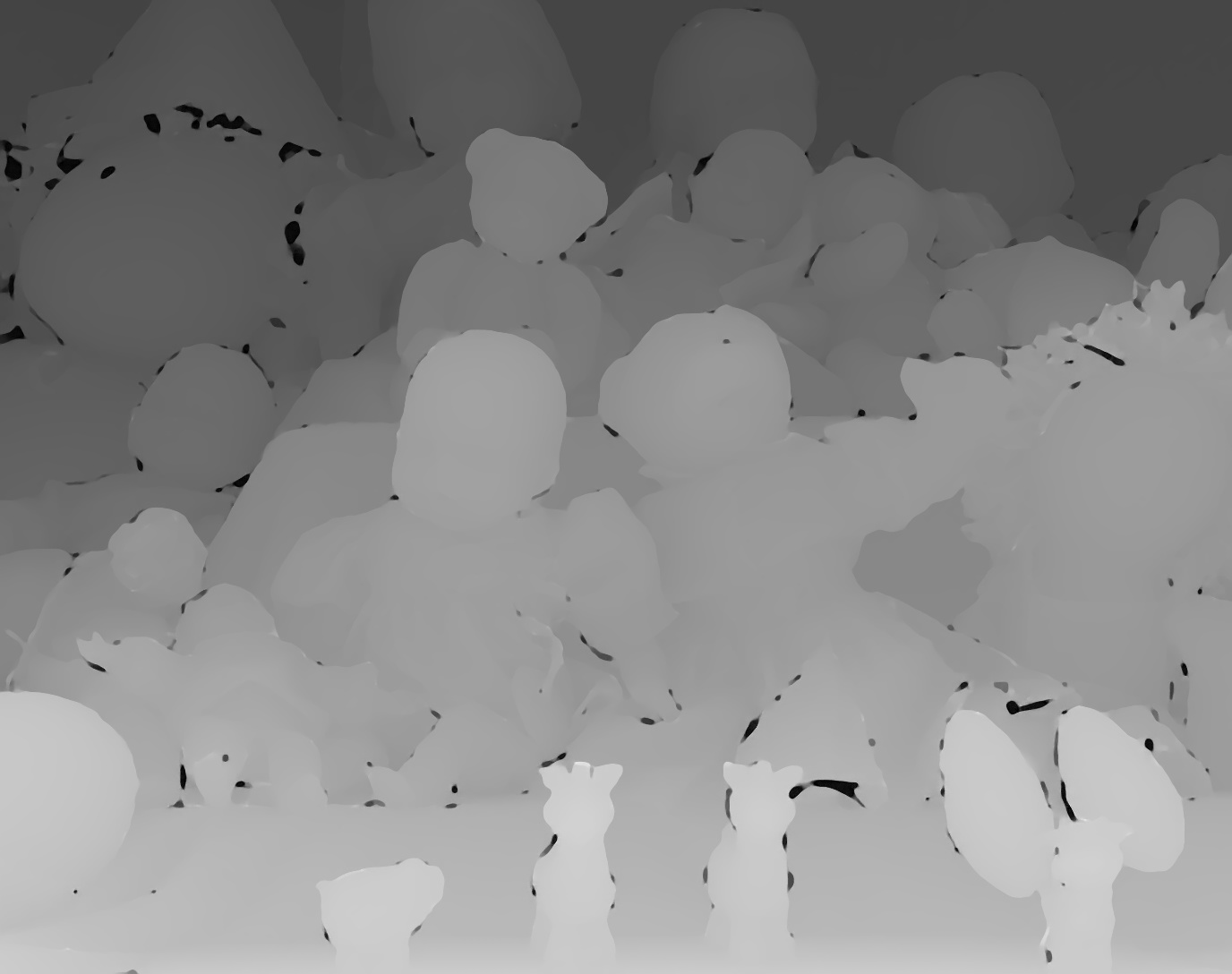}
    		\end{minipage}
    		\hspace{-0.13in}
    	}\subfigure[DKN]{
    		\begin{minipage}[b]{0.194\linewidth}
  		 	\includegraphics[width=1\linewidth]{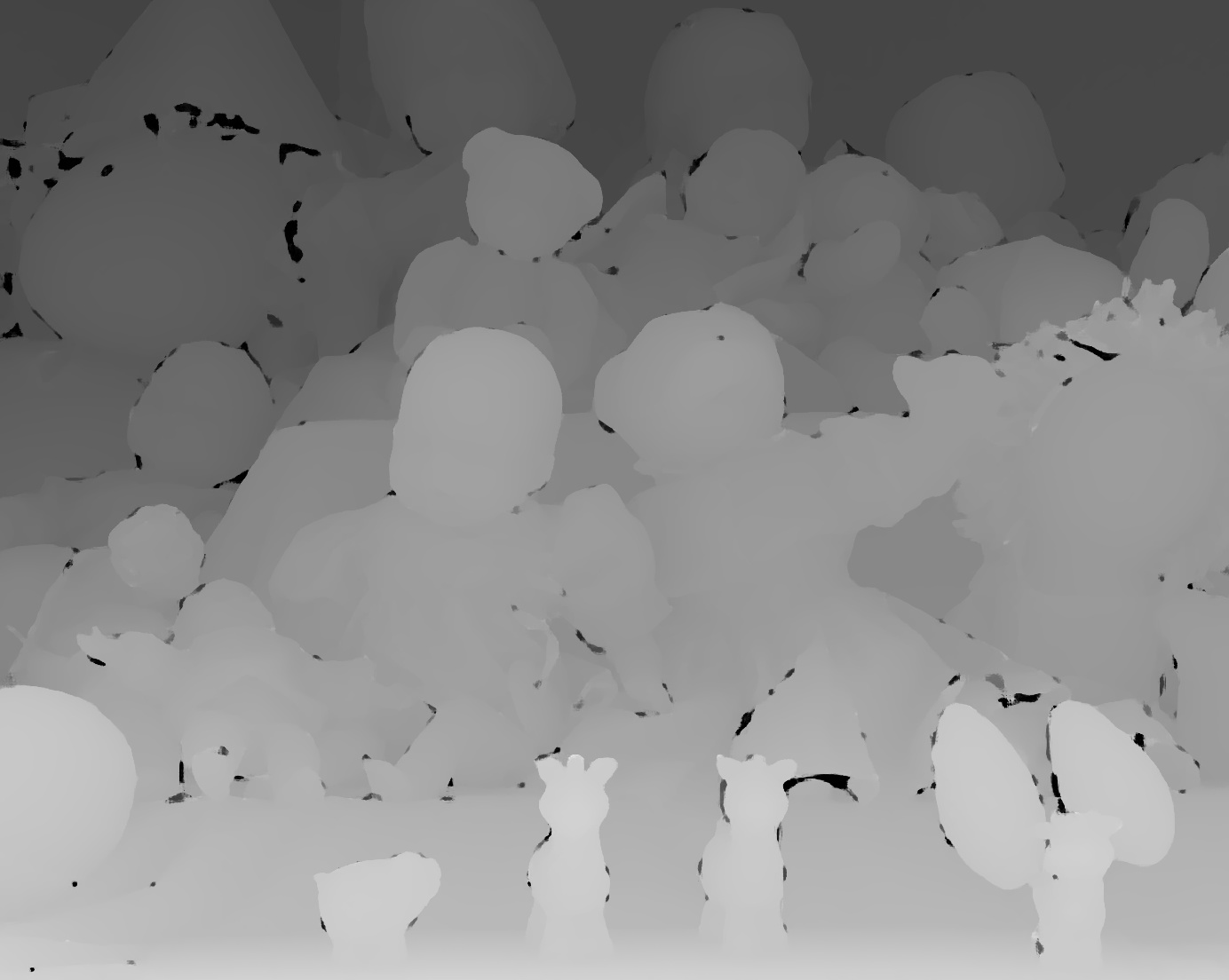}
    		\end{minipage}
    		\hspace{-0.13in}
    	}\subfigure[AHMF]{
    		\begin{minipage}[b]{0.194\linewidth}
  		 	\includegraphics[width=1\linewidth]{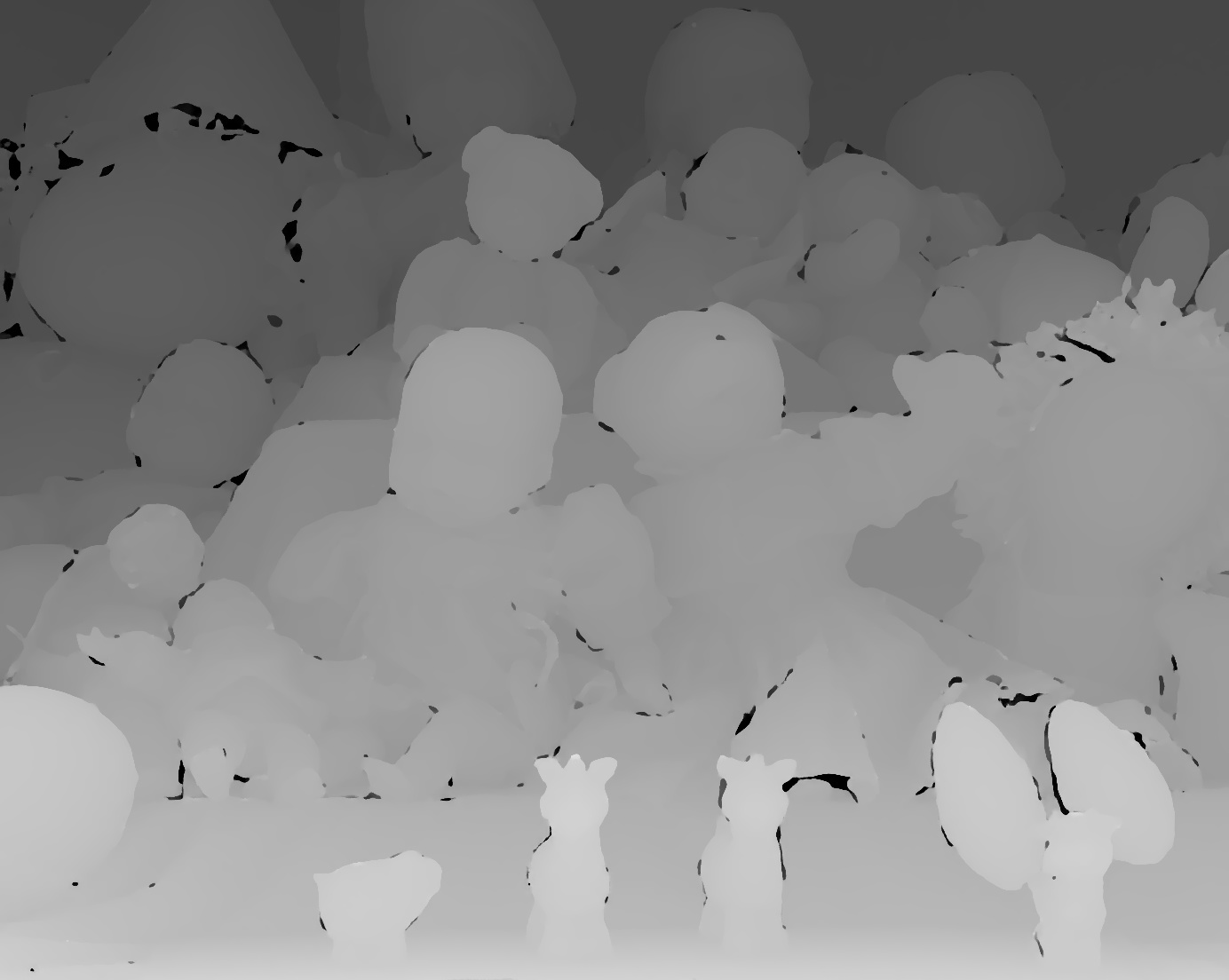}
    		\end{minipage}
    	}
    	
	\end{center}
	\vspace{-.17in}
	\caption{A failure case of $8\times$ depth map super-resolution. (a) guidance image, (b) input depth map with missing data, (c): DJFR~\cite{DJFR}, (d): DKN~\cite{DKN}, (e): Ours. Please enlarge the PDF for more details.}
	\label{fig:fail}
	\vspace{-0.25in}
\end{figure}
\subsection{Limitations}
\label{Limitations}
Although the proposed is capable of generating lower errors and visual appealing results for depth map super-resolution, some cases still pose challenges to our approach, which are listed as follows:
\begin{itemize}
    \item The proposed method tends to generate unsatisfactory results when dealing with raw depth maps with missing data. We show a failure case in Fig.~\ref{fig:fail}, from which we can see that all methods produce unnatural artifacts, and our method fails to reconstruct the missing boundaries. Since our model is trained on the hole-filled data, and the reconstructed results may be sub-optimal to the raw data. However, in the real case, the depth map obtained by consumer cameras suffers from both low-resolution and missing value artifacts. In the future, we will consider how to extend our idea for joint depth map super-resolution and completion.
    \item The upsampling factors of the proposed method are fixed instead of arbitrary scales, which may limit the potential applications of our method. This problem will also be considered in our future work.
\end{itemize}

\section{Conclusion}
\label{con}
In this paper, we presented a novel attention-based hierarchical multi-modal fusion (AHMF) network for guided depth map super-resolution. It consists of a multi-modal attention based fusion (MMAF) and a bi-directional hierarchical feature collaboration (BHFC) module. The MMAF can effectively select and combine relevant information from multi-modal features extracted from input depth and guidance images in a learning manner. The BHFC is designed for optimizing the use of hierarchical features fused by MMAF with the proposed bi-directional feature propagation and collaboration mechanism. Extensive experiments on widely used benchmark datasets demonstrate that the proposed method can achieve state-of-the-art performance in terms of  reconstruction accuracy, inference speed as well as peak GPU memory consumption.



\bibliographystyle{IEEEtran}
\bibliography{Mendeley}

\end{document}